\documentclass[10pt,twocolumn,letterpaper]{article}
\usepackage[dvipsnames]{xcolor}
\usepackage{amsmath}
\usepackage{multirow}
\usepackage{makecell}
\usepackage{amssymb}
\usepackage{adjustbox}
\usepackage{graphicx}
\usepackage{booktabs}
\usepackage{comment}
\usepackage{cases}
\usepackage{setspace}
\usepackage{overpic}
\usepackage{rotating}
\usepackage{subeqnarray}
\usepackage{multirow}
\usepackage{tabularx}
\usepackage{epstopdf}
\usepackage{tikz}
\usepackage{enumerate}
\usepackage{here}
\usepackage{bm}
\usepackage{bbm}
\usepackage[ruled,linesnumbered]{algorithm2e}
\usepackage{array}

\usepackage[pagenumbers]{cvpr}              

\definecolor{cvprblue}{rgb}{0.21,0.49,0.74}
\usepackage[pagebackref,breaklinks,colorlinks,allcolors=cvprblue]{hyperref}

\usepackage{pifont}
\newcommand{\cmark}{\ding{51}}%
\newcommand{\xmark}{\ding{55}}%

\def\paperID{9909} 
\def\confName{CVPR}
\def\confYear{2025}

\title{RipVIS: Rip Currents Video Instance Segmentation Benchmark for Beach Monitoring and Safety}

\author{Andrei Dumitriu$^{1, 2}$, Florin Tatui$^{2}$, Florin Miron$^{2}$, Aakash Ralhan$^{1}$, Radu Tudor Ionescu$^{2}$, Radu Timofte$^{1}$ \\
$^{1}$Computer Vision Lab, CAIDAS \& IFI, University of Würzburg, Germany\\
$^{2}$University of Bucharest, Romania\\
{\tt\small {andrei.dumitriu}@uni-wuerzburg.de}\\
}

\begin{document}

\twocolumn[{%
\renewcommand\twocolumn[1][]{#1}%
\maketitle
\begin{center}
    \centering
    \captionsetup{type=figure}
    
    \parbox{0.246\textwidth}{\centering Aerial - Bird's Eye}%
    \hfill
    \parbox{0.246\textwidth}{\centering Aerial - Tilted}%
    \hfill
    \parbox{0.246\textwidth}{\centering Elevated Beachfront}%
    \hfill
    \parbox{0.246\textwidth}{\centering Water-Level Beachfront}%
    
    \includegraphics[width=0.246\textwidth]{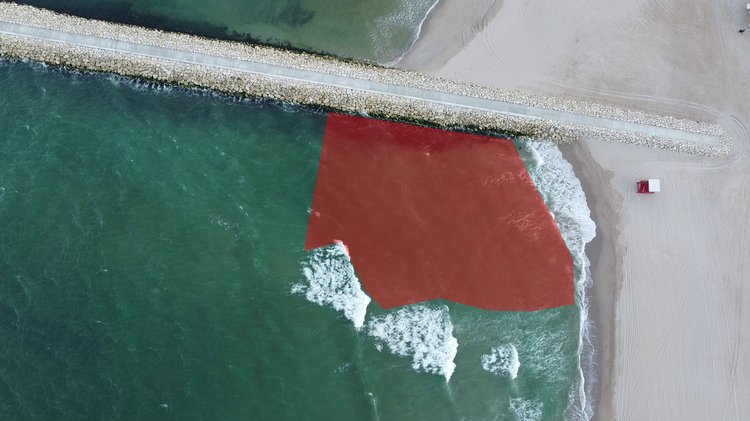}
    \includegraphics[width=0.246\textwidth]{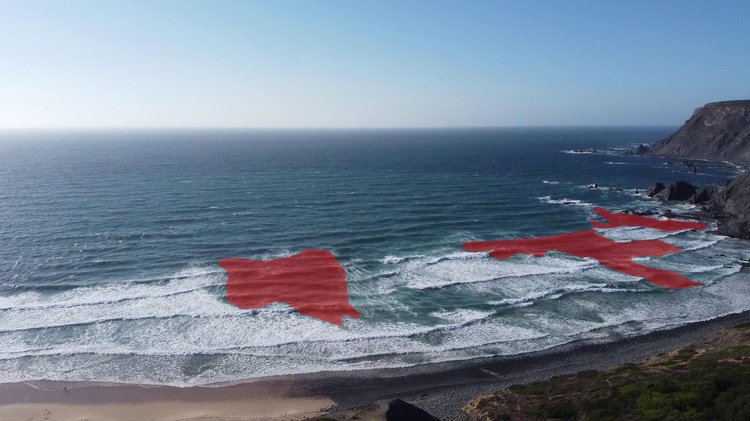}
    \includegraphics[width=0.246\textwidth]{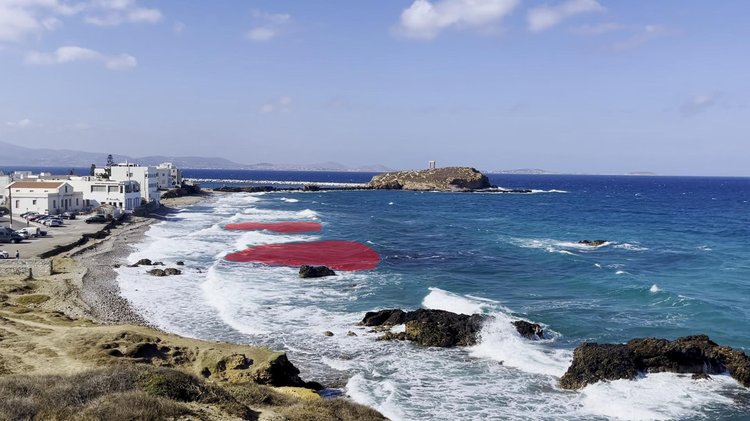}
    \includegraphics[width=0.246\textwidth]{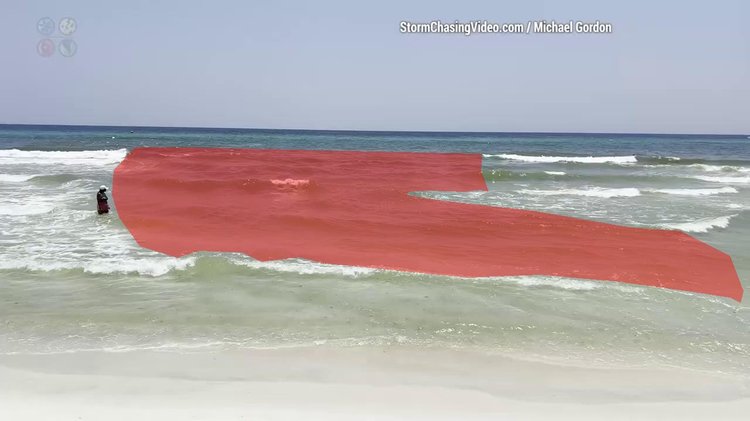}
\end{center}%
\vspace{-7mm} 
\begin{center}
    \centering
    \captionsetup{type=figure}
    \includegraphics[width=0.246\textwidth]{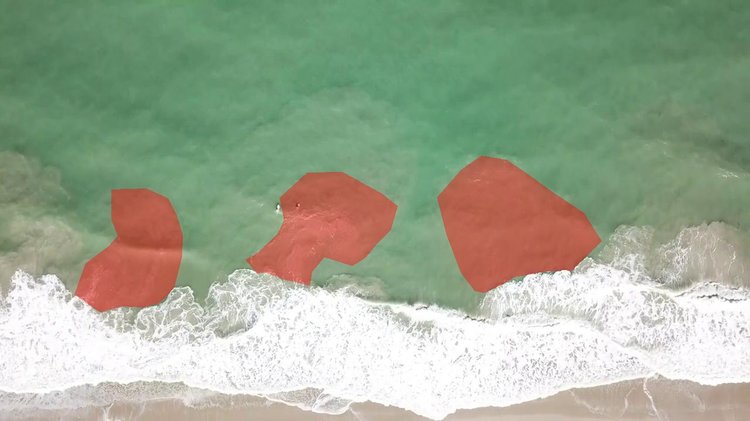}
    \includegraphics[width=0.246\textwidth]{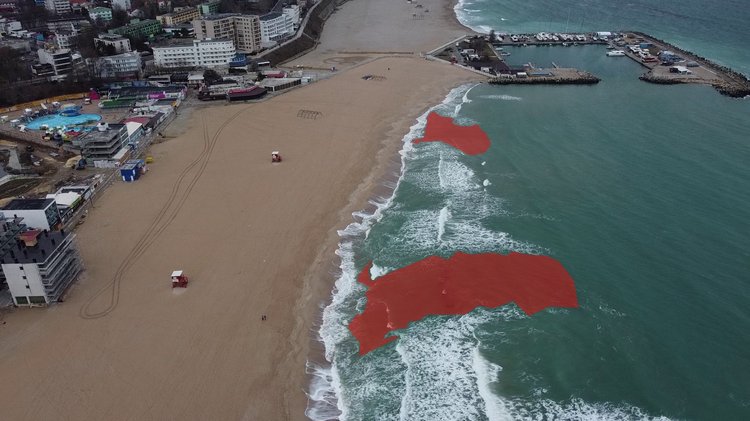}
    \includegraphics[width=0.246\textwidth]{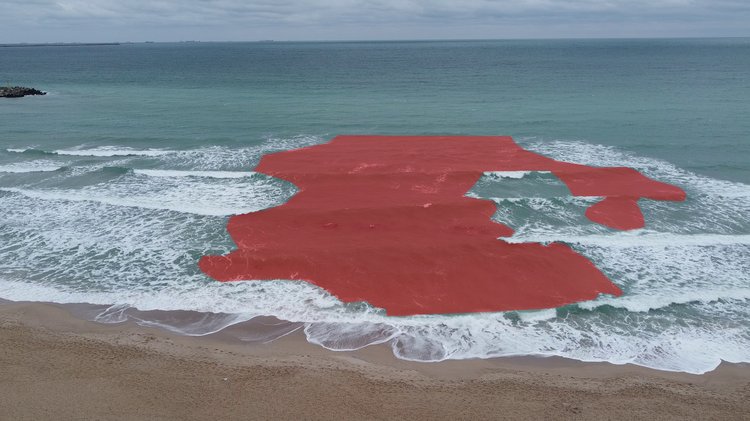}
    \includegraphics[width=0.246\textwidth]{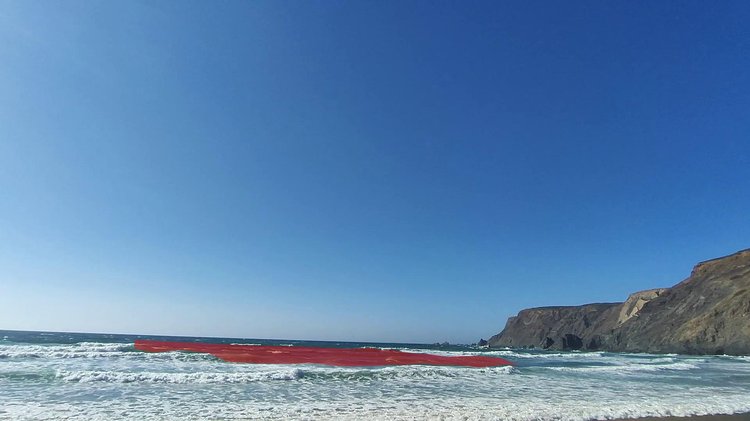}
    \vspace{-0.7cm}
    \captionof{figure}{Examples from our dataset, illustrating the diversity in locations, rip current types, viewpoint elevations and viewing angles. Rip currents are identifiable by distinct wave-breaking patterns, sediment transport, and instances of deflection rip currents. Rip current annotations are shown in red. Additional examples are provided in the supplementary material. Best viewed in color.}
\label{fig:intro_presentation}
\end{center}%
}]

\maketitle

\begin{abstract}
Rip currents are strong, localized and narrow currents of water that flow outwards into the sea, causing numerous beach-related injuries and fatalities worldwide. Accurate identification of rip currents remains challenging due to their amorphous nature and the lack of annotated data, which often requires expert knowledge. To address these issues, we present RipVIS, a large-scale video instance segmentation benchmark explicitly designed for rip current segmentation. RipVIS is an order of magnitude larger than previous datasets, featuring $184$ videos ($212,328$ frames), of which $150$ videos ($163,528$ frames) are with rip currents, collected from various sources, including drones, mobile phones, and fixed beach cameras. Our dataset encompasses diverse visual contexts, such as wave-breaking patterns, sediment flows, and water color variations, across multiple global locations, including USA, Mexico, Costa Rica, Portugal, Italy, Greece, Romania, Sri Lanka, Australia and New Zealand. Most videos are annotated at $5$ FPS to ensure accuracy in dynamic scenarios, supplemented by an additional $34$ videos ($48,800$ frames) without rip currents. We conduct comprehensive experiments with Mask R-CNN, Cascade Mask R-CNN, SparseInst and YOLO11, fine-tuning these models for the task of rip current segmentation. Results are reported in terms of multiple metrics, with a particular focus on the $F_2$ score to prioritize recall and reduce false negatives. To enhance segmentation performance, we introduce a novel post-processing step based on Temporal Confidence Aggregation (TCA). RipVIS aims to set a new standard for rip current segmentation, contributing towards safer beach environments. We offer a benchmark website to share data, models, and results with the research community, encouraging ongoing collaboration and future contributions, at \url{https://ripvis.ai}.
\end{abstract}

\vspace{-0.4cm}
\section{Introduction}
Rip currents are powerful, fast-moving surface currents that flow seaward from the shore. Detecting and understanding rip currents is critical, as accurate detection can prevent fatalities. Each year, numerous lives are lost globally due to these dangerous phenomena, emphasizing the urgent need for effective solutions. These hazardous currents are common along coastlines worldwide, including oceans, seas, and large lakes \cite{da2003analysis, lushine1991study, brewster2019estimations, brander2013brief, castelle2016rip}. They vary widely in size and speed, influenced by nearshore hydrodynamics, underwater morphology, and occasionally, by human activity near coastal structures \cite{brander2000morphodynamics, dusek2013rip}. Some rip currents reach speeds of up to $8.7$ km/h, faster than even Olympic swimmers \cite{noaa2023ripcurrents}. The main risk lies not only in their strength, but also in the widespread lack of public awareness on how to recognize and respond to them. Often, individuals caught in a rip current panic and attempt to swim directly against it, leading to exhaustion and even drowning. Effective safety measures include swimming parallel to the shore to escape and, ideally, early detection systems to warn beachgoers.

In computer vision, detection and segmentation methods for visual data have advanced considerably \cite{he2017mask, Jocher_Ultralytics_YOLO_2023, liang2022cbnet, wang2022image, liang2023mask, wei2022contrastive, fang2023eva, liang2022cbnet, kirillov2023segment, ravi2024sam, mei2024slvp}, largely due to the availability of high-quality datasets focused on object detection and segmentation in images \cite{lin2014microsoftCOCO, Everingham15Pascal, geiger2013vision, cordts2015cityscapes}. Recently, video instance segmentation has emerged as an active area of research, with datasets like DAVIS \cite{perazzi2016benchmarkDAVIS} and YouTube-VIS \cite{Yang2019vis}, supporting ongoing challenges \cite{vis2021, Yang2019vis}. Despite these advancements and the growing interest in automatic rip current detection \cite{choi2024explainable, dumitriu2023rip, de2023ripviz, rashid2023reducing, zhu2022yolo, mori2022flow, mcgill2022flow, rampal2022interpretable, desilva2021frcnn, rashid2021ripdet, rashid2020ripnet, maryan2019machine, philip2016flow}, the complex rip current detection task remains understudied. 
The primary barrier to further progress is the lack of sufficient high-quality data. Collecting and annotating this data is difficult due to several factors:
\begin{enumerate}
    \item Rip currents vary widely in appearance, being influenced by environmental factors, such as water body, beach structure, weather, and bathymetric conditions. Gathering diverse data requires global efforts across varied weather conditions, including hostile ones.
    \item While some rip currents are visually distinctive, others require expert knowledge to be identified (see Figure \ref{fig:dataset_diversity}).
    \item Rip currents are best observed from elevated viewpoints, often requiring the use of drones or elevated positions, like towers or cliffs. Not all beaches have elevated locations, making drones essential in many cases.
    \item Accurate annotation for instance segmentation of rip currents is challenging and labor-intensive, requiring expertise in rip current dynamics, alongside computer vision skills, patience and attention to details.
    \item Rip currents are amorphous objects (see Figure \ref{fig:intro_presentation} and \ref{fig:dataset_diversity_masked}). Unlike objects with consistent shapes and clear boundaries, rip currents are continuously changing in shape and form, making them particularly challenging to detect. While some amorphous objects, like fire or smoke, also undergo continuous shape changes, they usually stand out distinctly from their background, making them easier to identify. In contrast, rip currents blend seamlessly into the large water environment, often appearing as subtle patterns within a dynamic, constantly shifting background. This unique characteristic of rip currents demands diverse and extensive data to facilitate accurate and reliable detection.

\end{enumerate}
To address this problem, we introduce the RipVIS dataset. The culmination of three years of work and a team of over 30 people, involved in both data collection and annotation, has successfully materialized into this dataset. With diversity in types of rip currents, elevation, conditions and locations, RipVIS is a high-quality dataset, which is an order of magnitude larger than any existing alternative. 

In summary, our contribution is fourfold:
\begin{itemize}
    \item \textbf{RipVIS benchmark}: We introduce RipVIS, an open benchmark for rip current instance segmentation, featuring $184$ videos ($212,328$ frames), out of which $150$ videos ($163,528$ frames) contain rip currents annotated at an average sampling rate of $5$ FPS, and $34$ videos ($48,800$ frames) are without rip currents.
    \item \textbf{Baseline models and analysis}: We establish baselines using several state-of-the-art instance segmentation methods, analyzing their performance on this challenging dataset and highlighting the need for improvement.
    \item \textbf{Temporal Confidence Aggregation}: We propose a Temporal Confidence Aggregation (TCA) technique, which boosts segmentation quality by incorporating temporal consistency across frames, improving the results, both qualitatively and quantitatively.
    \item \textbf{Benchmark website and community engagement}: We host RipVIS on a dedicated website (\url{https://ripvis.ai}), promoting community collaboration and inviting researchers to contribute with new data and models. Each submitted video is carefully annotated, with credits given to both contributors and annotators, reinforcing our commitment to continually enhancing rip current segmentation quality for improved beach safety.
\end{itemize}

\section{Related Work}
Rip currents have been extensively researched in the natural sciences \cite{leatherman2017techniques, sonu1972field, inman1980field, brander2000morphodynamics, dusek2013rip, zhang2021rip, valipour2018analytical}. Traditional observation techniques include visual monitoring and camera-based systems \cite{prodger2012argus, holman2007history, dusek2019webcat}. Precision tracking with GPS-equipped drifters or floating devices \cite{castelle2014rip, castelle2016rip, short1994rip} is effective, but costly, location-dependent, and unsuitable for flash rip detection. Newer tools, such as laser rangefinders and drones with tracer dye, offer flexibility and broader perspectives \cite{clark2014aerial, kim2021analysis, pujianiki2020application}. In contrast, machine learning (ML) approaches are cost-effective, scalable, and capable of real-time detection, making rip current detection more accessible for public safety applications. We further discuss related studies introducing datasets to train ML methods for rip current detection, as well as studies proposing such methods.

\begin{table*}[t]
\centering
\renewcommand{\arraystretch}{1.1}
\setlength{\tabcolsep}{2.5pt} 
\begin{adjustbox}{width=\textwidth}
\begin{tabular}{|l|c|c|c|c|c|c|c|}
\hline
\multirow{2}{*}{\textbf{Dataset}}  & \multirow{2}{*}{\textbf{Total}} & \textbf{With} & \textbf{Without} & \multirow{2}{*}{\textbf{Train}} & \multirow{2}{*}{\textbf{Validation}} & \multirow{2}{*}{\textbf{Test}} & \textbf{Segmentation} \\
 &  & \textbf{Rip Currents} & \textbf{Rip Currents} & & &  & \textbf{Annotations} \\ 
\hline
\hline
Maryan~\etal (2019) \cite{maryan2019machine} & 5,310 images & 514 images & 4,796 images & 4,779 images (10-fold) & - & 531 images (10-fold) & \xmark \\ \hline
de Silva~\etal (2021) \cite{desilva2021frcnn} & 20,482 images & \begin{tabular}[c]{@{}c@{}} 10,793 images  \end{tabular} & 9,689 images & 2,440 images & - & \begin{tabular}[c]{@{}c@{}}23 videos\\ 18,042 frames \end{tabular} & \xmark  \\ \hline
YOLO-Rip (2022) \cite{zhu2022yolo} & 3,793 images & 2,486 images & 1,307 images & 3,793 images & - & same as de Silva~\etal \cite{desilva2021frcnn} & \xmark  \\ \hline
Dumitriu~\etal (2023) \cite{dumitriu2023rip}& 37,057 frames & \begin{tabular}[c]{@{}c@{}}26,761 frames \end{tabular} & \begin{tabular}[c]{@{}c@{}}10,296 frames \end{tabular} & 3,396 images (10-fold) & 377 images (10-fold) & \begin{tabular}[c]{@{}c@{}}25 videos\\ 33,284 frames \end{tabular} & \cmark  \\ \hline
\textbf{RipVIS (ours)} & \textbf{\begin{tabular}[c]{@{}c@{}}184 videos \\ 212,328 frames \end{tabular}} & \textbf{\begin{tabular}[c]{@{}c@{}}150 videos\\ 163,528 frames\end{tabular}} & \textbf{\begin{tabular}[c]{@{}c@{}}34 videos\\ 48,800 frames\end{tabular}} & \textbf{\begin{tabular}[c]{@{}c@{}}112 videos\\ 147,802 frames\end{tabular}} & \textbf{\begin{tabular}[c]{@{}c@{}}36 videos\\ 32,566 frames\end{tabular}} & \textbf{\begin{tabular}[c]{@{}c@{}}36 videos\\ 31,960 frames\end{tabular}} & \cmark  \\ \hline
\end{tabular}
\end{adjustbox}
\vspace{-0.2cm}
\caption{Comparison of public rip currents datasets. As observed, our dataset is an order of magnitude larger than any other publicly available dataset, with increased diversity and a train-validation-test split. All datasets, including ours, have bounding box annotations.}
\label{tab:rip_current_datasets}
\vspace{-0.3cm}
\end{table*}

\subsection{Datasets}

As shown in Table \ref{tab:rip_current_datasets}, a number of rip currents datasets are publicly available.
Maryan~\etal\cite{maryan2019machine} introduced a dataset containing 514 rip channel examples, including test samples. The dataset consists of small $24 \times 24$ pixel rip channel images, extracted from larger $1334\times1334$ timex images sourced from the Oregon State University beach imagery archive \cite{osuwebsite}. These timex images were orthorectified and time-averaged over $1,200$ frames at $2$ Hz, covering 10-minute intervals. Rip channel samples were isolated using the GIMP image editor, resized to $24 \times 24$ pixels, and converted to grayscale to reduce the impact of varying lighting conditions on model performance. To expand the dataset for deep learning applications, data augmentation techniques were applied, resulting in a dataset of over $4,000$ rip channel images. This dataset facilitated the training of a CNN, but it was also instrumental in training and evaluating several rip current detection algorithms, including studies by Rashid~\etal\cite{rashid2020ripnet, rashid2021ripdet, rashid2023reducing}.

In their study, de Silva~\etal\cite{desilva2021frcnn} introduced a training dataset primarily sourced from Google Earth, consisting of high-resolution aerial images of beach scenes both with and without rip currents. The dataset includes $1,740$ rip current images and 700 non-rip current images, with sizes ranging from $1086 \times 916$ to $234 \times 234$ pixels. Each rip current image was annotated with axis-aligned bounding boxes to serve as ground truth. Additionally, de Silva~\etal\cite{desilva2021frcnn} compiled a test dataset of $23$ videos with $18,042$ frames in total, out of which only $9,053$ frames contained rip currents. While the static images were captured from a high-elevation viewpoint, the test videos were recorded from a lower perspective, with resolutions varying between $1280 \times 720$ and $1080 \times 920$ pixels. Ground-truth annotations for these images were verified by an expert from NOAA, although the videos only received categorical labels without frame-level annotations. The dataset was used to train a Faster R-CNN \cite{ren2015faster} model, with frame averaging applied as a temporal aggregation technique for improved bounding box prediction and detection accuracy. While the bounding box annotations of Silva~\etal\cite{desilva2021frcnn} provided valuable insights, they lack the granularity of instance segmentation, limiting the precision of rip current localization.

The YOLO-Rip dataset \cite{zhu2022yolo} was created by expanding the dataset of de Silva~\etal\cite{desilva2021frcnn}. The authors collected additional real-world beach scene images along the South China coast, resulting in a total of 1,352 high-resolution images. Of these, 746 depict rip currents, while 606 do not, with image resolutions ranging from $4000 \times 2250$ to $480 \times 360$ pixels. Rip current boundaries in the images were annotated with axis-aligned bounding boxes. The extended dataset was designed to enhance model performance in recognizing rip currents across diverse image types, thereby improving its practical applicability in real-world scenarios.

Dumitriu~\etal \cite{dumitriu2023rip} introduced the first instance segmentation dataset for rip currents, specifically focusing on annotating images and videos to delineate rip currents with high precision. They extended the work of de Silva~\etal\cite{desilva2021frcnn} and Zhu~\etal\cite{zhu2022yolo} by adding detailed polygonal annotations to $2,466$ aerial images of rip currents sourced from Google Maps. In addition to these static images, Dumitriu~\etal included $17$ video sequences recorded at the Black Sea, totaling $24,295$ frames. These videos capture rip currents from an elevated and top perspective, with sampled frames annotated for segmentation. While this dataset marked a significant step forward by enabling instance segmentation, it is limited in geographic diversity, as all videos were recorded from a single location (the Black Sea).

\subsection{Detection Methods}

In recent years, the automatic identification of rip currents has garnered increasing attention \cite{choi2024explainable, dumitriu2023rip, de2023ripviz, rashid2023reducing, zhu2022yolo, mori2022flow, mcgill2022flow, rampal2022interpretable, desilva2021frcnn, rashid2021ripdet, rashid2020ripnet, philip2016flow}. Studies in this area generally fall into two categories: those using bounding boxes for rip current detection and those capturing the full shape. Most relevant approaches have been detailed along with the datasets they were published with. All of the approaches rely on video and image data \cite{prodger2012argus, holman2007history, dusek2019webcat, holland1997practical}, with some using time-exposure or “timex” images to highlight rip current patterns over time \cite{nelko2011rip, lippmann1989quantification}. Rashid~\etal\cite{rashid2020ripnet} used an anomaly detection framework, RipNet, to improve accuracy by reducing the need for additional negative samples. The same team later introduced RipDet and RipDet+ \cite{rashid2021ripdet, rashid2023reducing}, treating the task as a detection problem. Pitman~\etal \cite{pitman2016synthetic} employed synthetic imagery, but this often led to underestimations. Liu~\etal \cite{liu2019lifeguarding} utilized threshold and HSV-based segmentation, limited to sediment-visible currents. 

Optical flow has proven useful for rip current detection, particularly in cases lacking segment-level annotation. Philip~\etal \cite{philip2016flow} employed the Lukas-Kanade optical flow algorithm to determine water flow direction and isolate rip currents, though this approach requires a stable platform and captures only the main flow direction. Mori~\etal \cite{mori2022flow} enhanced flow visualization fields to improve detection, but similarly, their approach relies on a stationary camera. RipViz \cite{de2023ripviz} combines optical flow with an LSTM autoencoder to detect rip currents as flow anomalies in stationary videos, offering an intuitive visualization of dangerous currents. McGill~\etal \cite{mcgill2022flow} applied Farneb\"{a}ck optical flow on timex images, improving accuracy in channel detection, though the method is time-consuming and sensitive to camera positioning and beach morphology. While optical flow techniques enable rip current detection without wave-breaking patterns, they are generally limited to specific camera setups. In contrast, our dataset includes diverse camera types and orientations, allowing for a broader applicability.

\noindent
\textbf{Bounding box detection vs.~segmentation.} While bounding boxes provide valuable information, due to the amorphous property of rip currents, boxes can either include a significant amount of background information or leave out a significant part of the rip current, making precise beach monitoring a much more difficult task.

\section{RipVIS Dataset and Benchmark}

\subsection{General Description}

\begin{figure}[t!]
    \centering
    \includegraphics[width=\columnwidth,scale=0.5]{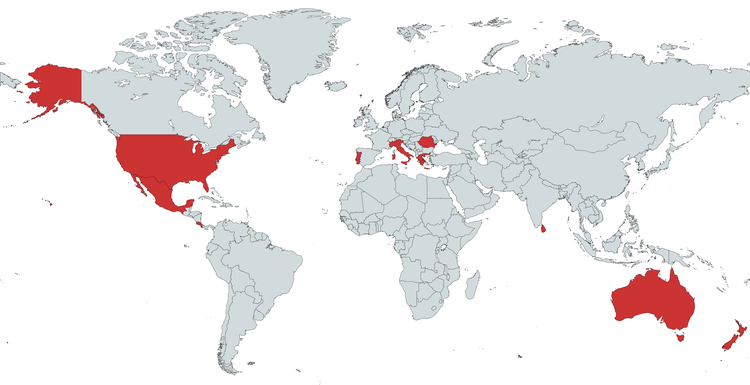}
    \vspace{-0.2cm}
    \caption[Rip Currents Map]{Map of the countries present in the RipVIS dataset. From left to right: USA, Mexico, Costa Rica, Portugal, Italy, Greece, Romania, Sri Lanka, Australia and New Zealand. Created with mapchart.net.}
    \label{fig:rip_current_map}
    \vspace{-0.3cm}
\end{figure}

The RipVIS Benchmark (\textbf{Rip} Currents \textbf{V}ideo \textbf{I}nstance \textbf{S}egmentation) is a large-scale and high-quality dataset designed to address the limitations of previous rip current datasets in terms of diversity, annotation quality, and data structure (see Table \ref{tab:rip_current_datasets}). With $184$ videos totaling $212,328$ frames, RipVIS is the most comprehensive dataset for rip current instance segmentation to date. It contains $150$ videos  ($163,528$ frames) featuring rip currents and $34$ videos ($48,800$ frames) without rip currents, allowing for both positive and negative sample training. The dataset was collected from diverse locations worldwide—including the USA, Mexico, Costa Rica, Portugal, Italy, Greece, Romania, Sri Lanka, Australia and New Zealand — capturing rip currents across varied visual contexts, environmental conditions, and geographic landscapes 
(see Figures \ref{fig:intro_presentation} and \ref{fig:rip_current_map}).

RipVIS introduces the first rip current dataset with a dedicated train-val-test split, manually curated by computer vision experts to mirror the data distribution accurately and prevent overfitting. This structured split, including a validation set, is crucial for effective hyperparameter tuning and robust model development, addressing a key limitation of prior datasets. Without it, models risk overfitting from tuning on test data or underperforming due to lack of tuning. Expert selection ensures balanced, reliable splits, enhancing evaluation consistency and laying a solid groundwork for advancing rip current detection and segmentation research.
\subsection{Sources}

RipVIS is compiled from multiple sources, with $76$ videos recorded directly by the authors using drones and phone cameras at different locations worldwide. An additional $87$ videos were collected from the Internet, providing real-world variability, while $21$ videos were sourced from the de Silva~\etal dataset \cite{desilva2021frcnn}. Each video source and annotator is credited individually, ensuring transparency and traceability across the dataset.

\subsection{Video and Rip Current Variety}

Our dataset captures a diverse range of rip current characteristics from multiple perspectives, enhancing its utility for comprehensive analysis. The videos encompass four types of elevation and orientation (see Figure \ref{fig:intro_presentation}): water-level beachfront (captured at beach level), elevated beachfront (from stationary elevated points, like hills, buildings, or low-altitude drones), aerial tilted view (drone recordings at an inclined angle), and aerial bird's-eye view (high-altitude drone recordings). This range provides a robust basis for detecting rip currents from both traditional and challenging angles, a necessary improvement over previous datasets that were limited in this perspective.


The dataset includes rip currents with considerable temporal and spatial variability, driven by shoreline geometry, underwater morphology, wave conditions, and tidal forces. Following the classification of Castelle~\etal \cite{castelle2016rip}, RipVIS features primarily bathymetrically-controlled rip currents, which are shaped by underwater sandbars or channels, and boundary-controlled rip currents, which flow along the edges of anthropogenic structures like piers or jetties. These rip currents were identified primarily by gaps in wave-breaking patterns or offshore sediment transport. While the dataset does not include flash or traveling rip currents—due to their unpredictability and transient nature—it focuses on stable rip currents that vary in strength but remain consistent in location, offering a structured basis for segmentation.

\subsection{Annotations}

The annotation process was carried out by a team of $30$ volunteers, trained and overseen by two academic experts with extensive experience in in-situ rip current measurements and analysis. Each volunteer received on-site training, and the experts annotated the first frame of each video using Roboflow \cite{roboflow2024} as a guide for consistency. All annotations were subsequently reviewed and validated by both experts, achieving an inter-annotator Cohen's $\kappa$ agreement of 0.82 (almost perfect agreement) on the entire dataset. This high agreement rate underscores the quality and reliability of the annotations.

The dataset includes pixel-level annotations for rip currents, with $15,784$ frames manually annotated using polygons for instance segmentation, totaling $25,298$ rip current instances (an average of $1.6$ rip currents per frame). Interpolated annotations were generated for intermediary frames to capture dynamics between manual annotations, with all interpolations verified for accuracy. This approach allows the dataset to provide $163,528$ frames with rip currents, retaining the average of $1.6$ rip currents per frame.

Out of the $150$ annotated videos, $28$ required major revisions from the experts, necessitating re-annotation and closer supervision to maintain high annotation standards. The sampling rate depends on video dynamics, varying from $1$ to $30$ FPS, with the most common being $5$ FPS. This rate is adjusted to suit video characteristics. For instance, stationary videos are annotated at a lower frame rate than moving camera footage, ensuring efficient yet accurate annotations.

Annotations capture various states of rip currents, with most videos containing at least one visible rip current in each frame. Some videos feature instances where the rip current is temporarily obscured or less visible due to lighting or environmental factors, reflecting realistic conditions for model training. In these cases, since we are performing a frame-by-frame annotation, the rip currents have not been annotated, as they are not visually obvious. The diversity in annotation detail and sampling frequency helps create a comprehensive dataset that accommodates different model requirements and evaluation scenarios, setting RipVIS apart as a pioneering benchmark in rip current detection and segmentation research.

Furthermore, the dataset also enables advanced analysis of rip current behavior over time. This positions RipVIS as a valuable resource for both computer vision researchers and coastal scientists, facilitating the development of robust detection models and contributing to improved rip current forecasting and public safety measures.

\section{Methods}

\begin{figure*}[t]
    \centering
    \includegraphics[width=1\textwidth]{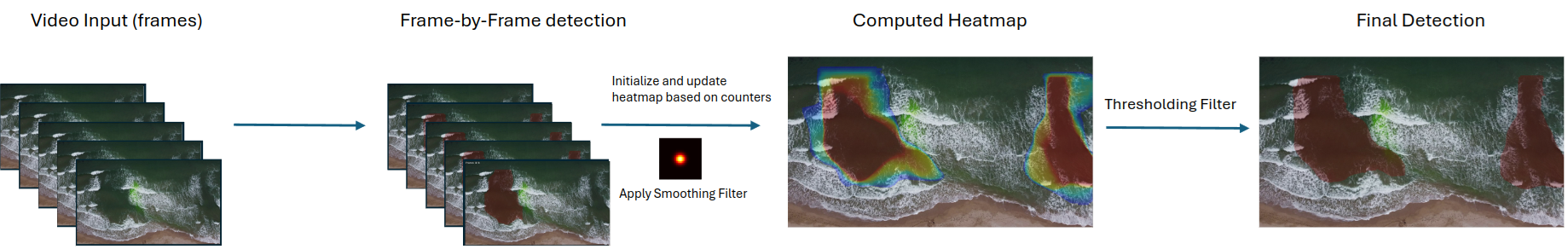}
    \caption{The proposed Temporal Confidence Aggregation (TCA) process, simplified. TCA leverages temporal coherence through downsampling, instance tracking, temporal smoothing, and hysteresis thresholding to create a stabilized temporal heatmap. Best viewed in color. }
\label{fig:tca_architecture}
\end{figure*}%

Segmentation is a much harder task than detection, and segmenting rip currents, due to their amorphous nature, is even harder. In this study, we evaluate a selection of popular, state-of-the-art segmentation models, focusing on their application to the RipVIS dataset. We distinguish between two-stage and one-stage detectors based on their architectural approach to the task. We then describe our TCA, used to improve both qualitative and quantitative results. We have selected the following methods based on their previous results, popularity and availability. 
For implementation details, see the supplementary material.

\subsection{Two-Stage Detectors}

\noindent
\textbf{Mask RCNN.} Mask RCNN \cite{he2017mask} is a two-stage instance segmentation model that combines region proposals with pixel-level segmentation masks, making it a widely-used baseline for segmentation tasks.

\noindent
\textbf{Cascade MASK RCNN.} Cascade Mask RCNN \cite{cai2018cascade} extends Mask RCNN with a multi-stage refinement approach, progressively improving bounding box and mask quality through stricter IoU thresholds at each stage.

\subsection{One-Stage Detectors}



\noindent
\textbf{YOLO11.} YOLO11 \cite{Jocher_Ultralytics_YOLO_2023} represents the latest evolution in the YOLO series, incorporating key improvements, particularly the C2PSA (Cross-Stage Partial Spatial Attention) module, which enhances spatial sensitivity and is useful for detecting small or partially occluded objects. YOLO11 is optimized for efficiency, achieving faster training times and reduced inference latency, which supports real-time applications. 

\noindent
\textbf{SparseInst.} SparseInst \cite{Cheng2022SparseInst} is a one-stage fully convolutional instance segmentation framework designed for real-time performance, leveraging sparse instance activation maps to directly segment objects in a single pass, without region proposals or post-processing. In our work, we implemented SparseInst with ResNet-50 (R-50) and ResNet-101 (R-101) backbones, as well as the transformer-based PVTv2-B1 backbone.

\subsection{Temporal Confidence Aggregation (TCA)}
We propose TCA, a pixel-level post-processing technique aimed at improving segmentation consistency over video frames, especially for amorphous, dynamic phenomena such as rip currents. Rip currents continuously change shape and intensity, making frame-by-frame segmentation noisy and inconsistent. TCA addresses this issue by leveraging temporal coherence, aggregating confidence scores across frames to create a ``temporal heatmap'' that stabilizes detection (see Figure \ref{fig:tca_architecture}). The method works by:
\begin{itemize}
    \item \textbf{Downsampling:} To enable efficient pixel-level analysis, prediction masks are downsampled, reducing computational complexity, while preserving spatial relationships.
    \item \textbf{Instance tracking:} Tracking of individual instances is performed by computing the Intersection over Union (IoU) between masks from consecutive frames. The Hungarian algorithm \cite{hungarian_algorithm} is then employed to optimally match previous and current instances, ensuring consistent identity assignment throughout the sequence.
    \item \textbf{Temporal smoothing:} TCA maintains a heatmap for each tracked instance, where pixel confidence scores incrementally increase with repeated detections and gradually decay in their absence. 

    \item \textbf{Thresholding:}
    To generate final masks, TCA applies hysteresis thresholding to the accumulated heatmaps, adapting the dual-threshold technique introduced in the Canny edge detector \cite{canny1986computational}. Pixels that surpass a high threshold are identified as strong object regions, serving as seeds to include neighboring pixels that exceed a lower threshold, while those below are excluded. 
\end{itemize}
The benefits of TCA for segmentation are (see Figure \ref{fig:testing_results}):
\begin{itemize}
    \item \textbf{Noise reduction:} TCA effectively reduces false positives by exploiting temporal accumulation to filter out noise, requiring sustained evidence across multiple frames to confirm object presence.
    \item \textbf{False negative mitigation:} TCA also reduces false negatives by leveraging the temporal heatmap to recover pixels missed in individual frames, where transient noise or occlusions might obscure detection, provided that they show sustained presence across the aggregated scores. 
    \item \textbf{Refined segmentation masks:}
    TCA refines predictions by smoothing instance boundaries over time, yielding more coherent and precise segmentation masks compared to inconsistent per-frame outputs.
    
\end{itemize}



Our aggregated confidence map offers a clearer and more stable representation of rip current locations over time, enhancing visualization. Integrating TCA with instance segmentation models improves the $F_2$ score by prioritizing fewer false negatives, making it ideal for safety-critical beach monitoring applications. TCA provides an effective solution for accurately determining the shape and position of rip currents in dynamic coastal environments.

\section{Experiments and Results}

\begin{table*}[t]
\centering
\renewcommand{\arraystretch}{1.1}
\small 
\setlength{\tabcolsep}{4pt} 
\begin{tabular}{|l|c|c|c|c|c|c|c|c|c|c||c|c|}
\hline
\multirow{4}{*}{\textbf{Model}} & \multicolumn{2}{c|}{\textbf{Precision}} & \multicolumn{2}{c|}{\textbf{Recall}} & \multicolumn{2}{c|}{\textbf{AP50}} & \multicolumn{2}{c|}{{$\mathbf{F_1}$}} & \multicolumn{2}{c||}{{$\mathbf{F_2}$}} & \multicolumn{2}{c|}{\textbf{FPS}}\\ 
\cline{2-13}
& \textbf{\rotatebox[origin=c]{90}{$\;$Original$\;$}} & \textbf{\rotatebox[origin=c]{90}{+TCA}} 
& \textbf{\rotatebox[origin=c]{90}{Original}} & \textbf{\rotatebox[origin=c]{90}{+TCA}} 
& \textbf{\rotatebox[origin=c]{90}{Original}} & \textbf{\rotatebox[origin=c]{90}{+TCA}} 
& \textbf{\rotatebox[origin=c]{90}{Original}} & \textbf{\rotatebox[origin=c]{90}{+TCA}} 
& \textbf{\rotatebox[origin=c]{90}{Original}} & \textbf{\rotatebox[origin=c]{90}{+TCA}} 
& \textbf{\rotatebox[origin=c]{90}{Original}} & \textbf{\rotatebox[origin=c]{90}{+TCA}} \\ 
\hline
\hline
\textbf{Mask-RCNN} \cite{he2017mask} & $0.492$ & $0.538$ & $0.625$ & $0.651$ & $0.530$ & $0.556$ & $0.550$ & $0.589$ & $0.593$ & $0.625$ & $7.84$ & $6.73$\\ 
\hline
\textbf{Cascade Mask-RCNN} \cite{cai2018cascade} & $0.606$ & $0.613$ & $0.660$ & $0.686$ & $0.628$ & $0.639$ & $0.632$ & $0.647$ & $0.648$ & $0.670$ & $9.53$ & $7.94$ \\ 
\hline
\textbf{YOLO11n} \cite{Jocher_Ultralytics_YOLO_2023} & $0.713$ & $0.719$ & $0.558$ & $0.591$ & $0.650$ & $0.648$ & $0.626$ & $0.648$ & $0.583$ & $0.613$ & \textcolor{blue}{$128.20$} & \textcolor{blue}{$34.48$} \\ 
\hline
\textbf{YOLO11s} \cite{Jocher_Ultralytics_YOLO_2023} & $0.757$ & $0.752$ & $0.612$ & $0.647$ & $0.705$ & $0.723$ & $0.677$ & $0.696$ & $0.636$ & $0.666$ & $116.27$ & $33.78$ \\ 
\hline
\textbf{YOLO11m} \cite{Jocher_Ultralytics_YOLO_2023} & $0.739$ & $0.745$ & $0.624$ & $0.648$ & $0.707$ & $0.726$ & $0.677$ & $0.693$ & $0.644$ & $0.665$ & $76.93$ & $29.41$ \\ 
\hline
\textbf{YOLO11l} \cite{Jocher_Ultralytics_YOLO_2023} & \textcolor{blue}{$0.812$} & \textcolor{blue}{$0.819$} & $0.588$ & $0.613$ & $0.713$ & $0.729$ & $0.682$ & $0.701$ & $0.622$ & $0.646$ & $57.14$ & $25.98$\\ 
\hline
\textbf{YOLO11x} \cite{Jocher_Ultralytics_YOLO_2023} & $0.746$ & $0.742$ & $0.609$ & $0.647$ & $0.682$ & $0.703$ & $0.671$ & $0.691$ & $0.632$ & $0.664$ & $34.01$ & $19.84$ \\ 
\hline
\textbf{SparseInst R-50} \cite{Cheng2022SparseInst} & $0.520$ & $0.583$ & \textcolor{blue}{$0.782$} & \textcolor{blue}{$0.807$} & $0.703$ & $0.722$ & $0.644$ & $0.677$ & $0.710$ & $0.749$ & $29.73$ & $18.32$ \\ 
\hline
\textbf{SparseInst PVTv2} \cite{Cheng2022SparseInst} & $0.683$ & $0.712$ & $0.770$ & $0.798$ & \textcolor{blue}{$0.721$} & \textcolor{blue}{$0.751$} & \textcolor{blue}{$0.724$} & \textcolor{blue}{$0.753$} & \textcolor{blue}{$0.751$} & \textcolor{blue}{$0.780$} & $27.99$ & $17.64$ \\ 
\hline
\end{tabular}
\vspace{-0.2cm}
\caption{Performance comparison of different models on the test split, with and without TCA. The models are applied on video and the metrics are calculated by evaluating on manually annotated frames. The best result on each metric is highlighted in \textcolor{blue}{blue}.}
\label{tab:model_comparison}
\vspace{-0.2cm}
\end{table*}

\subsection{Environment and Parameters}

Model training was executed on a $24$ GB NVIDIA GeForce RTX $4090$ GPU, using YOLO11 from ultralytics v$8.3.29$ \cite{Jocher_Ultralytics_YOLO_2023}, Mask RCNN \cite{he2017mask}, Cascade Mask RCNN \cite{cai2018cascade} and SparseInst \cite{Cheng2022SparseInst} using Detectron2 \cite{wu2019detectron2} v$0.6$, Python $3.10.4$, PyTorch $1.12.1$ and CUDA $12.2$. All reported FPS are measured on RTX $12$ GB $3060$ GPU, on $1920\times1080$ videos. 

\subsection{Evaluation Metrics}

To assess the effectiveness of our models in segmenting rip currents, we employ several standard evaluation metrics, following other video instance segmentation benchmarks \cite{perazzi2016benchmarkDAVIS, Yang2019vis, lin2014microsoftCOCO}, including IoU, Mean Average Precision (mAP), and the \( F_\beta \) score, with an emphasis on the \( F_{2} \) variant.


To evaluate the model’s detection quality across varying confidence thresholds, we use Average Precision (AP), which is derived from the Precision-Recall curve. AP is calculated by ranking model predictions by their confidence scores and integrating over the curve:
\begin{equation}\label{eq:AP}
AP = \sum_{n} (\text{Recall}_n - \text{Recall}_{n-1}) \cdot \text{Precision}_n.
\end{equation}
Since our model detects a single class (rip currents), the mean Average Precision (mAP) is equivalent to the AP for that class. Combining IoU for spatial accuracy and AP for threshold-independent detection quality provides a robust evaluation of the segmentation models.

Finally, we utilize the \( F_\beta \) score, a weighted harmonic mean of Precision and Recall, to offer a balanced metric. In our experiments, we specifically focus on \( F_{2} \), where \( \beta = 2 \):
\begin{equation}\label{eq:fbeta}
    F_\beta = \frac{(1 + \beta^2) \cdot (\text{precision} \cdot \text{recall})}{\beta^2 \cdot \text{precision} + \text{recall}}.
\end{equation}
Emphasizing recall with \( F_{2} \) aligns with the safety-critical nature of rip current detection, as false negatives—missed detections—pose significant risks. In a beach monitoring system, a false positive may simply disturb beachgoers, while a false negative could result in a potentially life-threatening situation. Thus, prioritizing recall with \( F_{2} \) allows us to reduce missed detections, enhancing safety.

\subsection{Baseline Results (without TCA)}

The performance on both validation and test sets starts off modest, underscoring the room for improvement on this task (see Table \ref{tab:model_comparison}). Unlike other datasets where bounding box detection often yields strong results with little effort, our dataset reveals the tougher challenge of accurately segmenting rip currents compared to simply detecting them. Different from the results of Dumitriu~\etal\cite{dumitriu2023rip}, where YOLOv8 was used with reasonably good results, we show that the increase in diversity also results in increased difficulty (see Table \ref{tab:cross_dataset}).

We observe notable performance differences among the evaluated models. SparseInst with PVTv2 and augmentation achieves the highest $F2$ score and a strong balance of precision and recall, while maintaining high FPS. YOLO11 variants, particularly the large model, lead in precision but exhibit lower recall, with YOLO11-nano being the fastest.


\begin{figure*}
\centering
\setlength{\tabcolsep}{1pt}
\begin{tabular}{c c c c c}
     Original Image & Prediction & Prediction + TCA & Pred. + Filtered TCA  & Ground Truth \tabularnewline
    
     \includegraphics[width=0.195\textwidth]{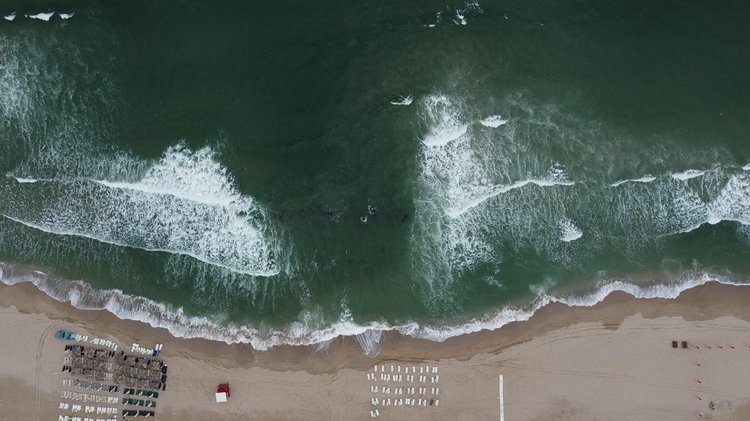} &
     \includegraphics[width=0.195\textwidth]{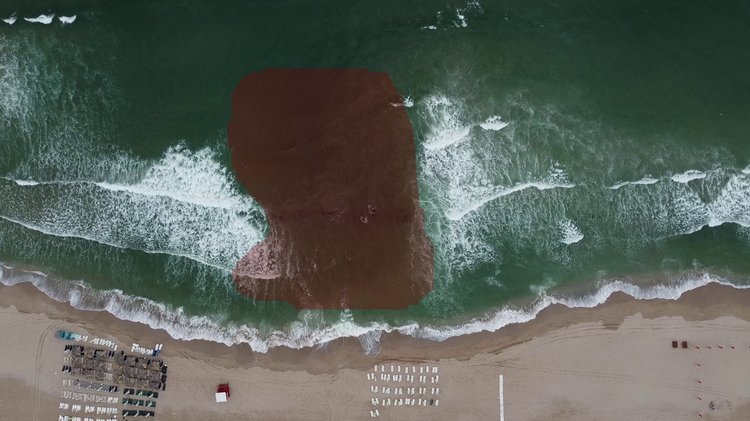} &
     \includegraphics[width=0.195\textwidth]{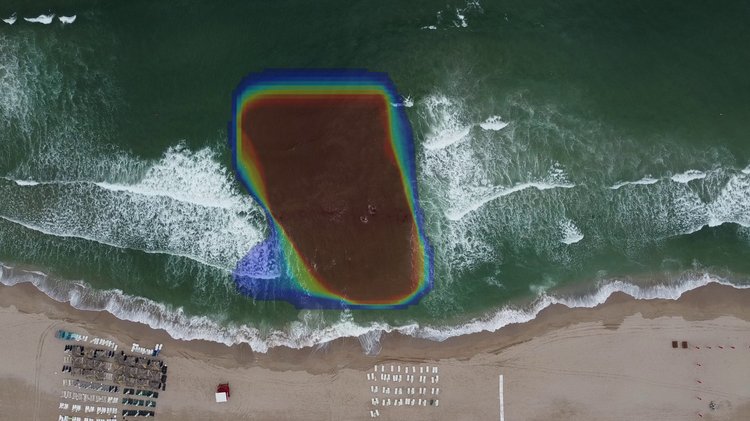} &
     \includegraphics[width=0.195\textwidth]{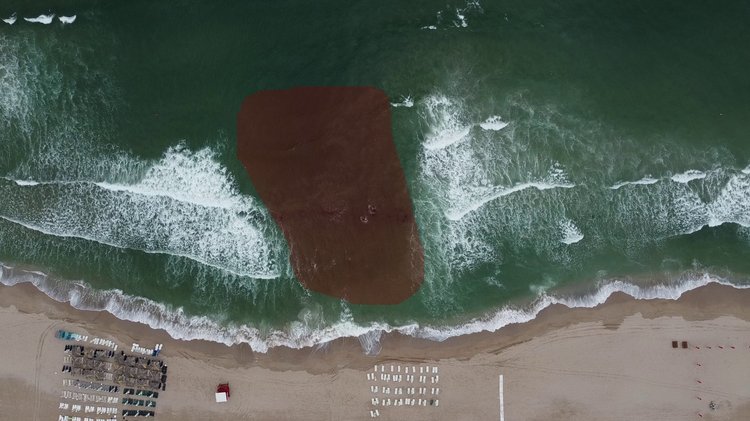} &
     \includegraphics[width=0.195\textwidth]{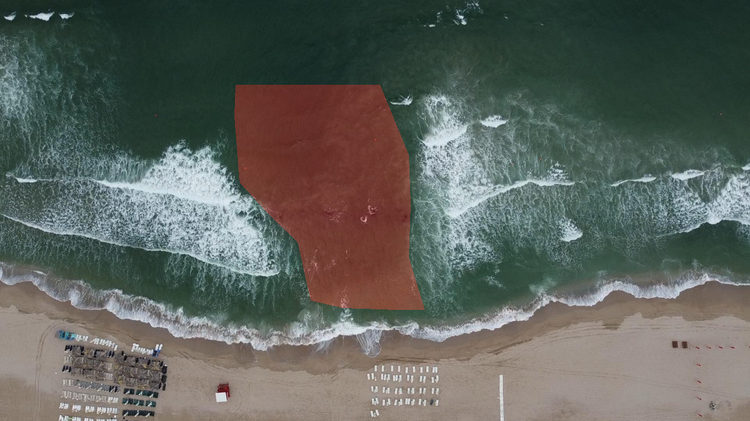} 
     \tabularnewline
     \includegraphics[width=0.195\textwidth]{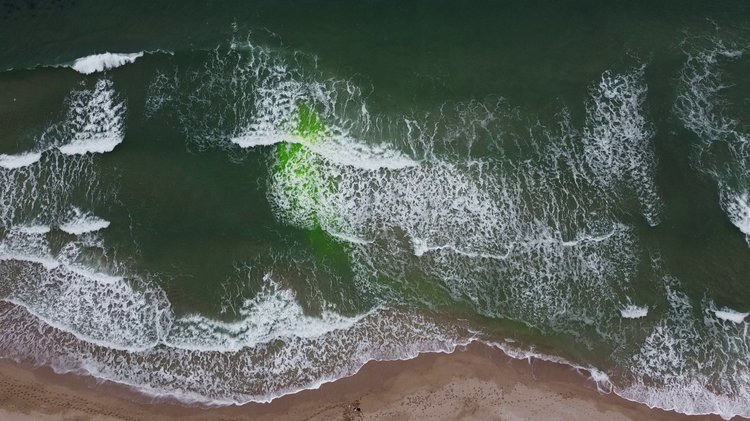} &
     \includegraphics[width=0.195\textwidth]{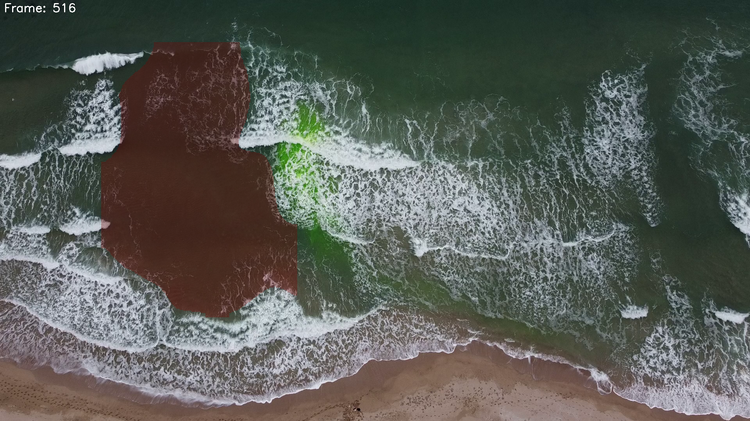} &
     \includegraphics[width=0.195\textwidth]{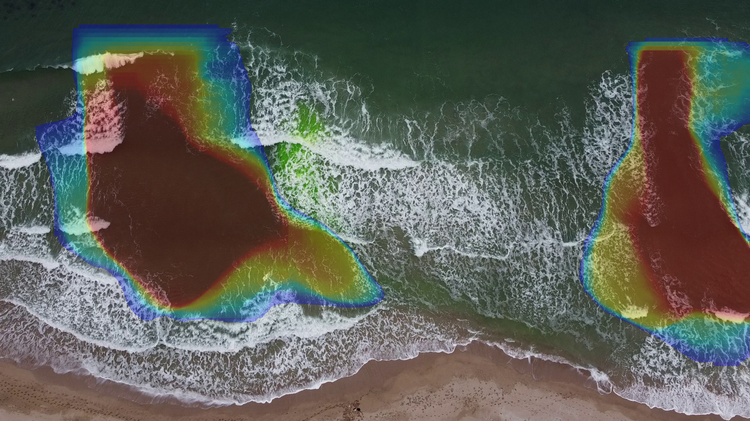} &
     \includegraphics[width=0.195\textwidth]{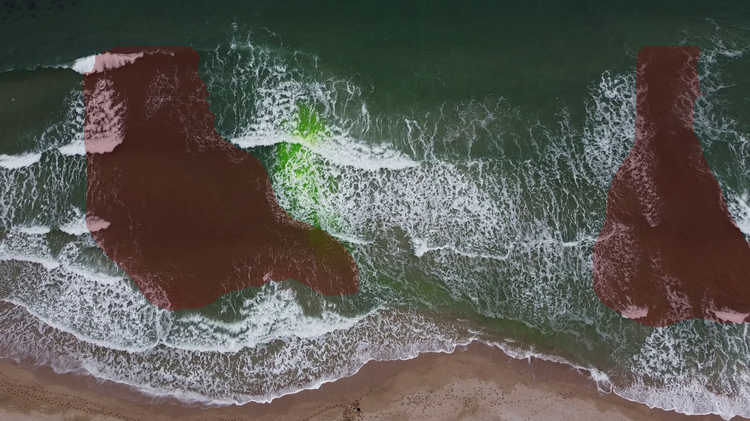} &
     \includegraphics[width=0.195\textwidth]{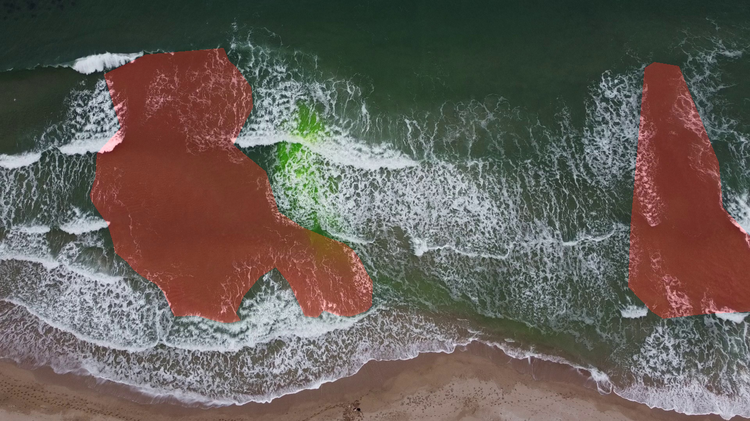} 
     \tabularnewline
     \includegraphics[width=0.195\textwidth]{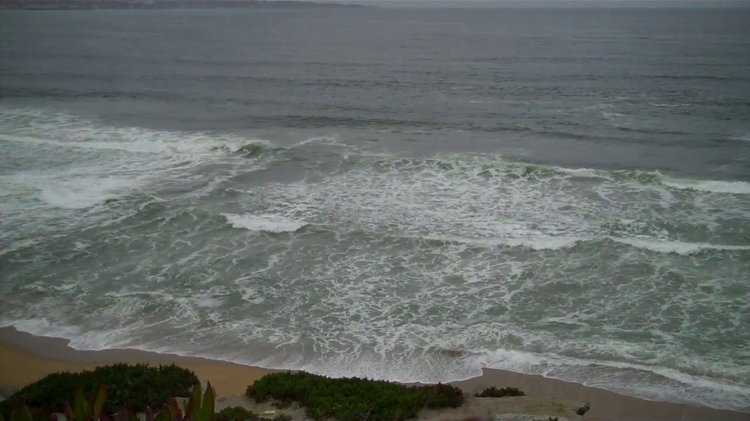} &
     \includegraphics[width=0.195\textwidth]{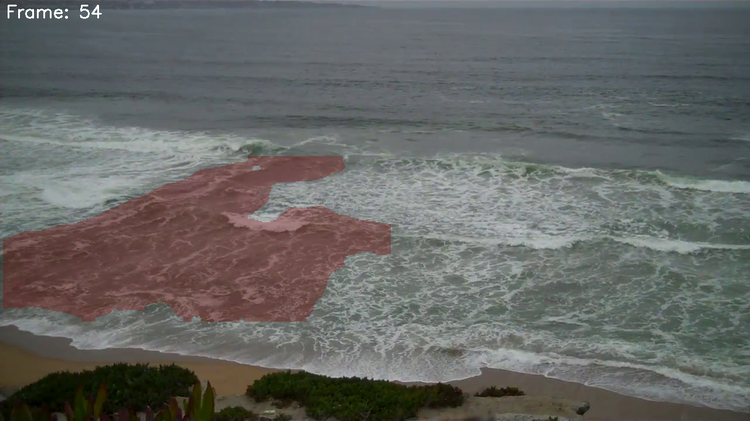} &
     \includegraphics[width=0.195\textwidth]{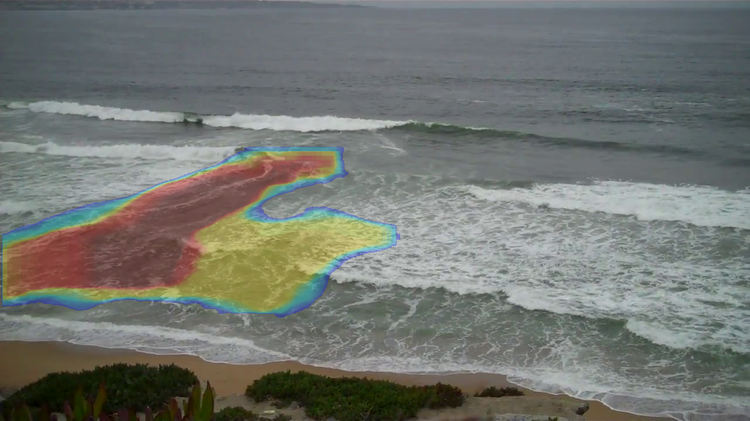} &
     \includegraphics[width=0.195\textwidth]{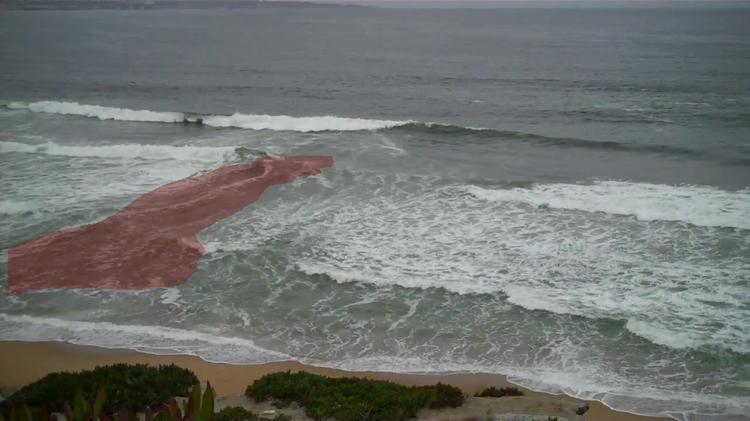} &
     \includegraphics[width=0.195\textwidth]{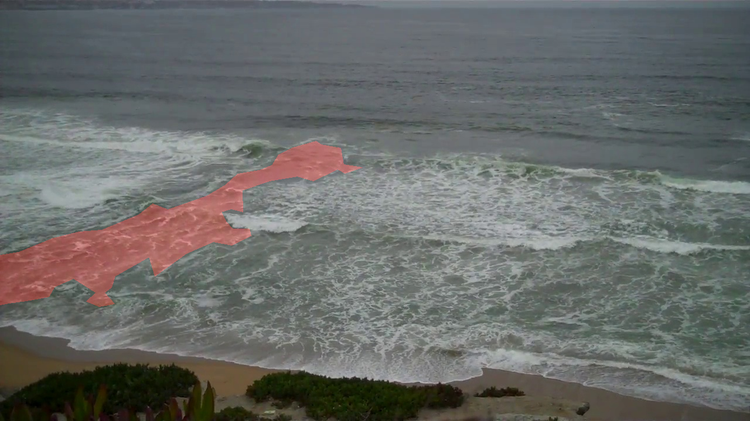} 
     \tabularnewline
     \includegraphics[width=0.195\textwidth]{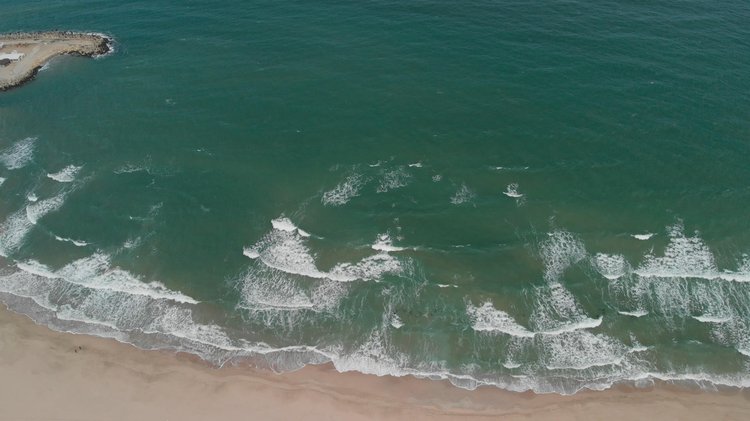} & 
     \includegraphics[width=0.195\textwidth]{figures/comparison/4/original_image_041.jpg} & 
     \includegraphics[width=0.195\textwidth]{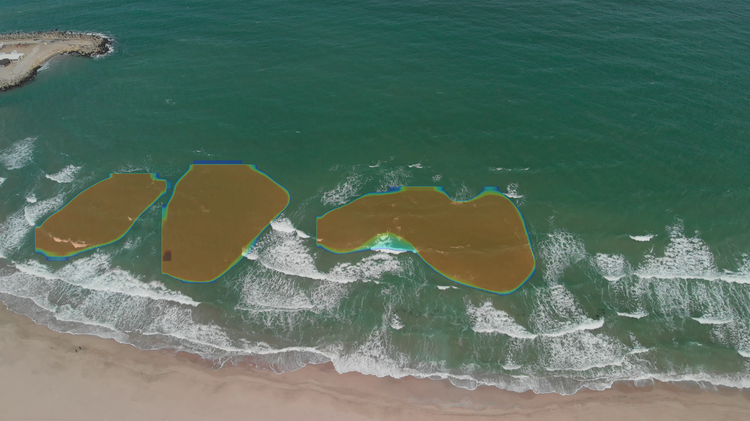} & 
     \includegraphics[width=0.195\textwidth]{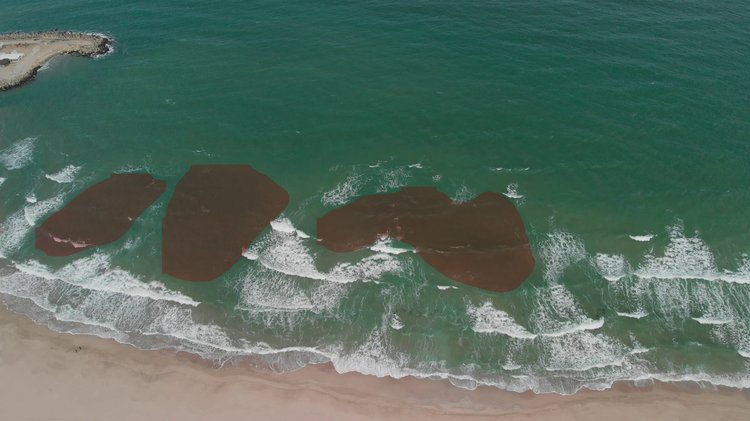} & 
     \includegraphics[width=0.195\textwidth]{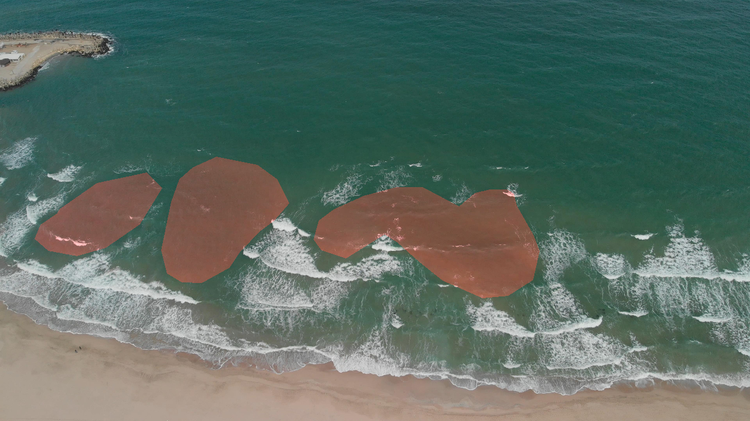} 
     \tabularnewline
     \includegraphics[width=0.195\textwidth]{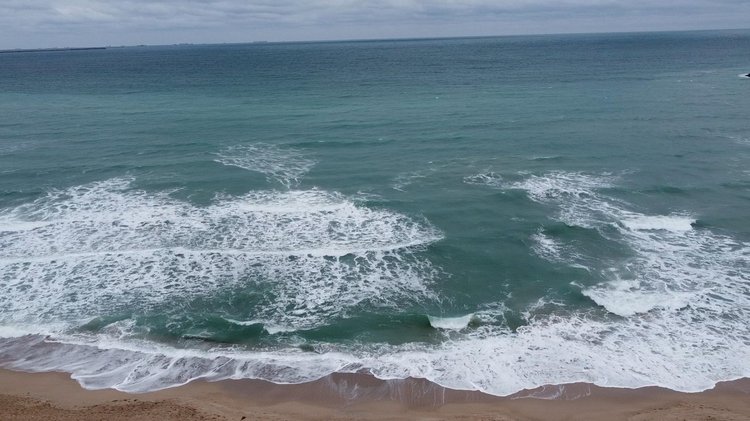} & 
     \includegraphics[width=0.195\textwidth]{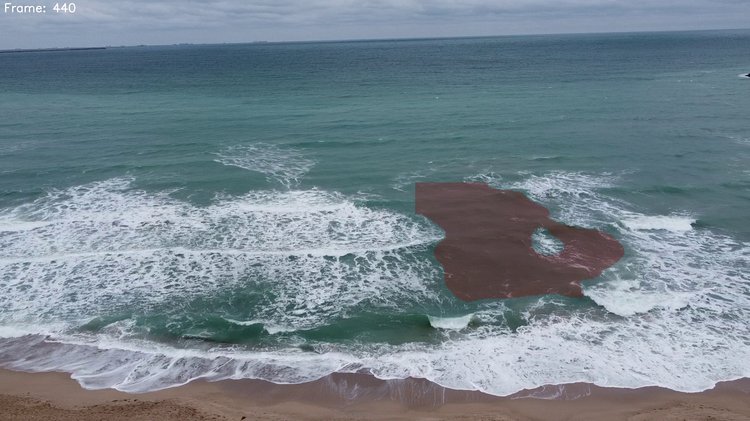} & 
     \includegraphics[width=0.195\textwidth]{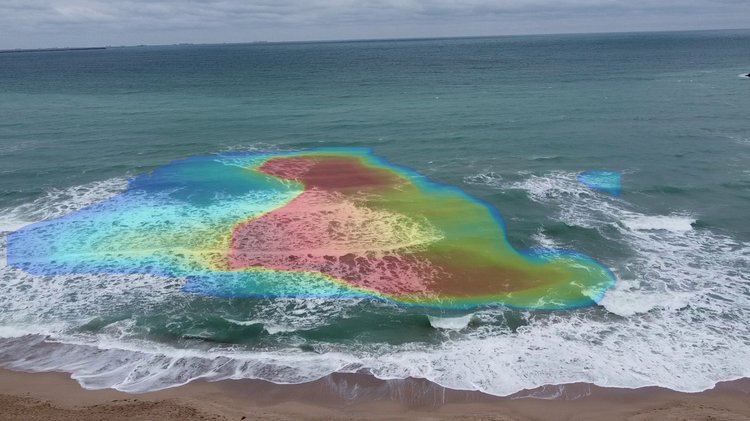} & 
     \includegraphics[width=0.195\textwidth]{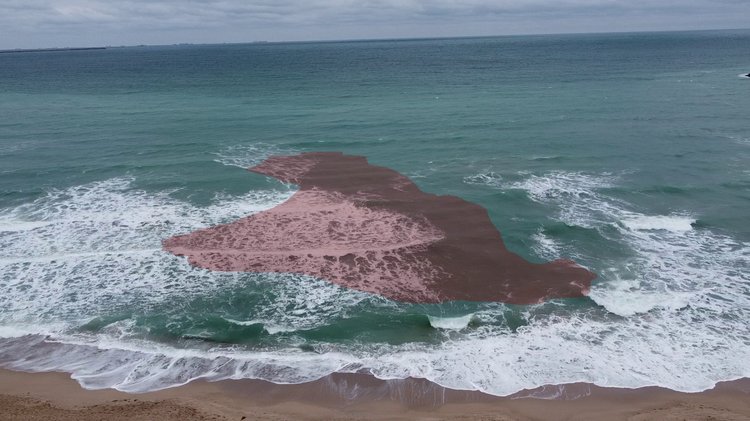} & 
     \includegraphics[width=0.195\textwidth]{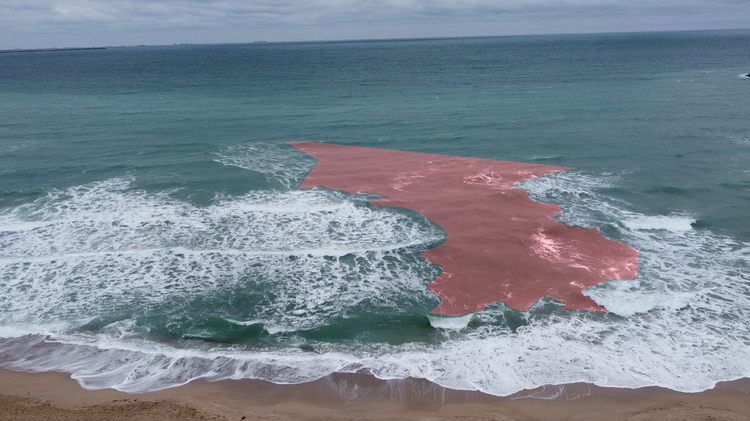} 
     \tabularnewline
\end{tabular}
\vspace{-0.3cm}
  \caption{Examples of rip current detection results across processing stages, with each row illustrating a distinct case for the impact of TCA:
1. TCA smooths the rip current shape on a successful detection.
2. TCA recovers false negatives on the right side.
3. TCA reduces false positives of an over-segmented mask to better match the ground truth.
4. TCA enables detection across frames with consecutive false negatives.
5. Failure case: TCA reduces detection accuracy due to initial stationary detection followed by rapid camera movement.
}
  \label{fig:testing_results}
  \vspace{-0.3cm}
\end{figure*}

\subsection{Results with TCA}
Applying TCA significantly improves segmentation stability across all models (see Figures \ref{fig:tca_good} and \ref{fig:testing_results} and Table \ref{tab:model_comparison}). TCA reduces false positives and enhances temporal consistency, especially in challenging, turbulent areas of water where rip currents appear intermittently. 

The $F_2$ score increases notably across models when TCA is applied, highlighting a reduction in false negatives, a critical improvement for safety-focused applications. In several cases of fast camera movement, TCA increases the number of false positives (see Figure \ref{fig:testing_results}). That number is negligible, as overall, the integration of TCA makes every model's output more reliable. This enables clearer and more stable segmentation of rip currents over video sequences, crucial for real-time beach monitoring.

However, different TCA implementations suit distinct scenarios. A slow-gain, slow-decay TCA excels with stationary video footage, but hampers performance on moving video, where a fast-gain, fast-decay approach is preferred, while a fast-gain, slow-decay TCA can be ideal for safety-critical environments requiring caution. Although TCA is highly effective in known contexts, optimizing it for a diverse dataset can be challenging without prior knowledge of the video type it will process.

\subsection{General Analysis}
One-stage detectors offer faster and more stable outputs, outperforming the two-stage methods. The results showcase the difficulty of the RipVIS dataset. While all methods perform well on standard benchmarks, and are still considered representative state-of-the-art models, they underperform on RipVIS. TCA enhanced segmentation stability across models, notably reducing false positives and false negatives, particularly for one-stage methods, making them more reliable for safety-critical applications. The diversity of our dataset introduces challenges not fully addressed by prior datasets, emphasizing the need for further improvements in segmentation models to generalize across various beach environments.

\subsection{Hyperparameter Tuning} 

All models were extensively trained on the dataset, exploring various hyperparameters to establish baseline comparisons. For YOLO, we trained all versions with both pre-trained and custom weights. We also tested most models from the \textit{Detectron2} model zoo, experimenting with different data augmentations, learning rates, schedulers, batch sizes, number of proposals, deformable convolutions, and various backbones. Additionally, the models were trained on an expanded version of the dataset, with automatically generated annotations for frames not manually annotated. While these annotations are reasonably accurate, they lack the precision of manual annotations, resulting in a slight drop in most metrics, likely due to overfitting on larger videos. We also experimented with various TCA methods, from linear to polynomial adjustments, with different upper bounds and additional thresholding for final prediction. A relevant trade-off is that different types of TCA are useful for moving cameras vs.~fixed cameras. Full details of the hyperparameter tuning and its impact are available in the supplementary material.

\section{Conclusion}

In this paper, we introduced RipVIS, a large-scale high-quality dataset specifically designed for rip current instance segmentation. RipVIS spans diverse environmental conditions, geographic locations, and video sources, making it the most comprehensive resource of its kind. By offering annotated videos with carefully curated training, validation, and test splits, RipVIS addresses the unique challenges posed by the amorphous and dynamic nature of rip currents, which traditional datasets and detection methods have struggled to overcome effectively.

Our analysis highlighted the limitations of popular instance segmentation models, including one-stage and two-stage detectors, on this challenging dataset. TCA demonstrated its effectiveness in improving segmentation consistency, particularly in difficult scenarios where rip currents appear intermittently. These results emphasize the need for more robust and accurate models to advance rip current detection—a safety-critical task where missed detections can lead to life-threatening consequences. The findings reinforce the importance of prioritizing recall and accuracy in such applications.
By releasing RipVIS and its benchmark website, we aim to foster a collaborative research environment within the global community. Openly sharing our data, code, and results not only supports innovation but also drives progress in the field of automatic rip current detection. Ultimately, we envision RipVIS as a pivotal resource for creating safer beaches and raising public awareness.

\section{Acknowledgments}
This work was partially supported by the Alexander von Humboldt Foundation and represents the culmination of a three-year collaboration between the University of W{\"u}rzburg and the University of Bucharest. We extend our sincere thanks to all the volunteer annotators, each of whom is credited individually in the video details sheet.

{
    \small
    \bibliographystyle{ieeenat_fullname}
    \bibliography{main}
}


   

\clearpage
\setcounter{page}{1}
\maketitlesupplementary

\section{Overview}

The supplementary material provides additional details and insights into the RipVIS dataset, experimental results, and methodology. While the main paper focuses on the major contributions and results, this document elaborates on the dataset's structure and diversity, the qualitative results of our experiments, and the impact of Temporal Confidence Aggregation (TCA) on rip current detection.

This supplementary aims to reinforce the robustness and reproducibility of our findings, offering a deeper understanding of the addressed challenges and proposed solutions. It also provides additional visualizations and metrics that could not be included in the main manuscript due to space limitations, including validation results (see Table \ref{tab:validation_results}).

RipVIS is a Video Instance Segmentation dataset, and it is challenging to convey its value 
 in a static format. The supplementary material starts with a short description of the dataset variety in Section \ref{sec:dataset}, with a visual showcase of all its diversity without masks, urging readers to see how many rip currents they can identify in Figure \ref{fig:dataset_diversity}, before looking at the ground truths in Figure \ref{fig:dataset_diversity_masked}.

\begin{figure*}
\centering
\setlength{\tabcolsep}{1pt}
\begin{tabular}{c c c c c}
     Original Image & Prediction & Prediction + TCA & Pred. + Filtered TCA  & Ground Truth \tabularnewline
    
     \vspace{-1mm} 
     \begin{turn}{90} {\raggedright Frame 158} \end{turn}
     \includegraphics[width=0.19\textwidth]{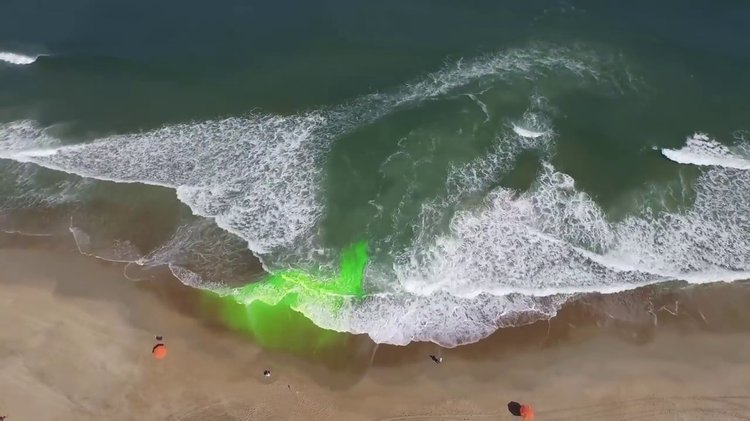} &
     \includegraphics[width=0.19\textwidth]{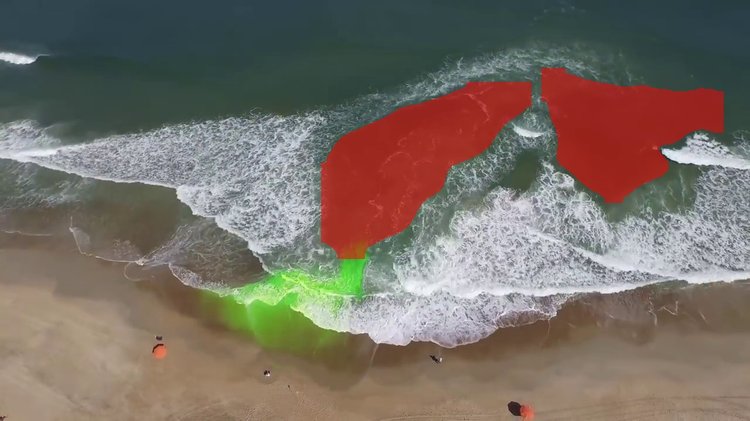} &
     \includegraphics[width=0.19\textwidth]{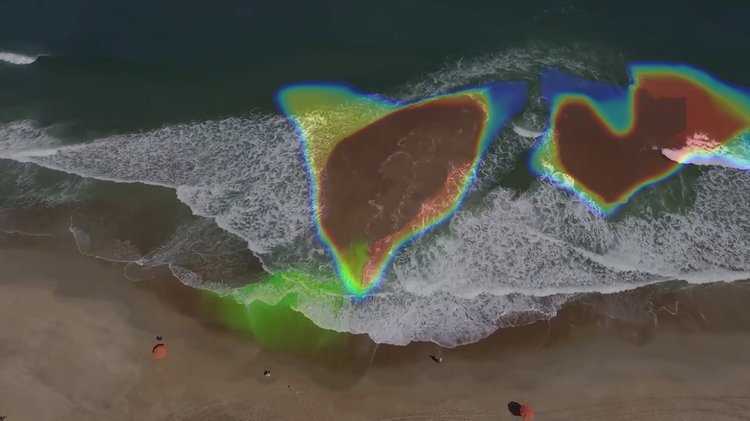} &
     \includegraphics[width=0.19\textwidth]{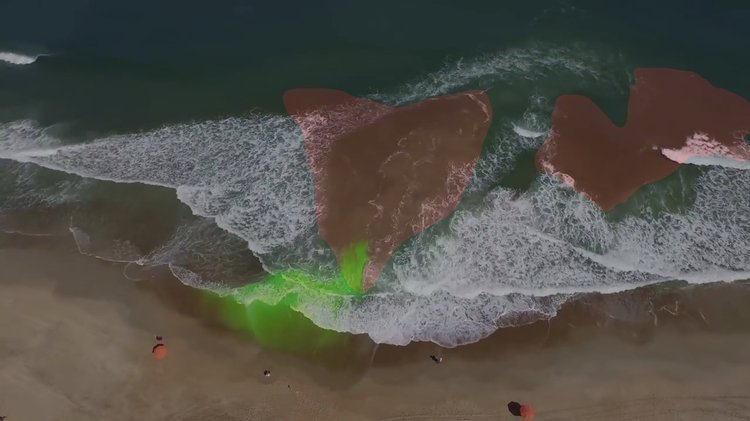} &
     \includegraphics[width=0.19\textwidth]{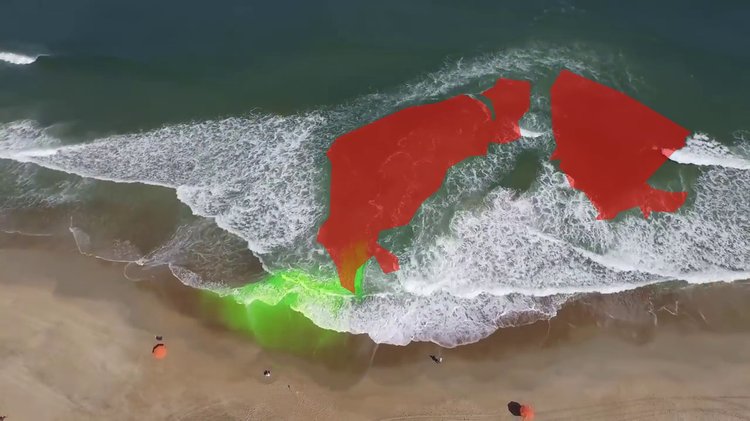} 
     \tabularnewline     
     \vspace{-1mm} 
     \begin{turn}{90} {\raggedright Frame 170} \end{turn}
     \includegraphics[width=0.19\textwidth]{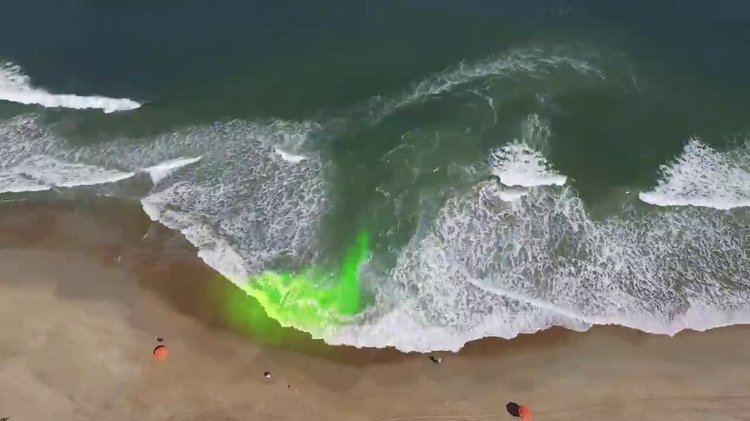} &
     \includegraphics[width=0.19\textwidth]{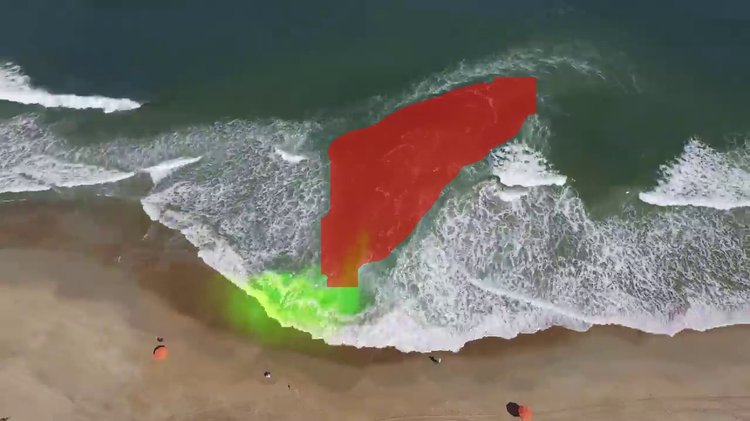} &
     \includegraphics[width=0.19\textwidth]{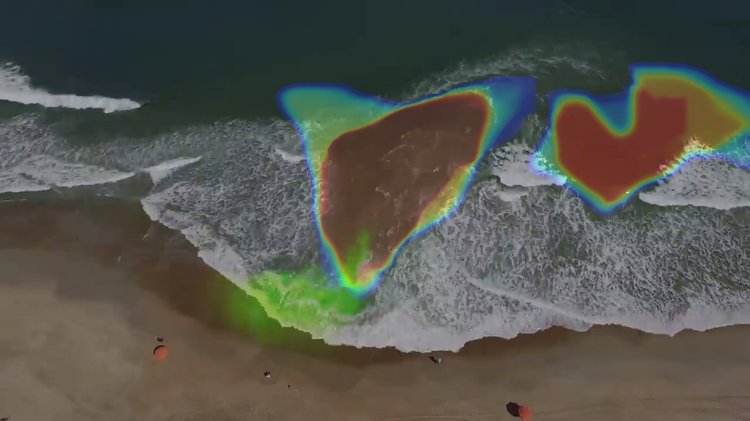} &
     \includegraphics[width=0.19\textwidth]{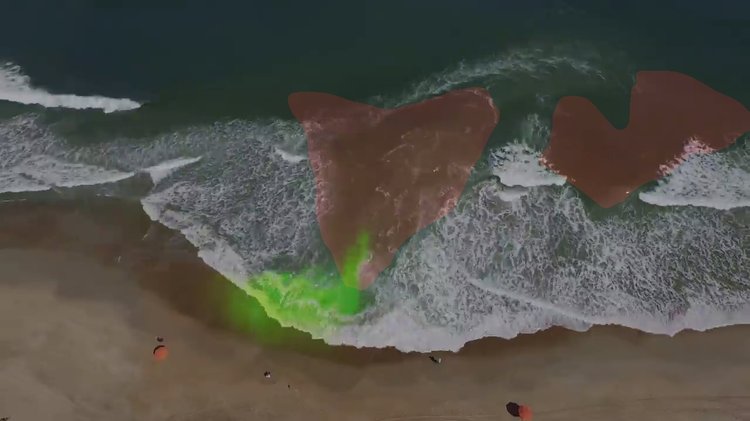} &
     \includegraphics[width=0.19\textwidth]{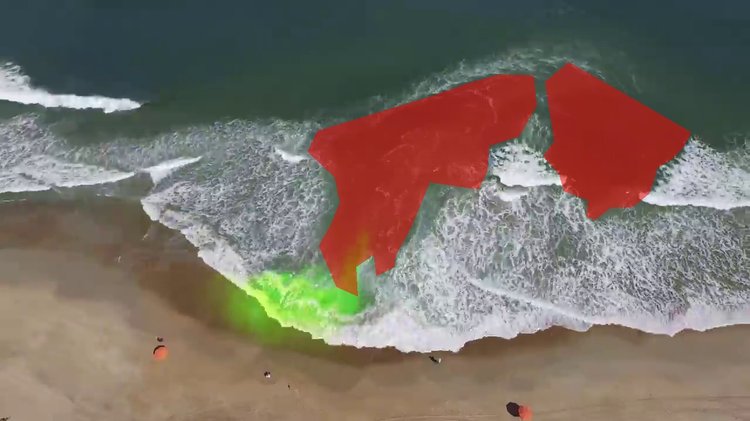} 
     \tabularnewline     
     \vspace{-1mm} 
     \begin{turn}{90} {\raggedright Frame 176} \end{turn}
     \includegraphics[width=0.19\textwidth]{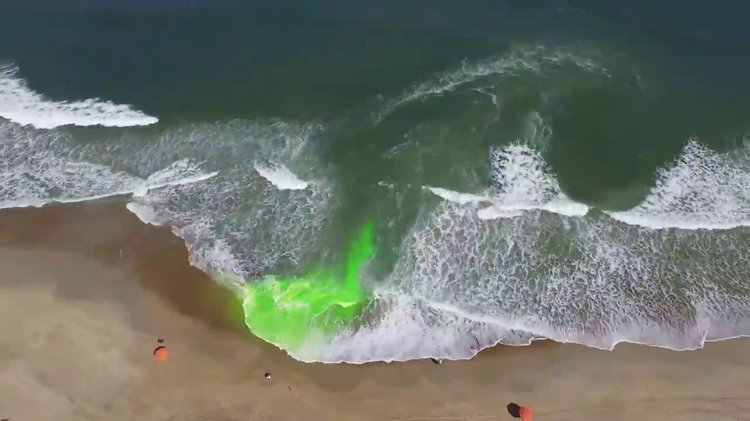} &
     \includegraphics[width=0.19\textwidth]{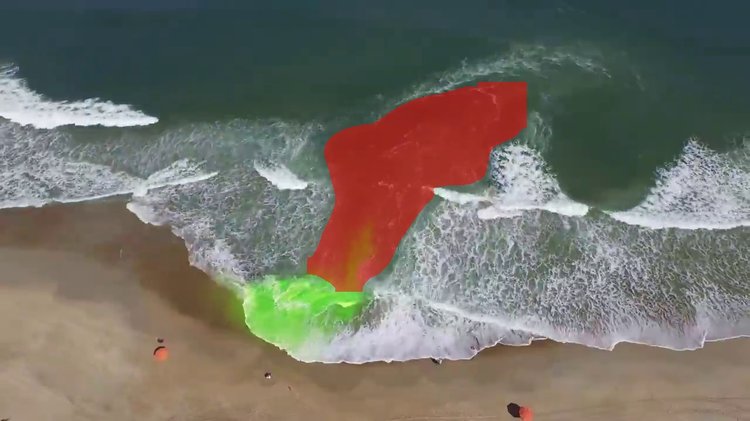} &
     \includegraphics[width=0.19\textwidth]{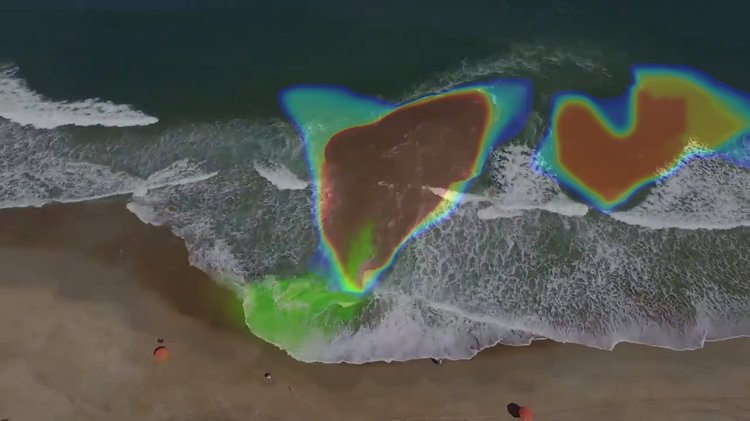} &
     \includegraphics[width=0.19\textwidth]{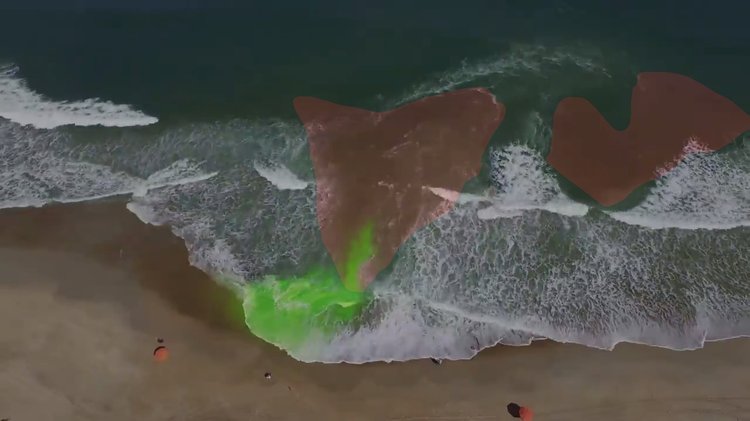} &
     \includegraphics[width=0.19\textwidth]{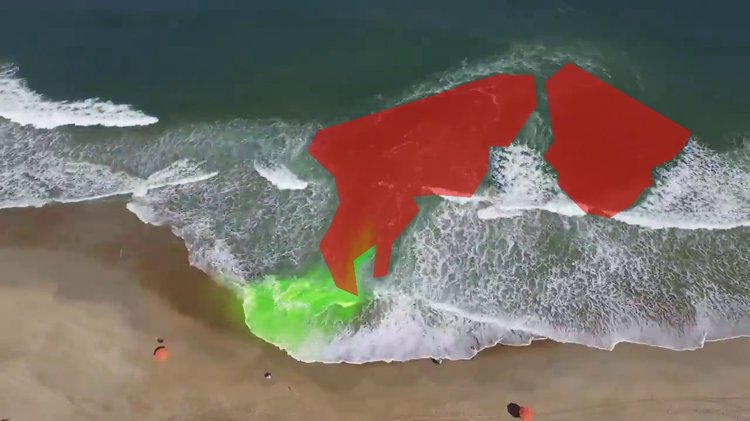} 
     \tabularnewline  
     \vspace{-1mm} 
     \begin{turn}{90} {\raggedright Frame 201} \end{turn}
     \includegraphics[width=0.19\textwidth]{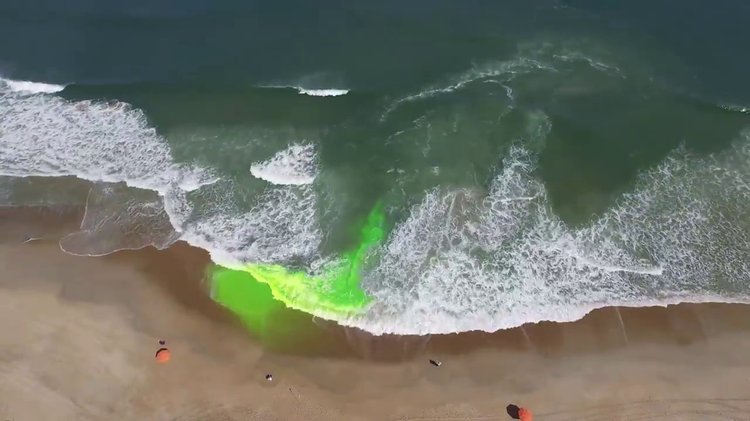} &
     \includegraphics[width=0.19\textwidth]{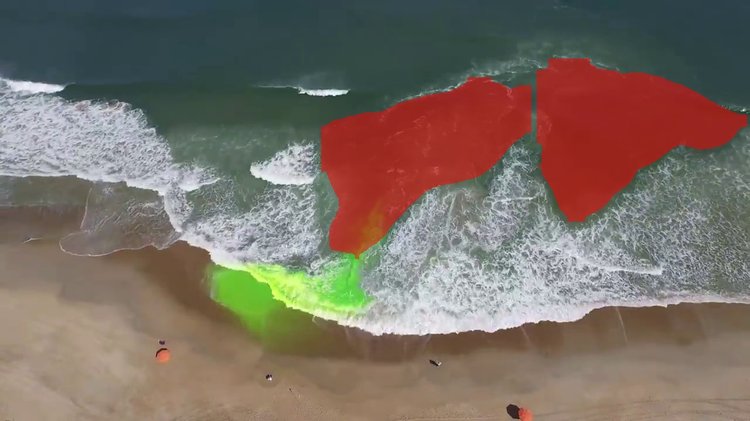} &
     \includegraphics[width=0.19\textwidth]{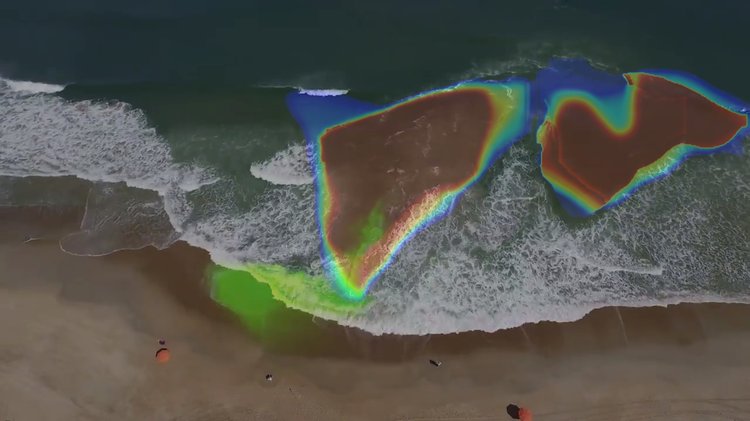} &
     \includegraphics[width=0.19\textwidth]{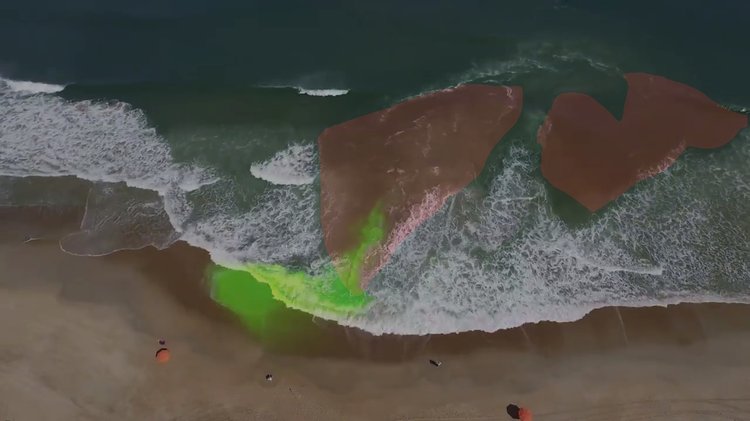} &
     \includegraphics[width=0.19\textwidth]{figures/qualitative_results/1/gt/frame_000176.jpg} 
     \tabularnewline
     \vspace{-1mm} 
     \begin{turn}{90} {\raggedright Frame 202} \end{turn}
     \includegraphics[width=0.19\textwidth]{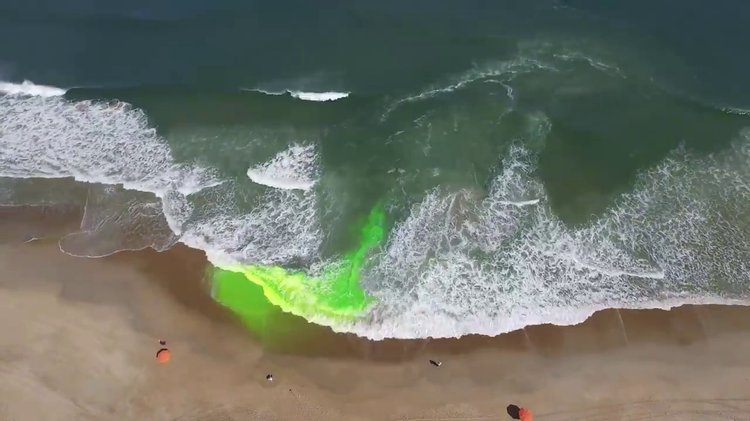} &
     \includegraphics[width=0.19\textwidth]{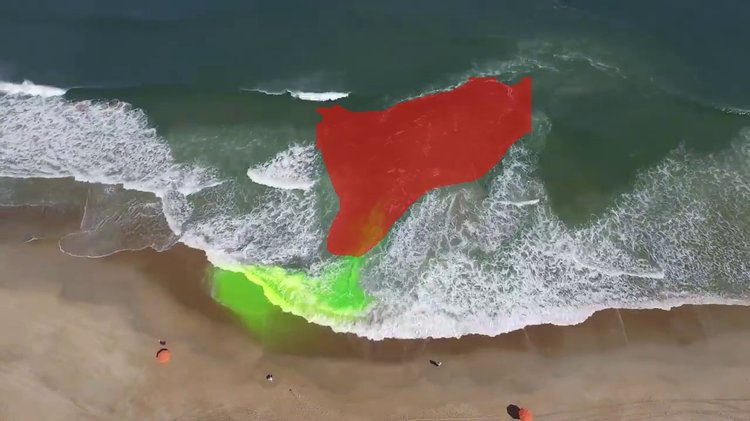} &
     \includegraphics[width=0.19\textwidth]{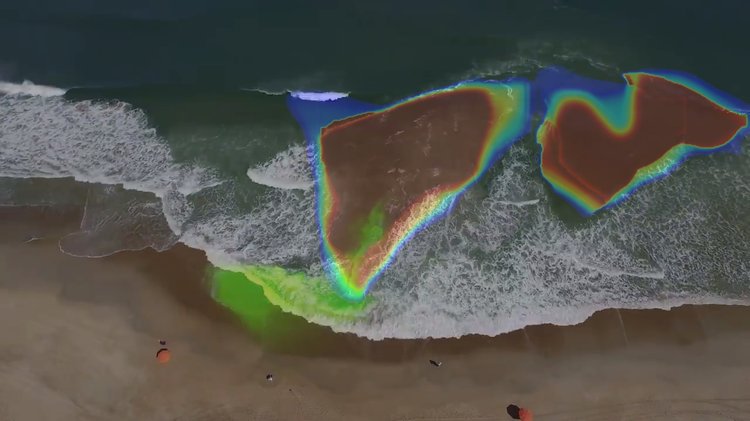} &
     \includegraphics[width=0.19\textwidth]{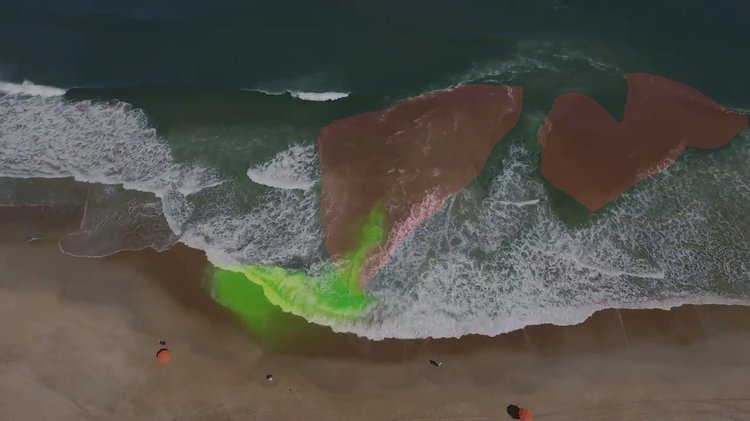} &
     \includegraphics[width=0.19\textwidth]{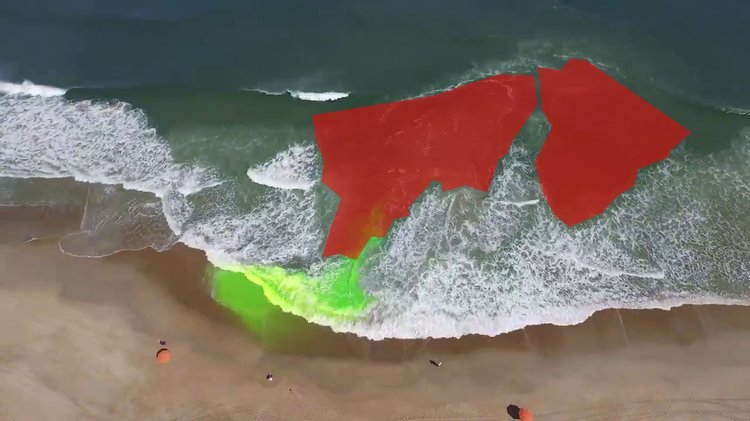} 
     \tabularnewline
\end{tabular}
  \caption{A more detailed example of TCA in action. All rows are of frames from the same video, showing how we mitigate for the false negative present in frames 176 (3rd row) and frame 202 (5th row). }
  \label{fig:tca_good}
\end{figure*}

We continue in Section \ref{sec:TCA} with a deep dive into TCA, as exemplified in Figure \ref{fig:tca_architecture}. We describe the approach, its implementation methodology, its improvements and limitations, as well as final results. We also showcase TCA in action in more detailed scenarios, by sampling more frames from the same video. In Figure \ref{fig:tca_good}, TCA can be seen filtering false negatives, while in Figure \ref{fig:tca_mixed}, it can be seen filtering false positives, with a strong success rate, albeit not $100\%$. Lastly, we provide Figure \ref{fig:tca_fail}, where TCA harms performance in a video transitioning from static to moving camera.

Finally, we finish with hyperparameter tuning in Section \ref{sec:Ablation}, diving deep into the hyperparameters that we used to train the different models, their strength, limitations and overall results. We analyze each model individually, discussing the approach used for hyperparameter tuning in each case. Ultimately, in Table \ref{tab:stddev_summary}, we present the standard deviations on all relevant metrics, for all models, on both validation and test sets.

\section{Dataset Variety}
\label{sec:dataset}
Rip currents are complex, dynamic phenomena, requiring datasets that reflect their diversity in form, environment, and conditions. The RipVIS dataset was designed to capture this variety comprehensively, spanning different geographic locations, camera perspectives, and environmental scenarios.

The dataset consists of 184 videos, totaling 212,328 frames. The videos are taken from multiple orientations and elevations, with different types of rip currents, in various weather conditions, from both seas and oceans. Figure \ref{fig:dataset_diversity} contains a large sampling from the videos, showcasing this variety, with Figure \ref{fig:dataset_diversity_masked} showcasing their annotation masks. RipVIS videos are mainly in landscape orientation, with a few in portrait, reflecting real-world diversity in camera setups. For a detailed breakdown of the resolution and FPS distribution of RipVIS videos, see Table \ref{tab:ripvis_video_stats}.

\begin{table}[t]
\centering
\small 
\resizebox{\columnwidth}{!}{
\begin{tabular}{|l|r||l|r|}
\hline
\textbf{Resolution} & \textbf{\#Videos} & \textbf{FPS} & \textbf{\#Videos} \\ 
\hline
\hline
$4,096 \times 2,160$ & 1 & 60 & 14 \\
$3,840 \times 2,160$ & 24 & 50 & 1 \\
$2,730 \times 1,440$ & 1 & 30 & 119 \\
$2,720 \times 1,530$ & 6 & 25 & 8 \\
$2,560 \times 1,440$ & 2 & 24 & 8 \\
$2,160 \times 3,840$ & 1 & & \\
$1,920 \times 1,080$ & 53 & & \\
$1,280 \times 720$ & 52 & & \\
$1,280 \times 676$ & 2 & & \\
$1,080 \times 1,920$ & 2 & & \\
$720 \times 1,280$ & 1 & & \\
$480 \times 360$ & 3 & & \\
$360 \times 640$ & 2 & & \\ \hline
\textbf{Total} & \textbf{150} & \textbf{Total} & \textbf{150} \\ \hline
\end{tabular}
}
\caption{Resolution and FPS distribution of the 150 RipVIS videos containing rip currents, sorted by decreasing resolution and FPS. Videos are primarily landscape-oriented, with a few in portrait, reflecting real-world camera diversity. This variation enables robust evaluation across video qualities.}
\label{tab:ripvis_video_stats}
\end{table}

\section{Temporal Confidence Aggregation (TCA)}
\label{sec:TCA}
TCA is an approach that enhances the consistency and reliability of rip current segmentation in video data by leveraging temporal information across consecutive frames. TCA effectively accumulates segmentation confidence over time, generating heatmaps that emphasize regions with stable rip current detections, while reducing noise from sporadic or transient detections.

\subsection{Methodology}
The TCA approach consists of several components that work together to aggregate segmentation confidence over time. Each component plays a role in dealing with the fluctuating and complex patterns of rip currents.

\noindent
\textbf{Heatmap initialization.} For each instance, a heatmap is initialized as a two-dimensional array, where each value represents the accumulated segmentation confidence for a corresponding pixel in the video frame. This heatmap captures areas of high and consistent rip current activity, ensuring that these remain prominent throughout the analysis.

\begin{table}[t]
\centering
\renewcommand{\arraystretch}{1.1}
\small 
\setlength{\tabcolsep}{2pt} 
\begin{tabular}{|l|c|c|c|c|c|c|c|c|c|c|}
\hline
{\textbf{Model}} & \rotatebox[origin=c]{90}{\textbf{$\;$Precision$\;$}} & \rotatebox[origin=c]{90}{\textbf{Recall}} & \rotatebox[origin=c]{90}{\textbf{AP50}} & \rotatebox[origin=c]{90}{$\mathbf{F_1}$} & \rotatebox[origin=c]{90}{$\mathbf{F_2}$}\\ 
\hline
\hline
\textbf{Mask-RCNN} \cite{he2017mask} & $0.415$ & $0.615$ & $0.550$ & $0.496$ & $0.561$ \\ 
\hline
\textbf{Cascade Mask-RCNN} \cite{cai2018cascade} & $0.550$ & $0.531$ & $0.548$ & $0.540$ & $0.535$ \\ 
\hline
\textbf{YOLO11n} \cite{Jocher_Ultralytics_YOLO_2023} & $0.679$ & $0.492$ & $0.610$ & $0.571$ & $0.521$ \\ 
\hline
\textbf{YOLO11s} \cite{Jocher_Ultralytics_YOLO_2023} & $0.670$ & $0.514$ & $0.596$ & $0.582$ & $0.534$ \\ 
\hline
\textbf{YOLO11m} \cite{Jocher_Ultralytics_YOLO_2023} & $0.679$ & $0.543$ & $0.630$ & $0.603$ & $0.566$  \\ 
\hline
\textbf{YOLO11l} \cite{Jocher_Ultralytics_YOLO_2023} & \textcolor{blue}{$0.729$} & $0.521$ & $0.619$ & $0.608$ & $0.553$ \\ 
\hline
\textbf{YOLO11x} \cite{Jocher_Ultralytics_YOLO_2023} & $0.612$ & $0.628$ & \textcolor{blue}{$0.649$} & \textcolor{blue}{$0.620$} & \textcolor{blue}{$0.625$} \\ 
\hline
\textbf{SparseInst R-50} \cite{Cheng2022SparseInst} & $0.477$ & \textcolor{blue}{$0.664$} & $0.564$ & $0.555$ & $0.615$ \\ 
\hline
\textbf{SparseInst PVTv2} \cite{Cheng2022SparseInst} & $0.606$ & $0.615$ & $0.617$ & $0.610$ & $0.613$ \\ 
\hline

\end{tabular}
\caption{Performance comparison of different models on the validation split. The models are applied on video and the metrics are calculated by evaluating on manually annotated frames. The best result on each metric is highlighted in \textcolor{blue}{blue}.}
\label{tab:validation_results}
\end{table}

\noindent
\textbf{Heatmap update.} The core of TCA lies in updating the heatmap over time by leveraging the current segmentation mask and information from previous frames. The confidence scores for each pixel are averaged across a short temporal window using the formula:
\[ C_{avg}(t) = \alpha \cdot C(t) + (1 - \alpha) \cdot C_{avg}(t-1), \]
where \(C(t)\) is the confidence score at time \(t\), \(C_{avg}(t)\) is the aggregated confidence score, and \(\alpha\) is the decay factor, set between $0$ and $1$, which dictates the influence of the current frame's confidence on the moving average. This step boosts the scores of consistently detected pixels. Additionally, every instance associated with a heatmap is accompanied by two supporting arrays:
\begin{itemize}
\item \textbf{Present counter:} 
This pixel-wise counter tracks the cumulative number of detections for each pixel within an instance’s mask. Upon a detection, the counter increments for corresponding pixels, and growth is triggered only when the counter reaches a minimum threshold. This delay ensures that transient or spurious detections do not prematurely inflate heatmap values.

\item \textbf{Absence counter:} 
In contrast, this counter tracks the consecutive frames without a detection for each pixel. In the absence of a detection, the counter increases, triggering a reduction of heatmap values by a decay factor.

\end{itemize}


The heatmap update process is implemented using vectorized GPU operations, allowing efficient processing even for high-resolution video frames.

\noindent
\textbf{Heatmap smoothing.} Rip currents often have amorphous shapes that change rapidly across frames. To maintain stability, while accommodating their fluid nature, a Gaussian smoothing filter is applied to the heatmap.

\noindent
\textbf{Hysteresis thresholding.}
TCA employs hysteresis thresholding to derive final binary masks from accumulated heatmaps, operating on the principle of differentiating strong and weak confidence scores within the heatmap. It uses an upper and a lower threshold. Pixels above the upper threshold are marked as strong detections, while those between the lower threshold and the upper thresholds form a weak detection. To connect these pixels, TCA applies a morphological dilation operation to each strong region, slightly expanding it to overlap with the weak mask. The final segmentation mask comprises strong pixels alongside weak pixels that are spatially connected to them.

\noindent
\textbf{Instance tracking.} For each new frame, TCA tracks instances by matching them to IDs assigned in earlier frames.



\subsection{Results and Discussion}
The output of TCA is a heatmap that provides a confidence-weighted visualization of rip current segmentation over time. This aggregated heatmap is particularly beneficial for applications such as:
\begin{itemize}
    \item \textbf{Rip current tracking:} Providing a stable representation of rip current activity, even when individual segmentations are noisy or inconsistent.
    \item \textbf{Beach safety monitoring:} Emphasizing regions of high rip current activity, which can help in developing early warning systems to alert beachgoers and lifeguards.
\end{itemize}

By aggregating temporal information, TCA effectively reduces the impact of sporadic false positives and false negatives, ensuring that only regions with consistent rip current activity are highlighted, making it a robust approach for rip current segmentation.

\subsection{Limitations}
While TCA provides significant improvements in the consistency of rip current segmentation, there are several limitations:

\begin{itemize}
    \item \textbf{Increased computational requirements:} TCA requires maintaining and updating a heatmap in real-time, which can be computationally demanding, particularly for high-resolution video. Although GPU acceleration helps, substantial computational resources are still required.
    \item \textbf{Latency in highlighting rip currents:} Due to the need for multiple consistent segmentations before increasing confidence, TCA introduces some latency in highlighting newly detected rip currents. This can be a drawback for short videos or fast changing camera movement.
    \item \textbf{Parameter sensitivity:} The success of TCA hinges on well-adjusted parameters and thresholds. Consequently, although TCA can boost performance in tailored setups, achieving this becomes progressively more difficult as the setup broadens in scope.
\end{itemize}

\begin{table*}[t!]
\centering
\renewcommand{\arraystretch}{1.1}
\small 
\setlength{\tabcolsep}{4pt} 
\begin{tabular}{|l|c|c|c|c|c||c|c|c|c|c|}
\hline
\multirow{4}{*}{\textbf{Model}} & \multicolumn{5}{c||}{\textbf{Validation Stddev}} & \multicolumn{5}{c|}{\textbf{Test Stddev}} \\ 
\cline{2-11}
& \textbf{\rotatebox[origin=c]{90}{\ Precision\ }} & \textbf{\rotatebox[origin=c]{90}{Recall}} & \textbf{\rotatebox[origin=c]{90}{AP50}} & \rotatebox[origin=c]{90}{$\mathbf{F_1}$} & \rotatebox[origin=c]{90}{$\mathbf{F_2}$} & \textbf{\rotatebox[origin=c]{90}{Precision}} & \textbf{\rotatebox[origin=c]{90}{Recall}} & \textbf{\rotatebox[origin=c]{90}{AP50}}  & {\rotatebox[origin=c]{90}{$\mathbf{F_1}$}} & {\rotatebox[origin=c]{90}{$\mathbf{F_2}$}} \\ 
\hline
\hline
\textbf{Mask-RCNN} \cite{he2017mask} & $0.06$ & $0.09$ & $0.07$ & $0.07$ & $0.08$ & $0.05$ & $0.08$ & $0.07$ & $0.06$ & $0.07$ \\ 
\hline
\textbf{Cascade Mask-RCNN} \cite{cai2018cascade} & $0.05$ & $0.08$ & $0.07$ & $0.06$ & $0.07$ & $0.06$ & $0.07$ & $0.06$ & $0.06$ & $0.07$ \\ 
\hline
\textbf{YOLO11n} \cite{Jocher_Ultralytics_YOLO_2023} & $0.03$ & $0.03$ & $0.03$  & $0.03$ & $0.03$ & $0.04$ & $0.04$ & $0.03$ & $0.03$ & $0.04$ \\ 
\hline
\textbf{YOLO11s} \cite{Jocher_Ultralytics_YOLO_2023} & $0.03$ & $0.03$ & $0.03$ & $0.03$ & $0.03$ & $0.02$ & $0.04$ & $0.03$ & $0.03$ & $0.04$ \\ 
\hline
\textbf{YOLO11m} \cite{Jocher_Ultralytics_YOLO_2023} & $0.04$ & $0.03$ & $0.03$ & $0.04$ & $0.03$ & $0.05$ & $0.04$ & $0.03$ & $0.03$ & $0.04$ \\ 
\hline
\textbf{YOLO11l} \cite{Jocher_Ultralytics_YOLO_2023} & $0.06$ & $0.04$ & $0.04$ & $0.03$ & $0.04$ & $0.04$ & $0.03$ & $0.03$ & $0.04$ & $0.03$ \\ 
\hline
\textbf{YOLO11x} \cite{Jocher_Ultralytics_YOLO_2023} & $0.04$ & $0.04$ & $0.03$ & $0.04$ & $0.04$ & $0.05$ & $0.04$ & $0.04$ & $0.04$ & $0.04$ \\ 

\hline
\textbf{SparseInst} \cite{Cheng2022SparseInst} & $0.04$ & $0.04$ & $0.03$ & $0.04$ & $0.04$ & $0.09$ & $0.01$ & $0.05$ & $0.06$ & $0.03$ \\ 
\hline
\end{tabular}
\caption{Standard deviation summary for all models evaluated on the RipVIS dataset, with varied results across validation and test splits based on experiments.}
\label{tab:stddev_summary}
\end{table*}
\section{Hyperparameter Tuning} 
\label{sec:Ablation}

This section provides an extended analysis of our experimental results, focusing on model performance on the RipVIS dataset and insights from hyperparameter tuning studies. The experiments are aimed to assess popular instance segmentation models for rip current detection and evaluate key hyperparameter impacts.

Most experiments are focused on varying backbones, optimizers, schedulers, and learning rates, as these hyperparameters greatly affect a model's ability to generalize and detect complex rip current patterns. Other parameters, like training epochs, early stopping patience, and batch size, were tested but showed minimal impact. To further enhance robustness, we extensively tested image augmentations for models implemented in Detectron2 (all except YOLO11, for which we used the built-in ones), exploring their effect on performance under diverse conditions.

In the following subsections, we provide a detailed description of the employed models, their configurations, and the conducted experiments. Each model was extensively evaluated under varying settings to identify the optimal configurations, understand their strengths and limitations, and assess their suitability for the challenging task of rip current segmentation in diverse video settings.

\noindent
\textbf{Mask R-CNN:} Mask R-CNN \cite{he2017mask}, a two-stage model, extends Faster R-CNN with a segmentation branch, enabling simultaneous object detection and pixel-level masking. Using a Region Proposal Network (RPN) to generate Regions of Interest (RoIs), it excels at capturing irregular shapes like rip currents but sacrifices speed due to its complexity. In our tests, its performance was hampered by the dynamic nature of rip currents. For our experiments, we conducted an extensive study focusing primarily on different backbones, as these are critical for feature extraction. The backbones included ResNet-50-FPN \cite{he2016deep}, ResNet-101-FPN, ResNet-50-DC, and ResNet-101-DC, with FPN (Feature Pyramid Networks) enabling multi-scale feature extraction. Dilated Convolutions (DC), applied to specific stages of the backbone, expand the receptive field in these layers, enhancing spatial context capture for dense prediction tasks. In the experiments, we tested learning rates of 0.0025 and 0.005 with the SGD optimizer and the Warmup Multi-Step LR scheduler.



\noindent
\textbf{Cascade Mask R-CNN:} Cascade Mask R-CNN \cite{cai2018cascade} builds on the Mask R-CNN architecture by introducing a multi-stage cascade of detectors and mask predictors, where each subsequent stage is trained to refine the outputs from the previous one with progressively stricter IoU thresholds. This cascading refinement process can enhance detection and segmentation accuracy, particularly for objects with complex or occluded boundaries. In principle, this approach is beneficial for segmenting ambiguous boundaries, such as those seen in rip currents, which often exhibit irregular and shifting patterns. While the multi-stage architecture helps mitigate false positives and improve instance mask quality, it does increase computational overhead. In practice, however, the model's performance on rip currents was limited, indicating potential challenges in handling highly amorphous and dynamic shapes.

Similar to Mask R-CNN, we conducted experiments focusing on backbone variations, using ResNet-50-FPN, ResNet-101-FPN, ResNet-50-DC, and ResNet-101-DC. Learning rates of 0.0025 and 0.005 were tested, alongside the SGD optimizer and Warmup Multi-Step LR scheduler. 


\begin{table}[t]
\centering
\small 
\setlength{\tabcolsep}{1.2pt} 
\begin{tabular}{|l|l|c|}
\hline
\textbf{Model} & \textbf{Train$\rightarrow$Test} & \textbf{Accuracy} \\
\hline
\hline
\multirow{2}{*}{\textbf{YOLO8n}} & Dumitriu \etal \cite{dumitriu2023rip}$\rightarrow$Dumitriu \etal \cite{dumitriu2023rip} & {\color{BrickRed}$0.750$} \\
& Dumitriu \etal~\cite{dumitriu2023rip}$\rightarrow$RipVIS & {\color{Blue}$0.205$} \\
 \hline
\multirow{2}{*}{\textbf{YOLO11n}} & RipVIS$\rightarrow$RipVIS & {\color{Blue}$0.530$} \\
& RipVIS$\rightarrow$Dumitriu \etal \cite{dumitriu2023rip} & {\color{BrickRed}$0.803$} \\ 
\hline
\end{tabular}
\vspace{-0.3cm}
\caption{Cross-dataset experiments on RipVIS vs.~Dumitriu \etal \cite{dumitriu2023rip} dataset.}
\label{tab:cross_dataset}
\vspace{-0.3cm}
\end{table}

\noindent
\textbf{YOLO11:} In our experiments, YOLO11 \cite{Jocher_Ultralytics_YOLO_2023} achieved reasonably high performance among the models tested for rip current segmentation, while also being the fastest. However, while it outperformed some models, it still struggled to accurately segment the complex rip current patterns present in our dataset, indicating that even advanced models like YOLO11 require further refinement to address the unique challenges of this task effectively. This performance highlights the difficulty of the problem and the need for continued work in developing specialized approaches for rip current detection.

For YOLO11, we performed the most extensive study, testing multiple configurations to maximize its performance. The study included all size variants (nano, small, medium, large, and x) and tested learning rates of 0.01 and 0.001, along with a weight decay of 0.0005. The models were trained using various optimizers, including SGD with momentum, Adam, AdamW, and standard SGD. The learning rate schedulers included both linear and cosine decay strategies.

We evaluated YOLO11 with both pre-trained weights and custom-trained weights, allowing us to analyze the impact of transfer learning on rip current detection. Pre-trained weights generally resulted in faster convergence and higher initial accuracy, while custom-trained weights offered more flexibility in adapting to the unique characteristics of the RipVIS dataset. 



\noindent
\textbf{SparseInst:} SparseInst \cite{Cheng2022SparseInst} uses sparse instance activation maps for efficient, real-time segmentation, leveraging feature aggregation and bipartite matching to skip post-processing. This lightweight design minimizes computational overhead, making it ideal for dynamic tasks like rip current detection. We tuned it with ResNet-50, ResNet-101, and PVTv2 backbones, adjusting learning rates, optimizers (SGD, AdamW), batch sizes, and sparsity thresholds to balance sensitivity and noise. PVTv2 with data augmentation achieved the highest $F_2$ score among all models, alongside top $F_1$ and fast inference, making SparseInst the best overall choice for rip current detection.

\begin{figure*}
\centering
\setlength{\tabcolsep}{1pt}
\begin{tabular}{c c c c c}
    
     \vspace{-1mm} 
     \includegraphics[width=0.195\textwidth]{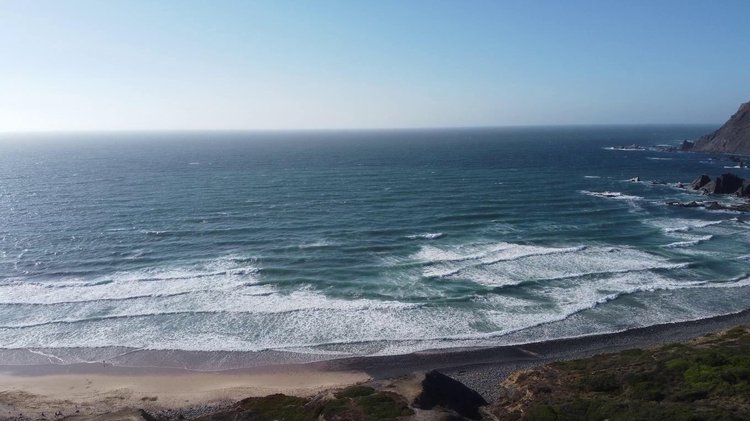} &
     \includegraphics[width=0.195\textwidth]{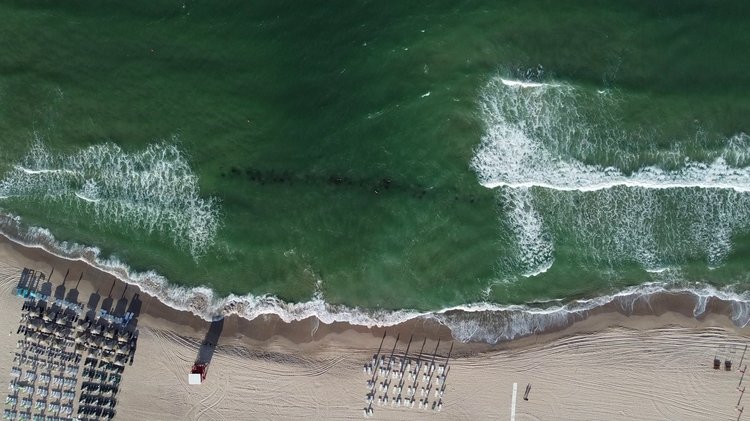} &
     \includegraphics[width=0.195\textwidth]{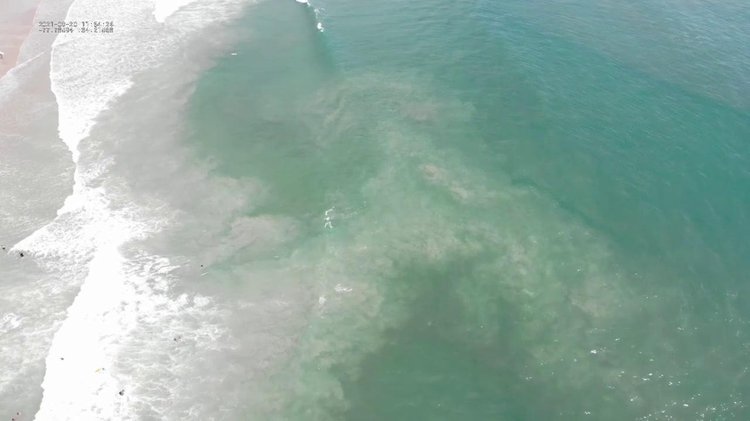} &
     \includegraphics[width=0.195\textwidth]{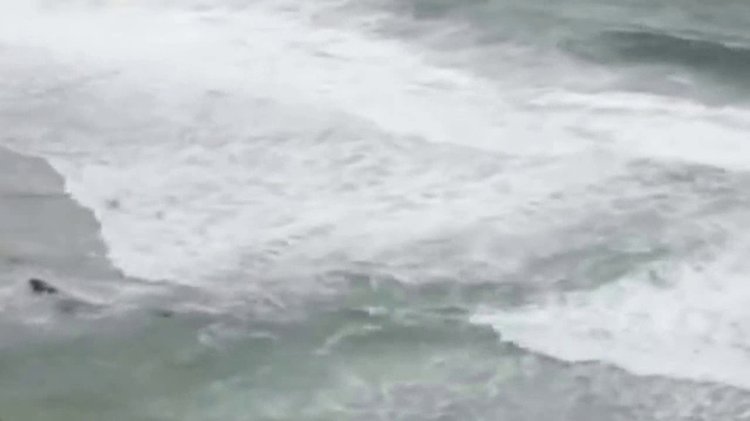} &
     \includegraphics[width=0.195\textwidth]{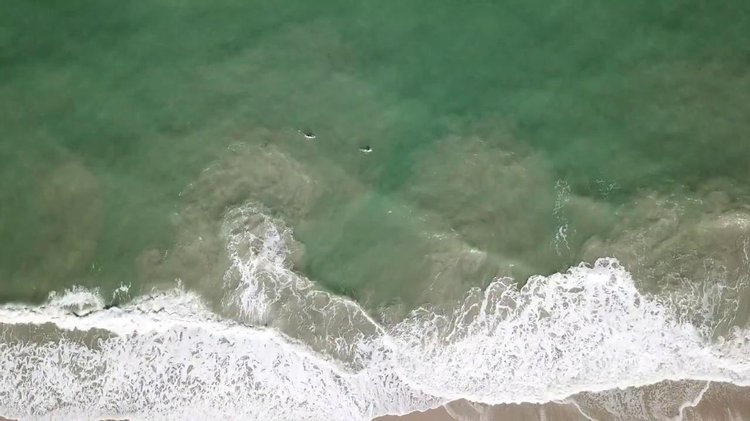} 
     \tabularnewline     
     \vspace{-1mm} 
     \includegraphics[width=0.195\textwidth]{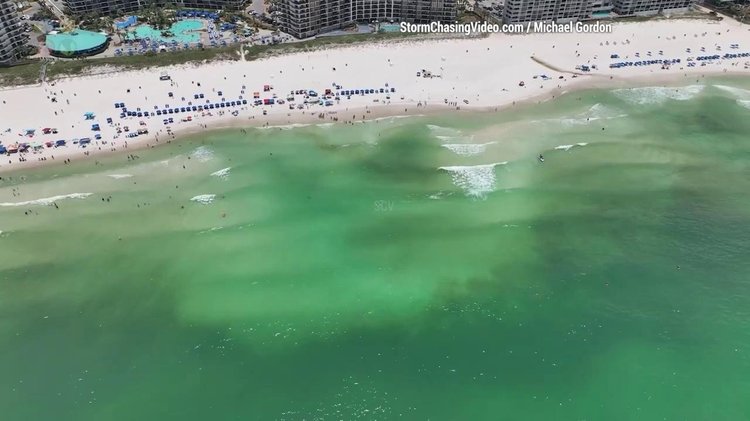} &
     \includegraphics[width=0.195\textwidth]{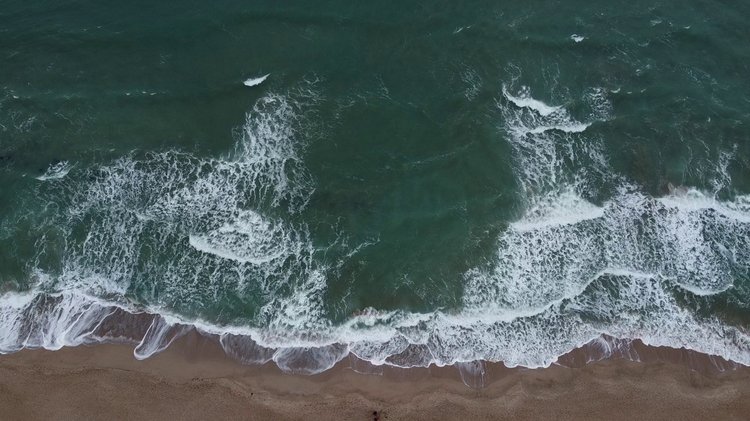} &
     \includegraphics[width=0.195\textwidth]{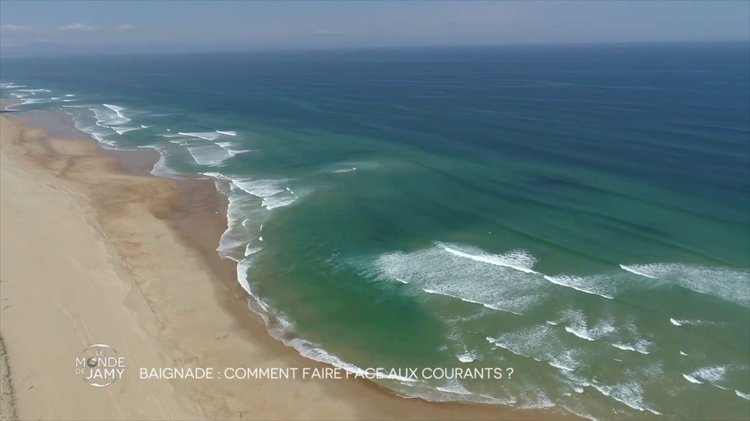} &
     \includegraphics[width=0.195\textwidth]{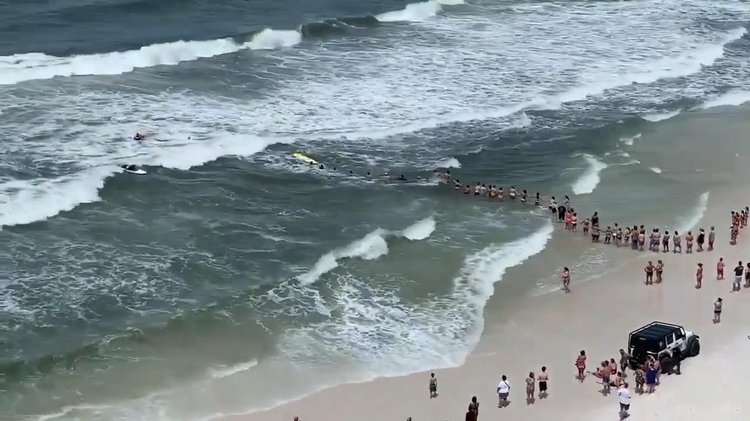} &
     \includegraphics[width=0.195\textwidth]{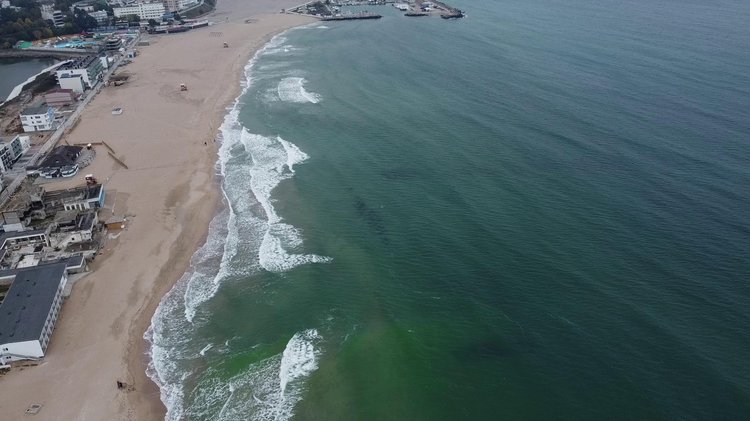} 
     \tabularnewline
     \vspace{-1mm} 
     \includegraphics[width=0.195\textwidth]{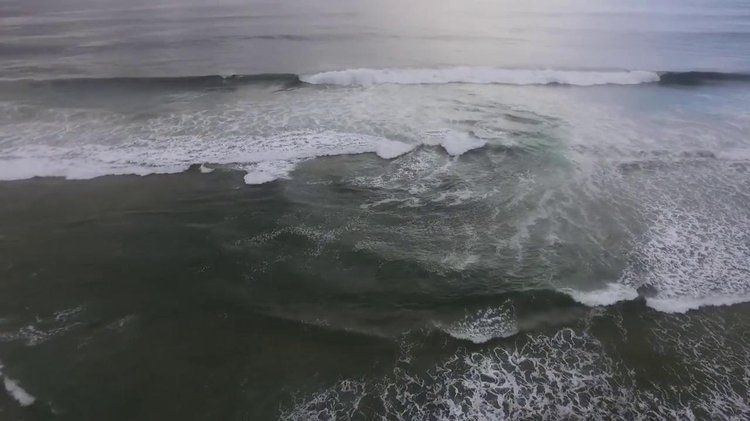} &
     \includegraphics[width=0.195\textwidth]{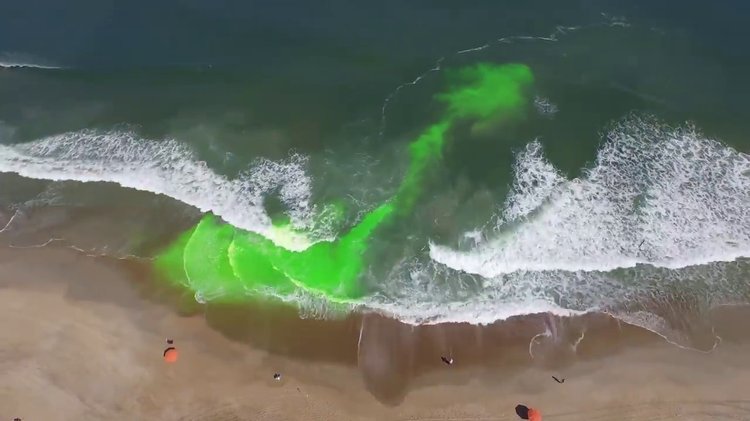} &
     \includegraphics[width=0.195\textwidth]{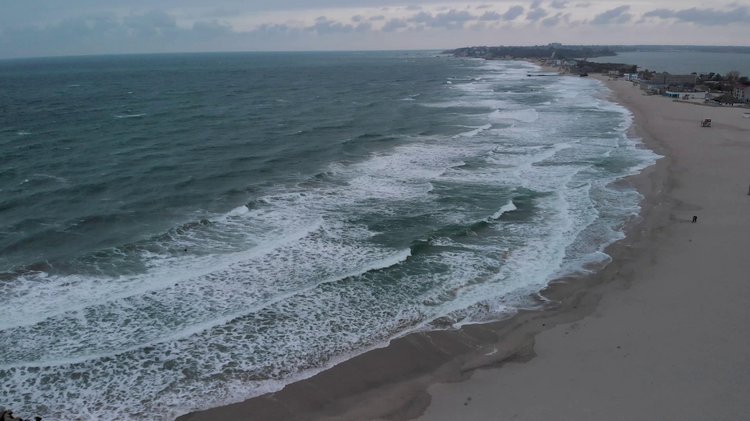} &
     \includegraphics[width=0.195\textwidth]{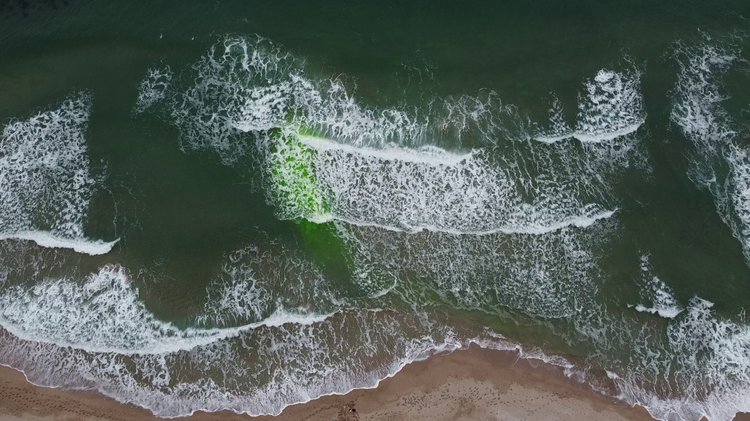} &
     \includegraphics[width=0.195\textwidth]{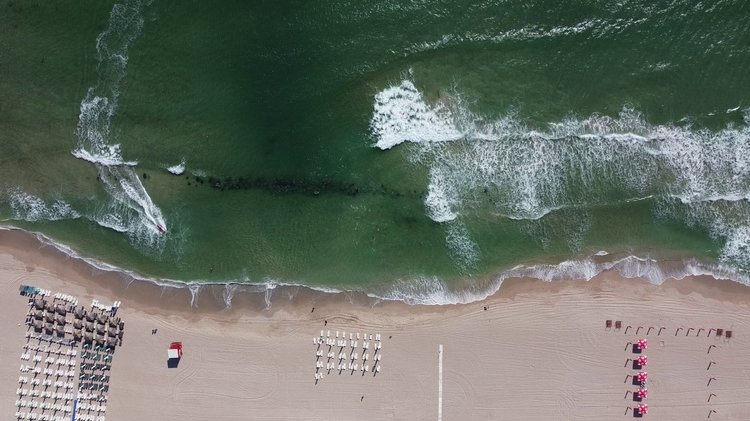} 
     \tabularnewline
     \vspace{-1mm} 
     \includegraphics[width=0.195\textwidth]{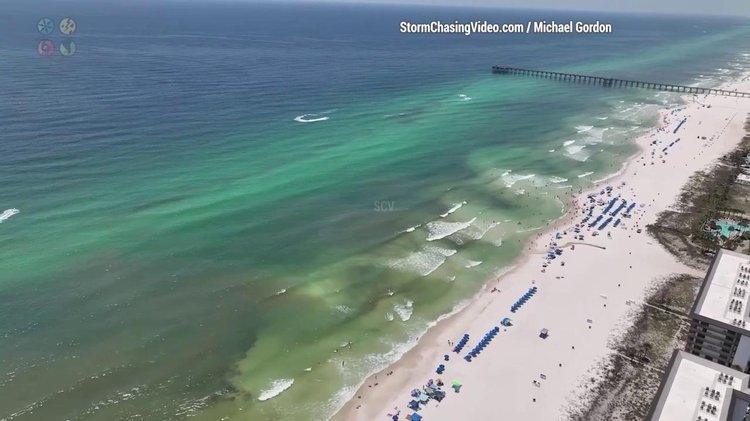} &
     \includegraphics[width=0.195\textwidth]{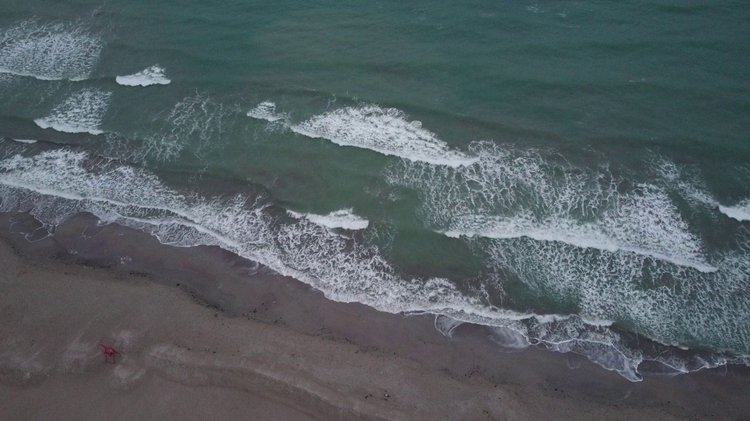} &
     \includegraphics[width=0.195\textwidth]{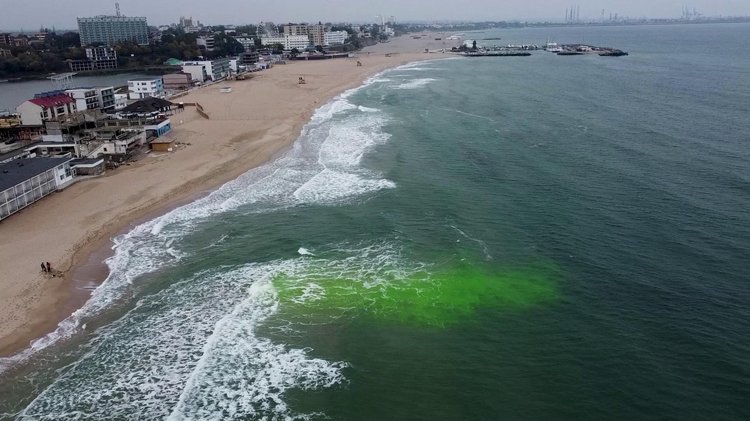} &
     \includegraphics[width=0.195\textwidth]{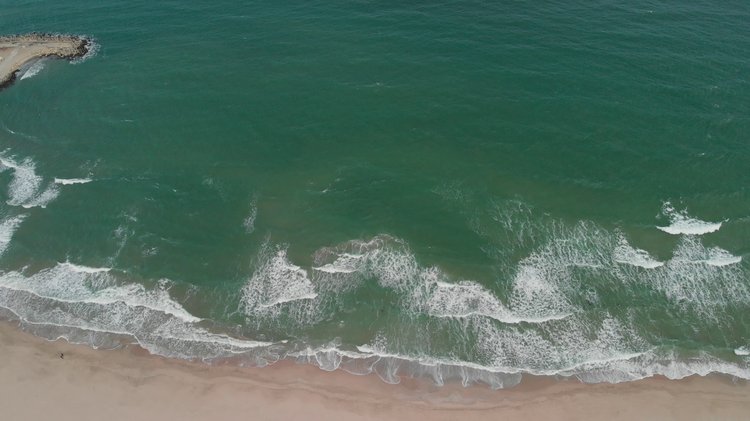} &
     \includegraphics[width=0.195\textwidth]{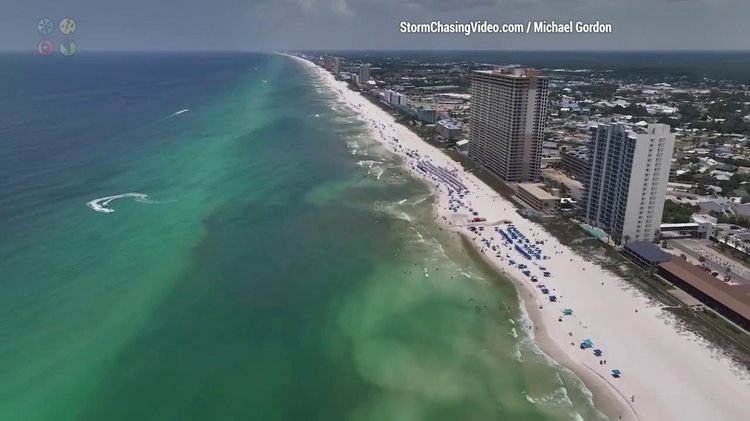} 
     \tabularnewline
     \vspace{-1mm} 
     \includegraphics[width=0.195\textwidth]{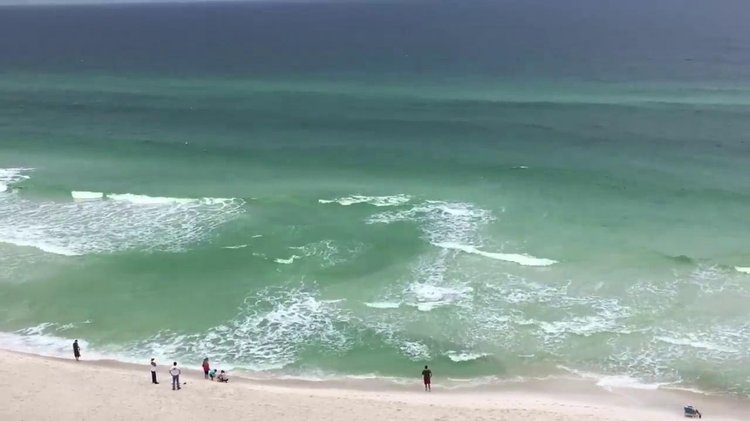} &
     \includegraphics[width=0.195\textwidth]{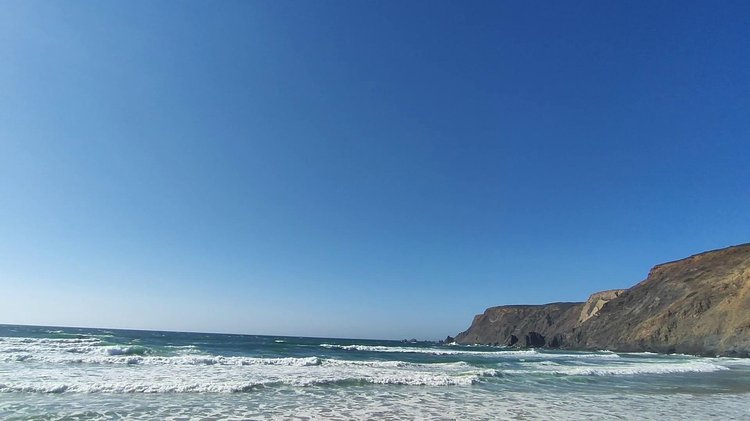} &
     \includegraphics[width=0.195\textwidth]{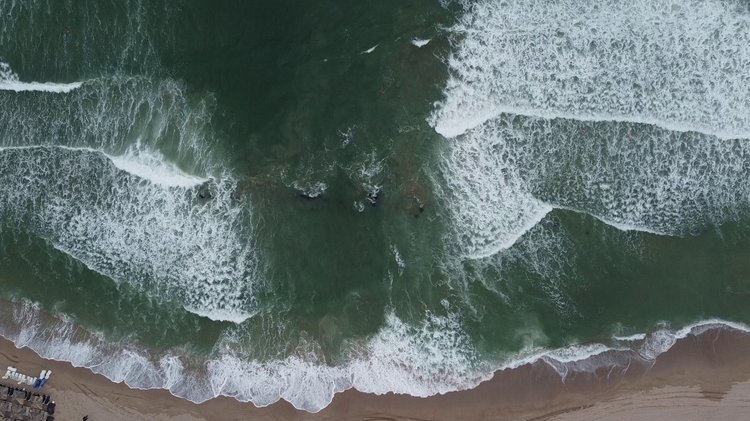} &
     \includegraphics[width=0.195\textwidth]{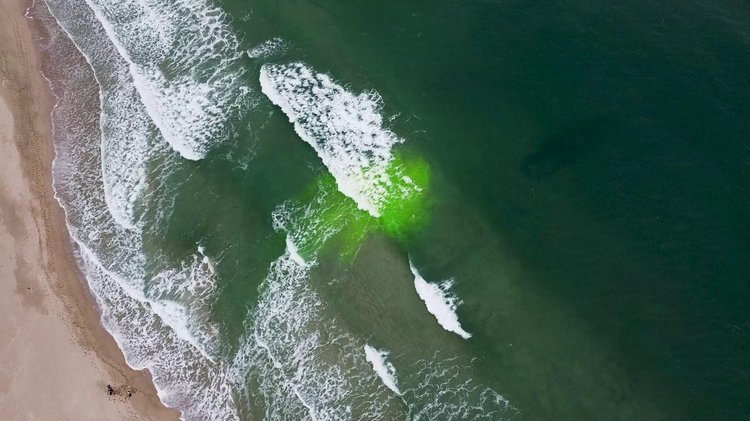} &
     \includegraphics[width=0.195\textwidth]{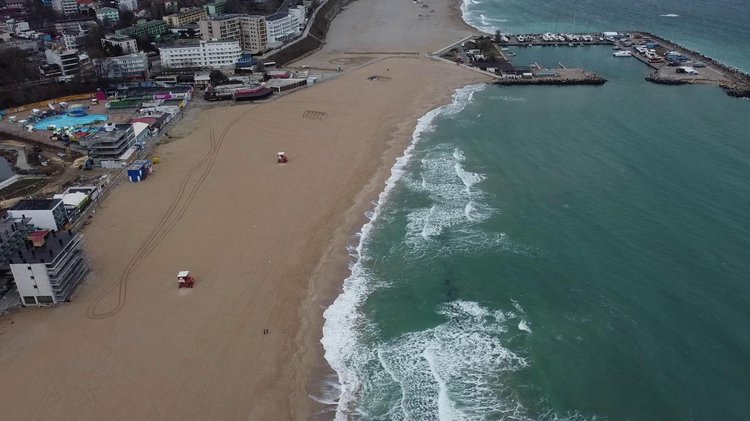} 
     \tabularnewline
     \vspace{-1mm} 
     \includegraphics[width=0.195\textwidth]{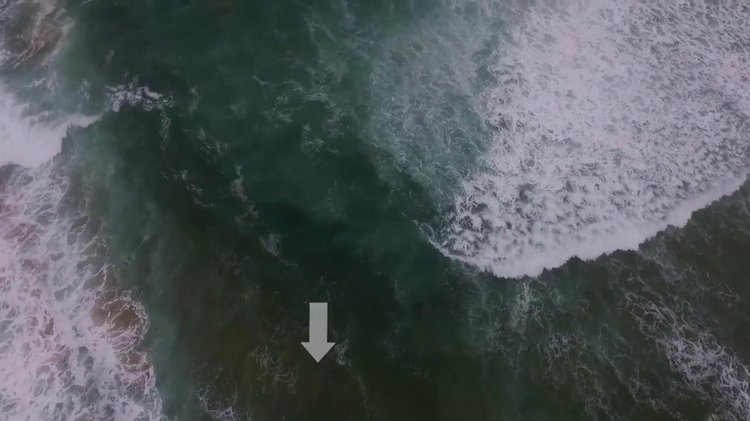} &
     \includegraphics[width=0.195\textwidth]{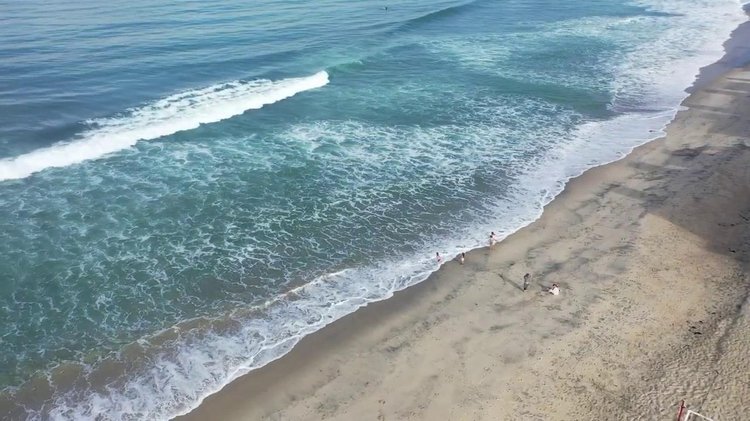} &
     \includegraphics[width=0.195\textwidth]{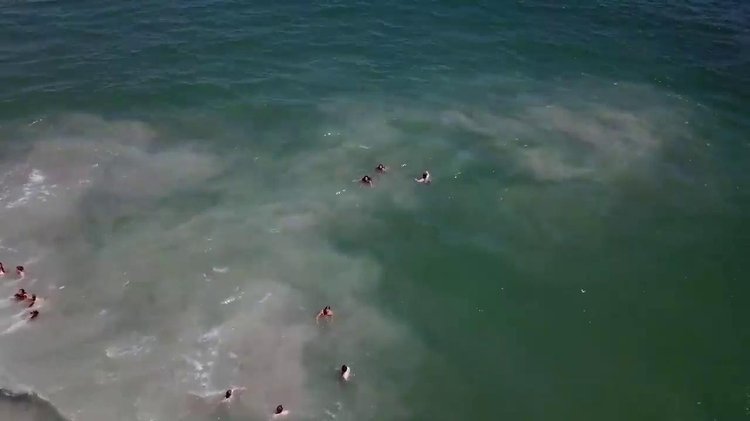} &
     \includegraphics[width=0.195\textwidth]{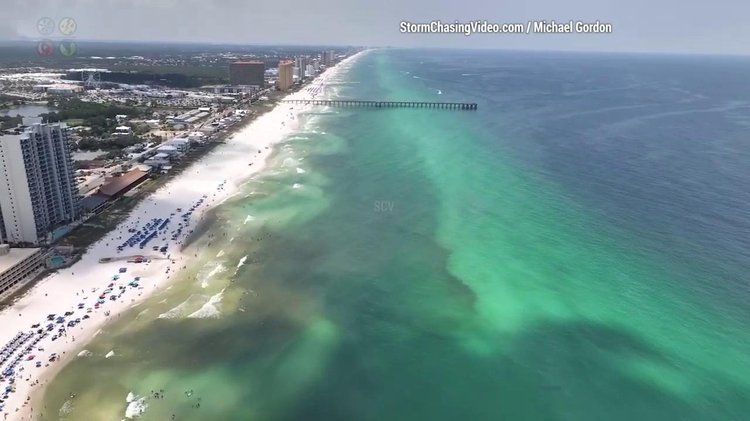} &
     \includegraphics[width=0.195\textwidth]{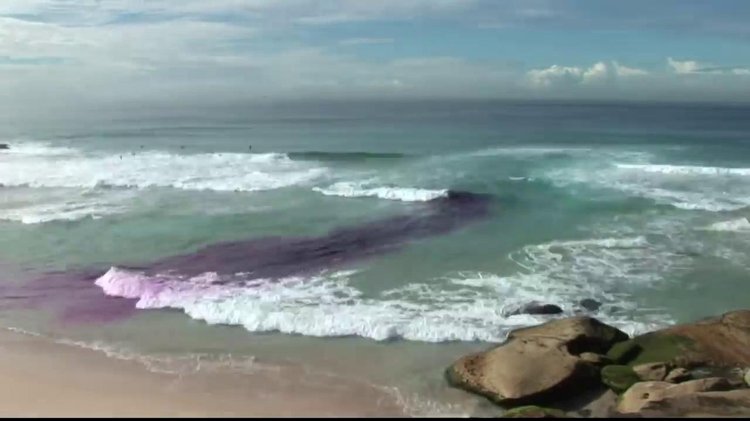} 
     \tabularnewline
     \vspace{-1mm} 
     \includegraphics[width=0.195\textwidth]{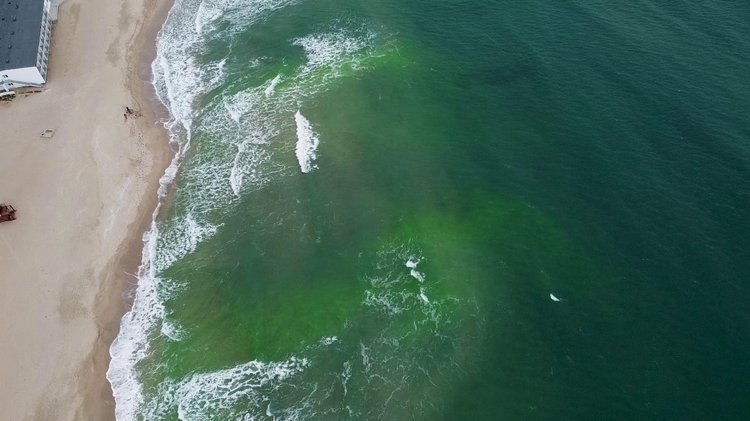} &
     \includegraphics[width=0.195\textwidth]{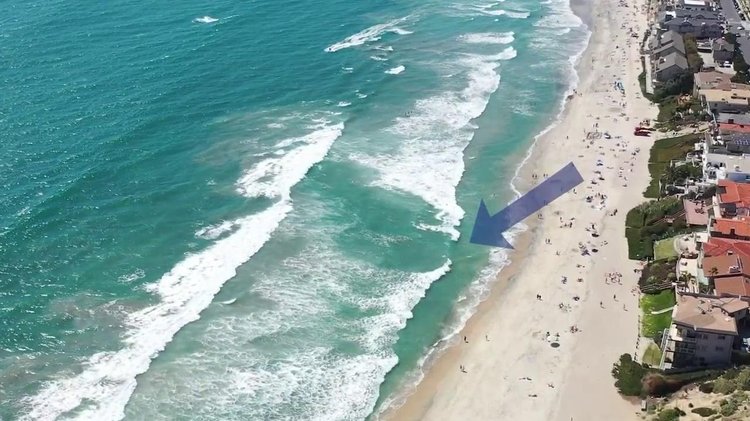} &
     \includegraphics[width=0.195\textwidth]{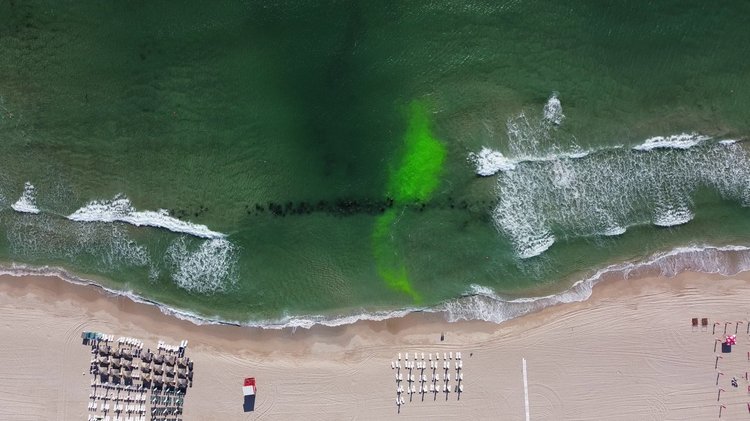} &
     \includegraphics[width=0.195\textwidth]{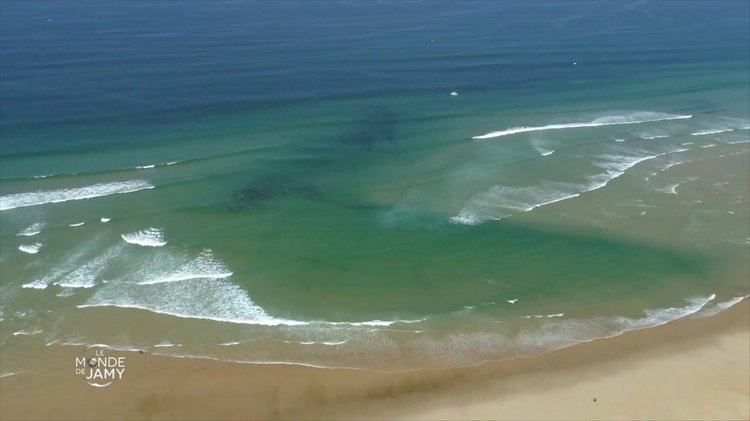} &
     \includegraphics[width=0.195\textwidth]{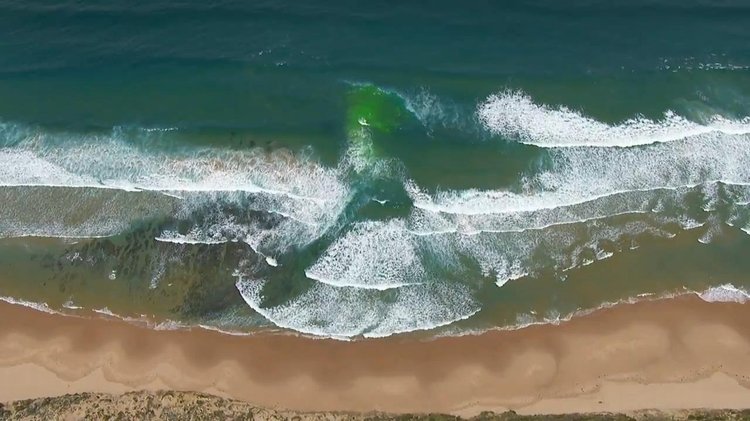} 
     \tabularnewline
     \vspace{-1mm} 
     \includegraphics[width=0.195\textwidth]{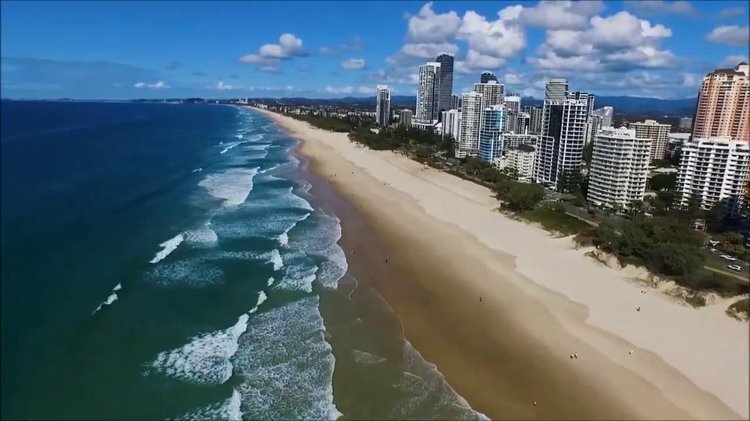} &
     \includegraphics[width=0.195\textwidth]{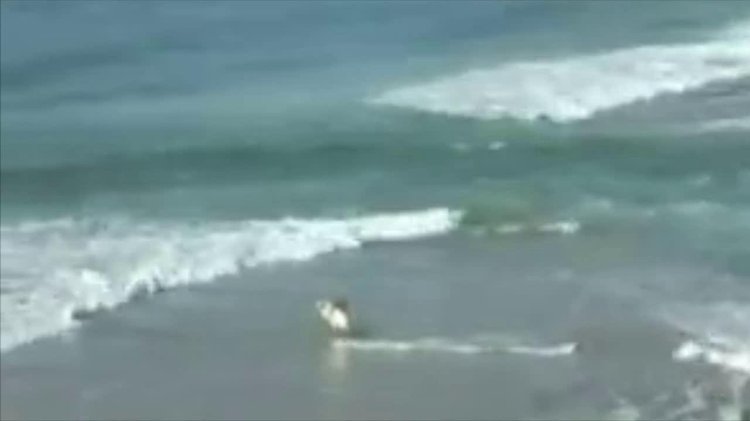} &
     \includegraphics[width=0.195\textwidth]{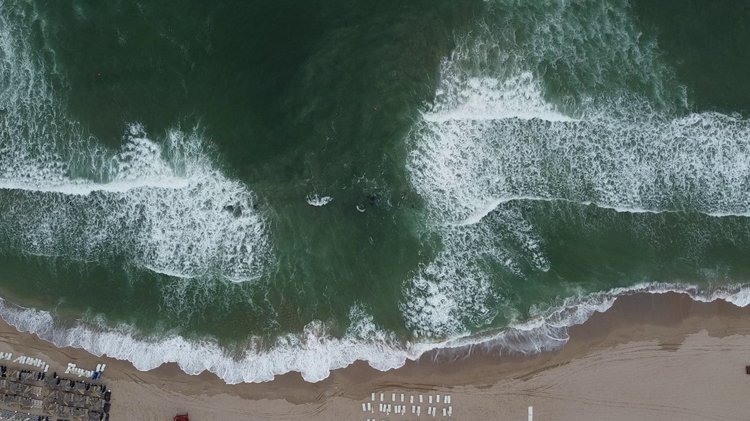} &
     \includegraphics[width=0.195\textwidth]{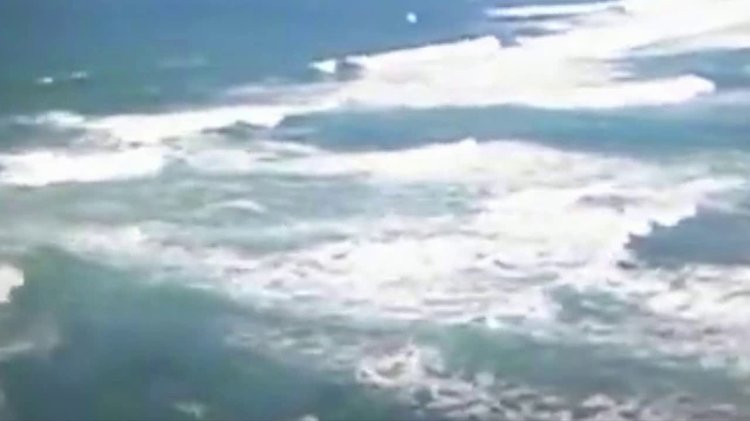} &
     \includegraphics[width=0.195\textwidth]{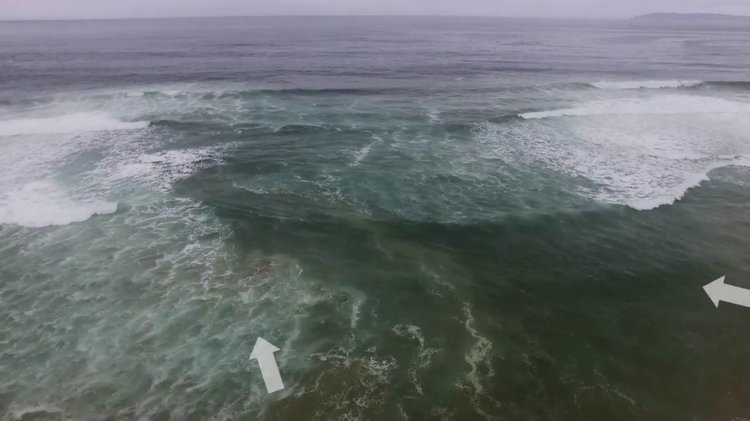} 
     \tabularnewline
     \vspace{-1mm} 
     \includegraphics[width=0.195\textwidth]{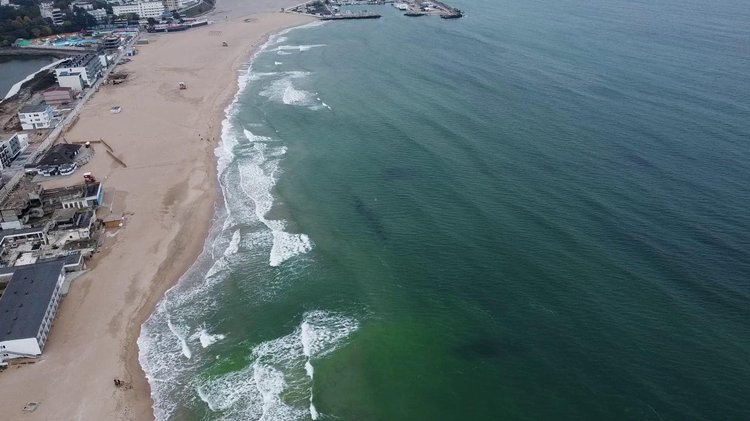} &
     \includegraphics[width=0.195\textwidth]{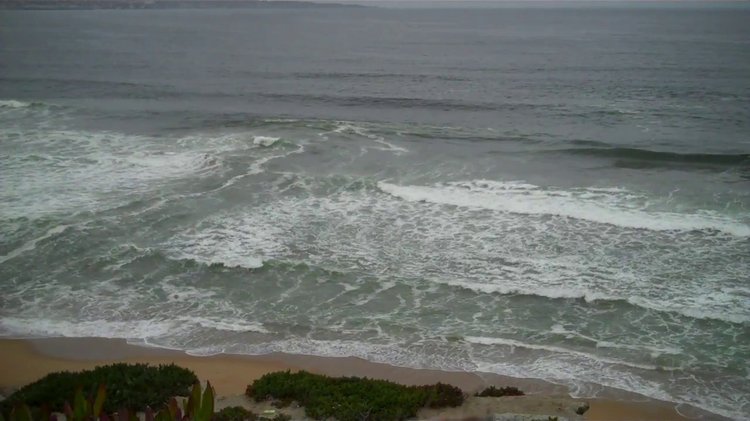} &
     \includegraphics[width=0.195\textwidth]{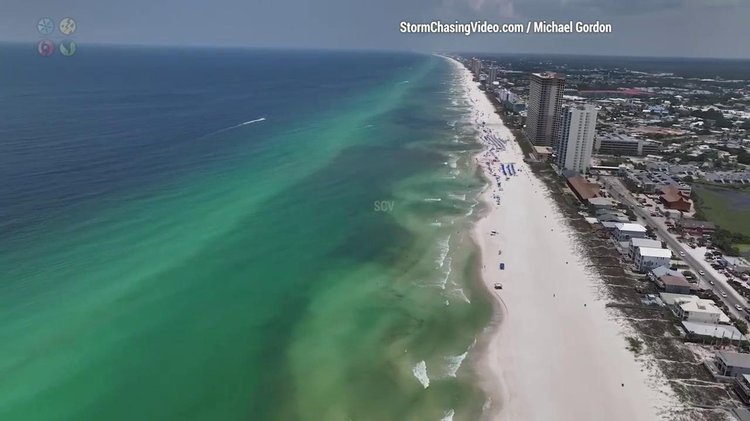} &
     \includegraphics[width=0.195\textwidth]{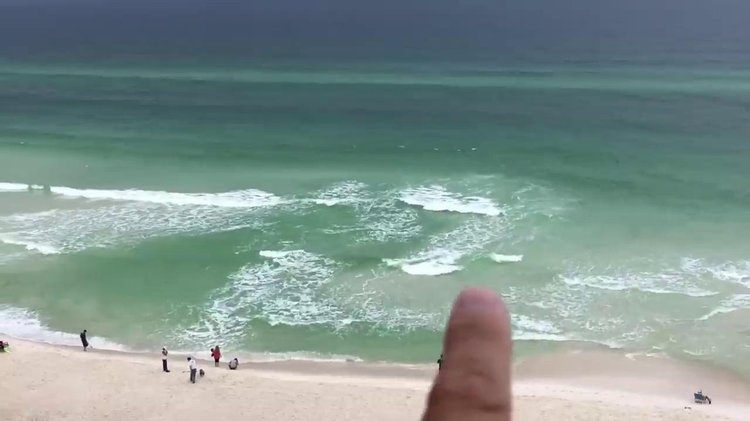} &
     \includegraphics[width=0.195\textwidth]{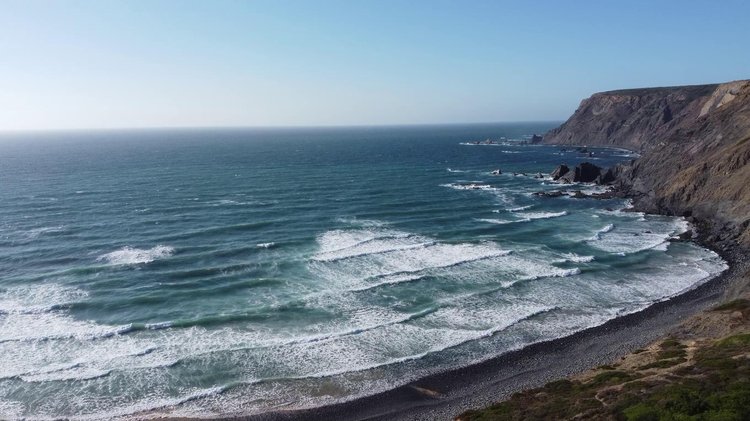} 
     \tabularnewline
     \vspace{-1mm} 
     \includegraphics[width=0.195\textwidth]{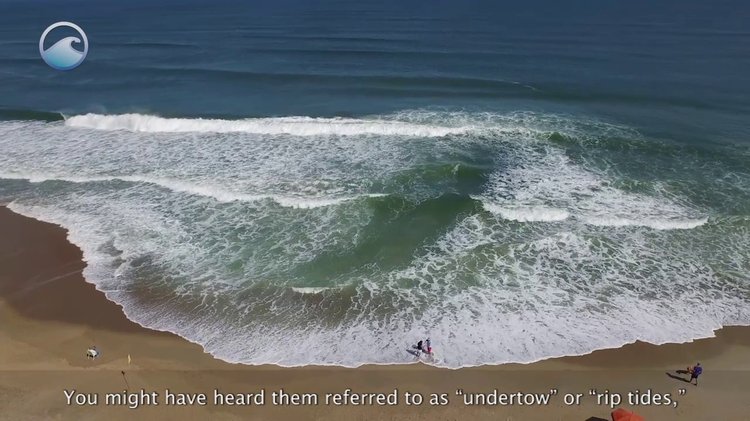} &
     \includegraphics[width=0.195\textwidth]{figures/dataset-variety/no-mask/RipVIS-084-middle_frame.jpg} &
     \includegraphics[width=0.195\textwidth]{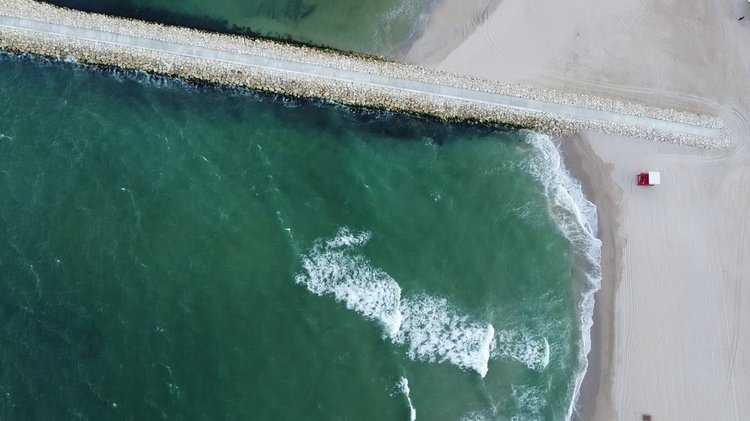} &
     \includegraphics[width=0.195\textwidth]{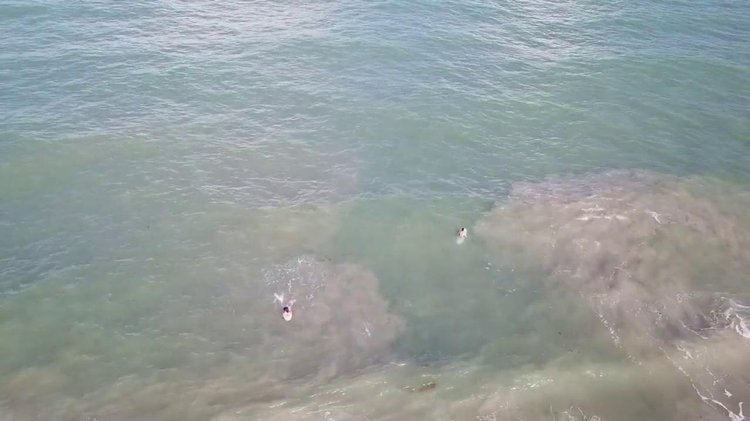} &
     \includegraphics[width=0.195\textwidth]{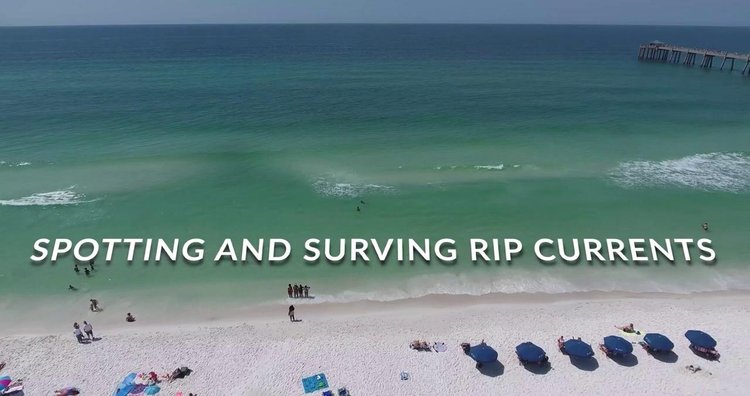} 
     \tabularnewline
\end{tabular}
  \caption{Examples of rip currents from the dataset, showcasing its diverse nature. Here we show frames from 55 randomly selected videos (out of 115 with rip currents). \textcolor{red}{\textbf{Can you spot them all?}} Some are easy, while others can be deceiving at first glance. }
  \label{fig:dataset_diversity}
\end{figure*}

\begin{figure*}
\centering
\setlength{\tabcolsep}{1pt}
\begin{tabular}{c c c c c}
    
     \vspace{-1mm} 
     \includegraphics[width=0.195\textwidth]{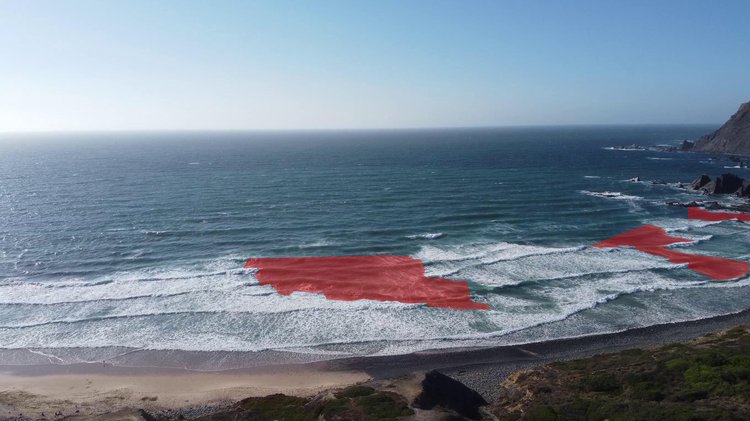} &
     \includegraphics[width=0.195\textwidth]{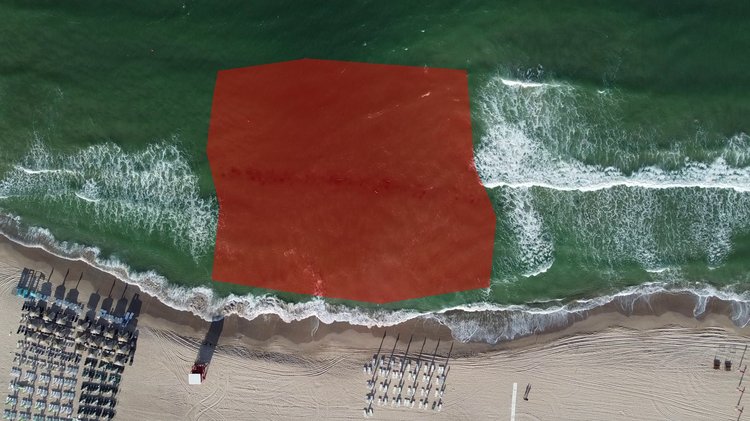} &
     \includegraphics[width=0.195\textwidth]{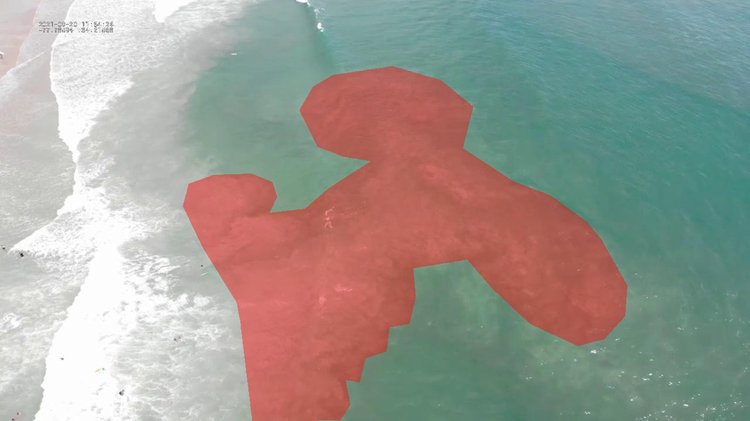} &
     \includegraphics[width=0.195\textwidth]{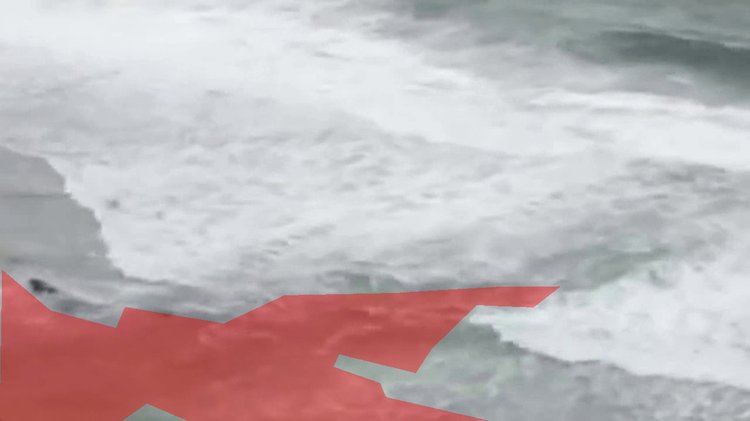} &
     \includegraphics[width=0.195\textwidth]{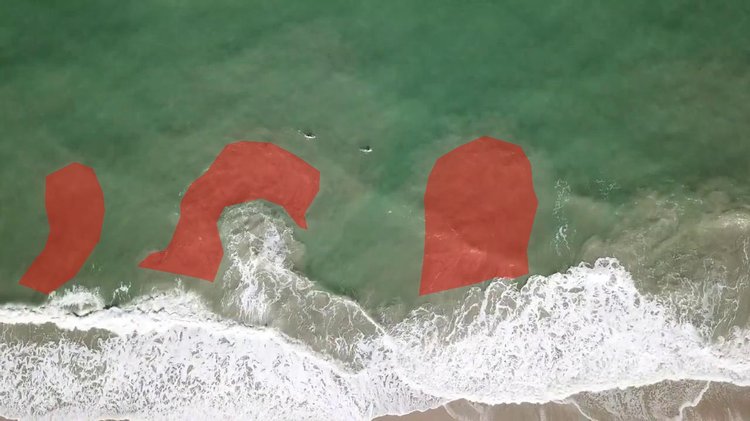} 
     \tabularnewline     
     \vspace{-1mm} 
     \includegraphics[width=0.195\textwidth]{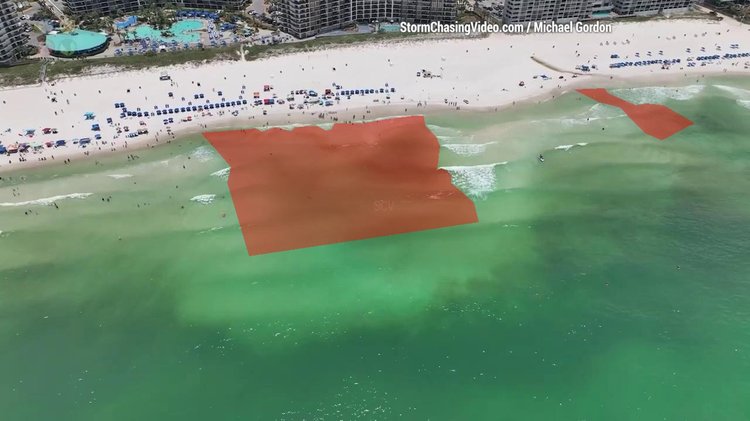} &
     \includegraphics[width=0.195\textwidth]{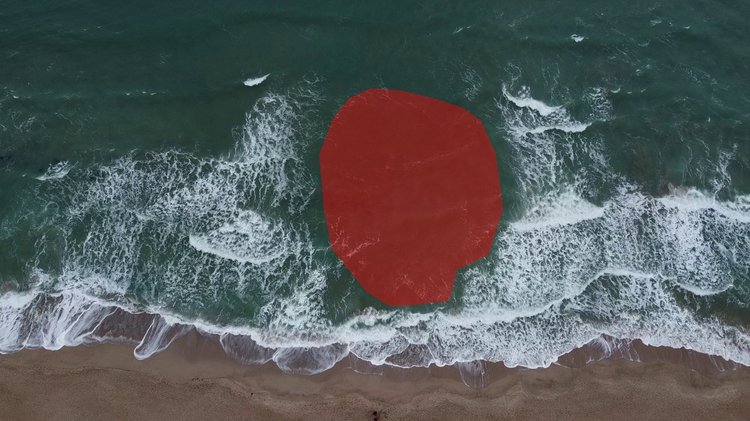} &
     \includegraphics[width=0.195\textwidth]{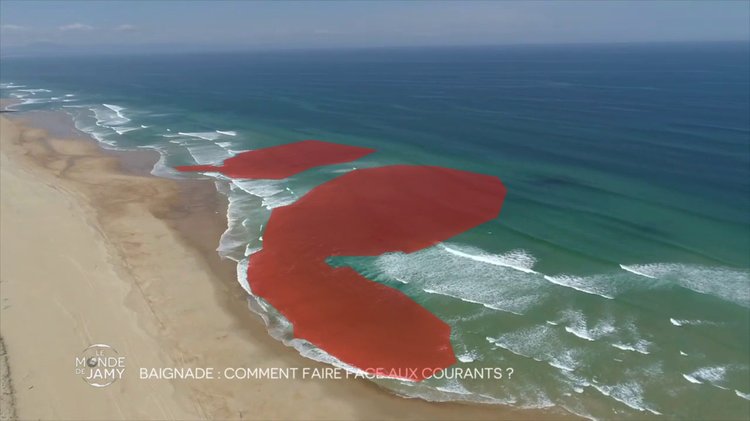} &
     \includegraphics[width=0.195\textwidth]{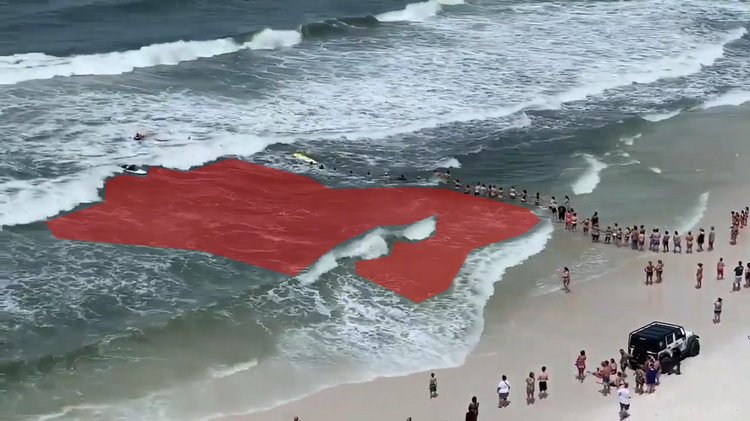} &
     \includegraphics[width=0.195\textwidth]{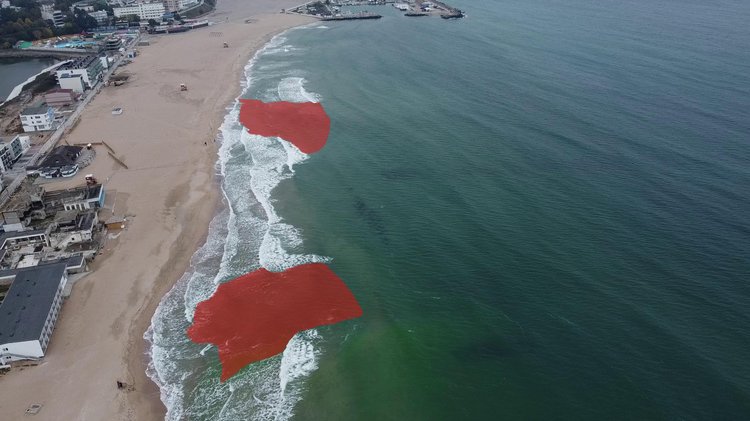} 
     \tabularnewline
     \vspace{-1mm} 
     \includegraphics[width=0.195\textwidth]{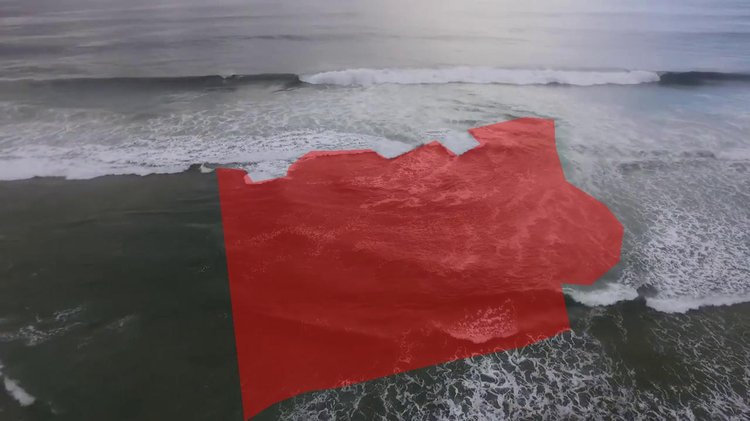} &
     \includegraphics[width=0.195\textwidth]{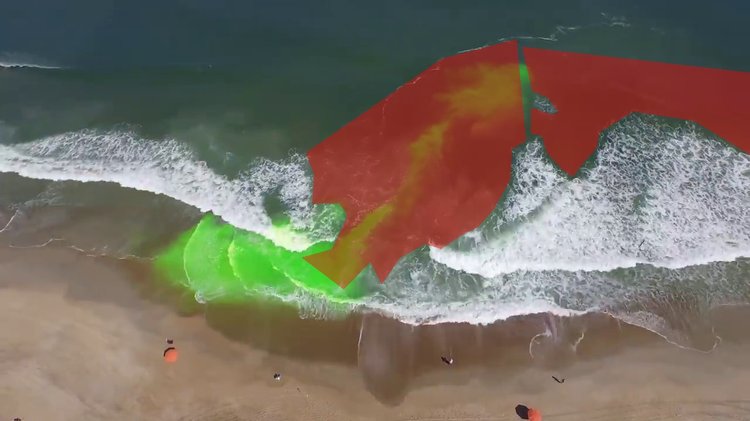} &
     \includegraphics[width=0.195\textwidth]{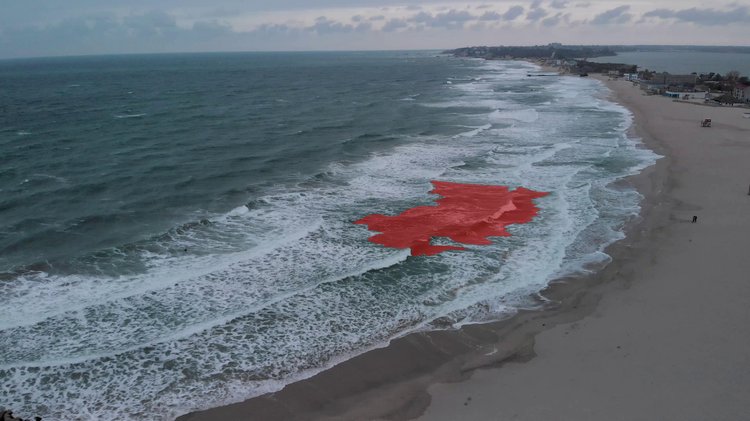} &
     \includegraphics[width=0.195\textwidth]{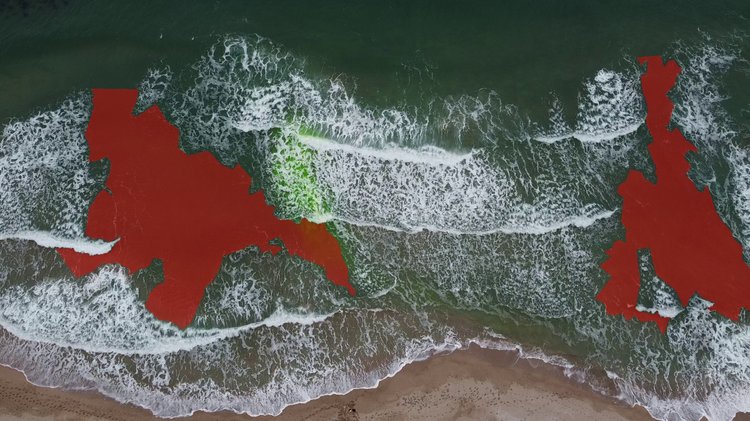} &
     \includegraphics[width=0.195\textwidth]{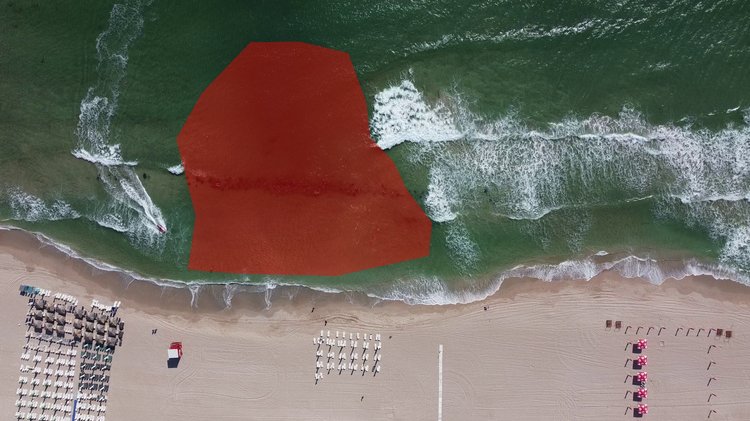} 
     \tabularnewline
     \vspace{-1mm} 
     \includegraphics[width=0.195\textwidth]{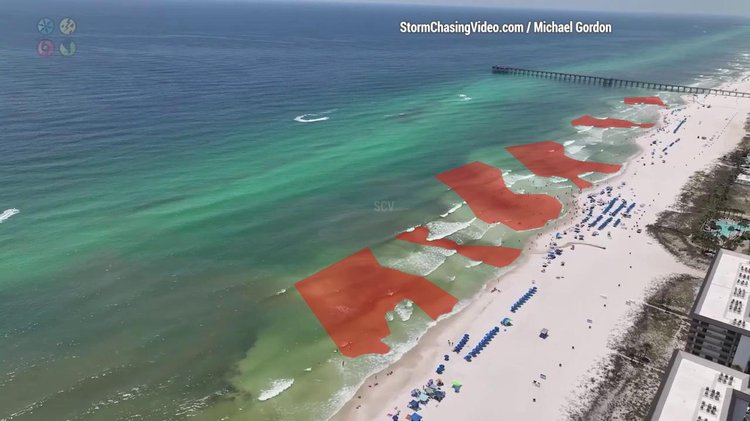} &
     \includegraphics[width=0.195\textwidth]{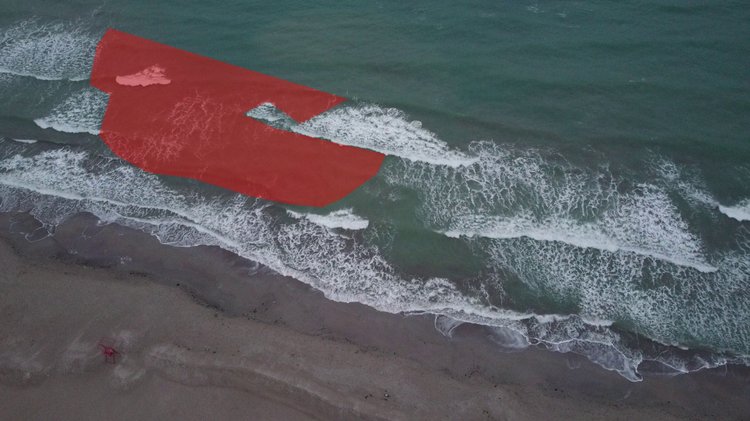} &
     \includegraphics[width=0.195\textwidth]{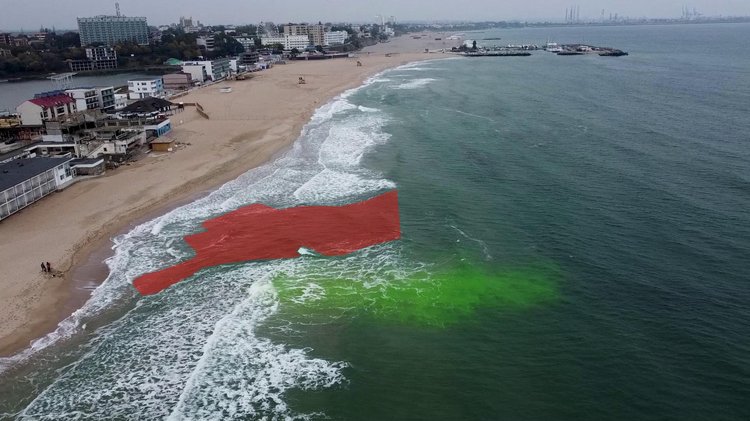} &
     \includegraphics[width=0.195\textwidth]{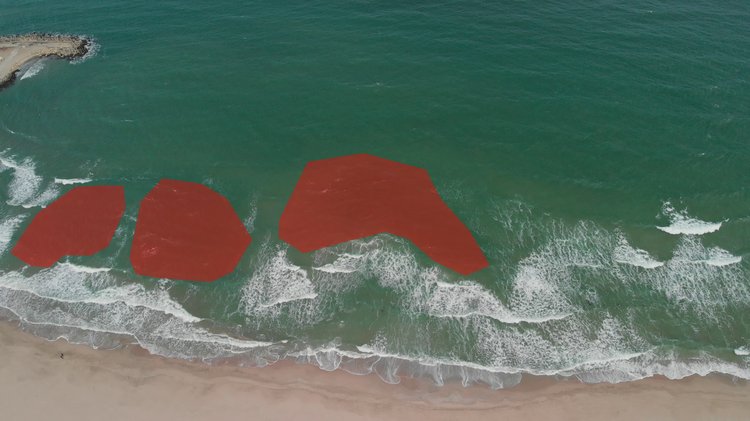} &
     \includegraphics[width=0.195\textwidth]{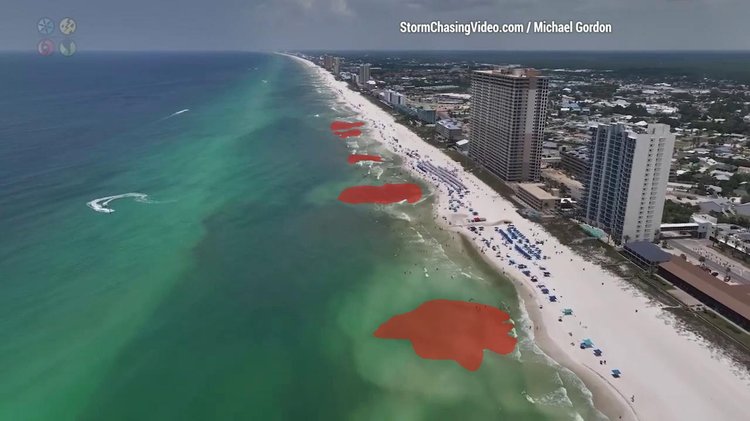} 
     \tabularnewline
     \vspace{-1mm} 
     \includegraphics[width=0.195\textwidth]{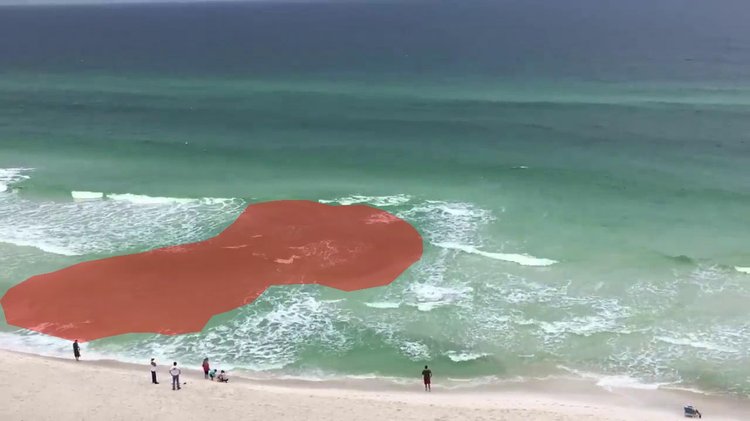} &
     \includegraphics[width=0.195\textwidth]{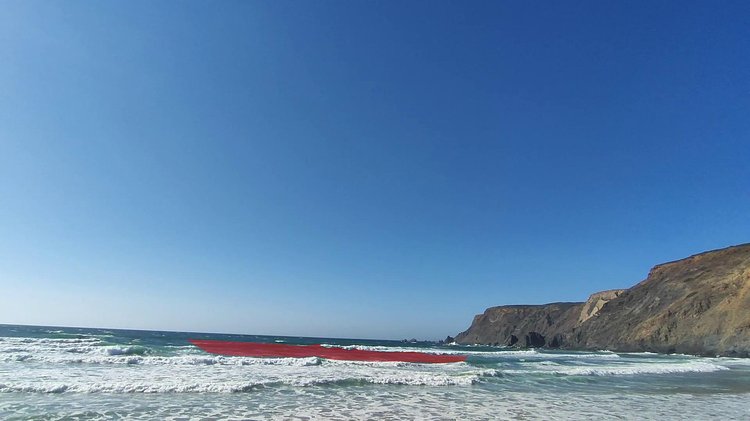} &
     \includegraphics[width=0.195\textwidth]{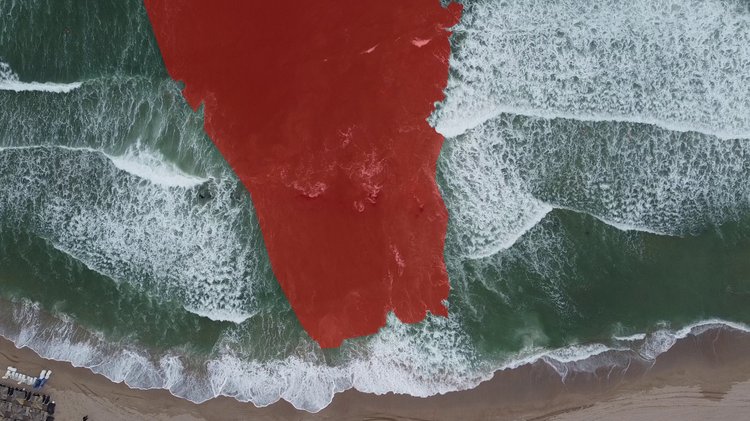} &
     \includegraphics[width=0.195\textwidth]{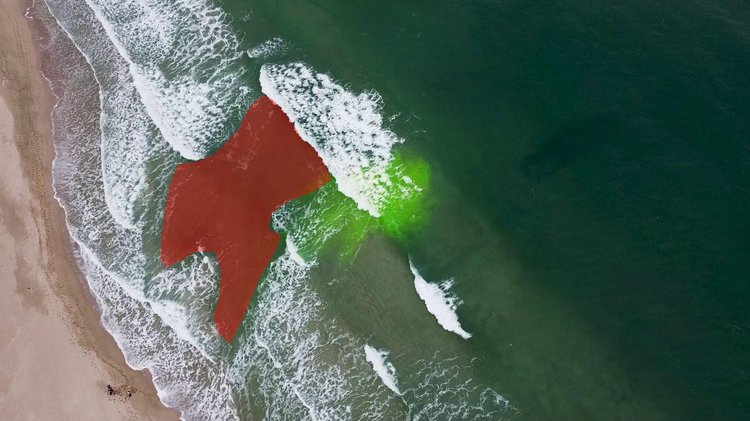} &
     \includegraphics[width=0.195\textwidth]{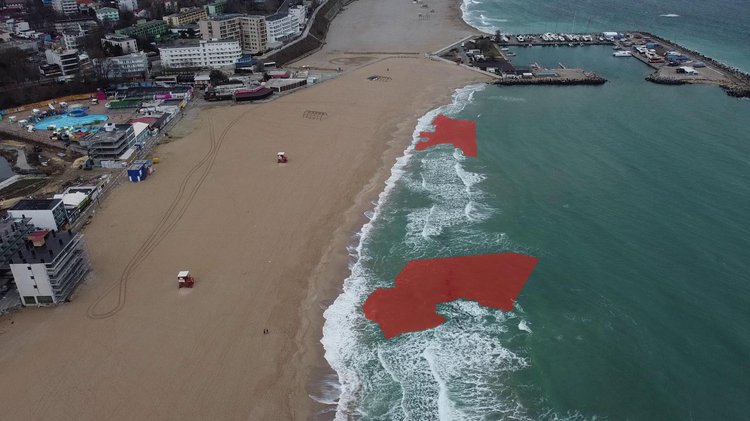} 
     \tabularnewline
     \vspace{-1mm} 
     \includegraphics[width=0.195\textwidth]{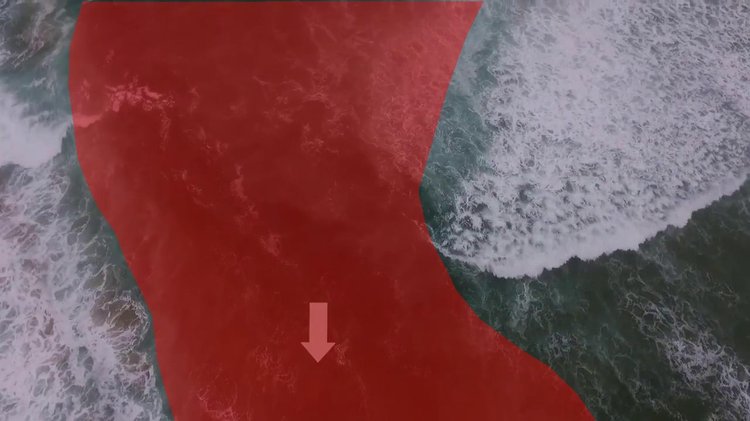} &
     \includegraphics[width=0.195\textwidth]{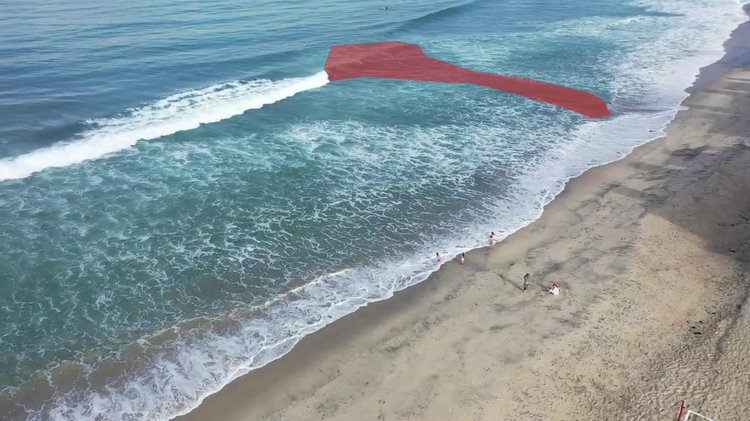} &
     \includegraphics[width=0.195\textwidth]{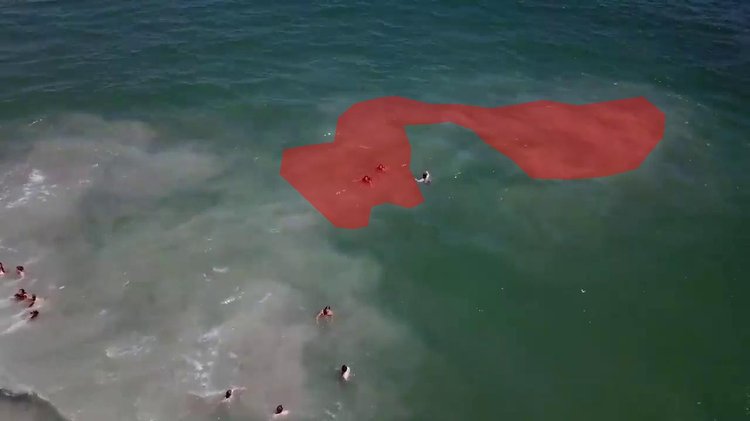} &
     \includegraphics[width=0.195\textwidth]{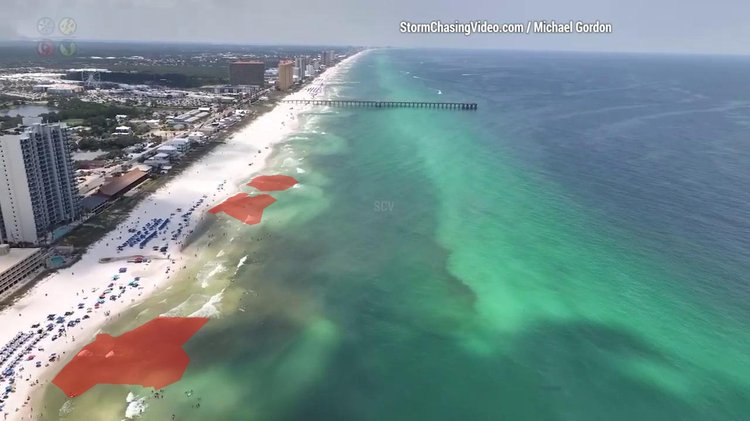} &
     \includegraphics[width=0.195\textwidth]{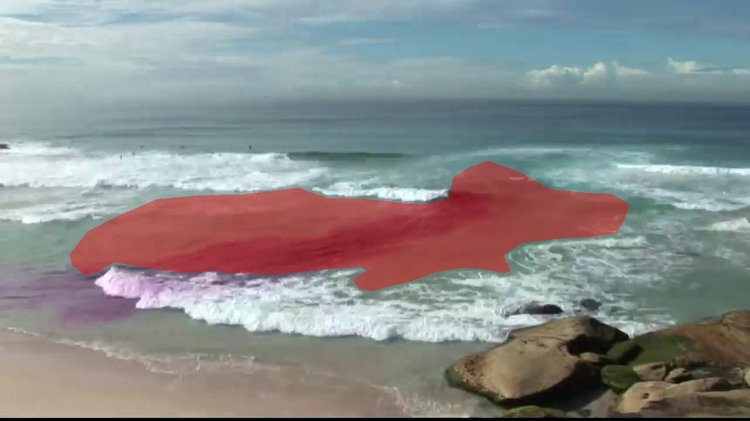} 
     \tabularnewline
     \vspace{-1mm} 
     \includegraphics[width=0.195\textwidth]{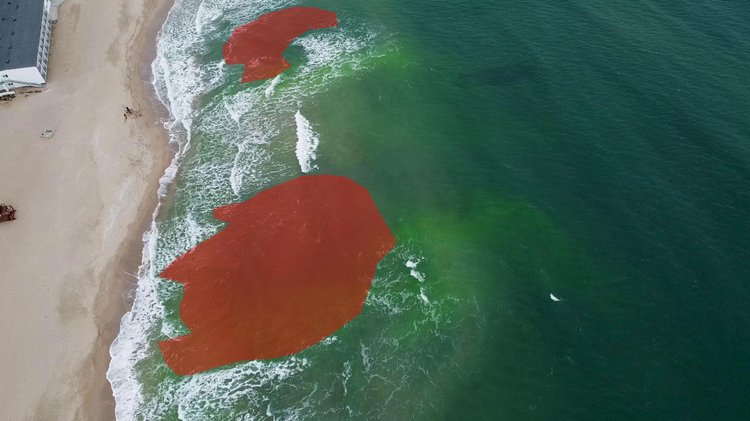} &
     \includegraphics[width=0.195\textwidth]{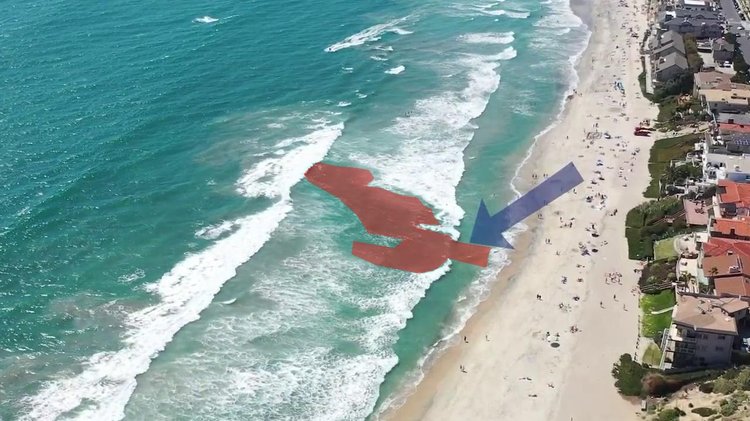} &
     \includegraphics[width=0.195\textwidth]{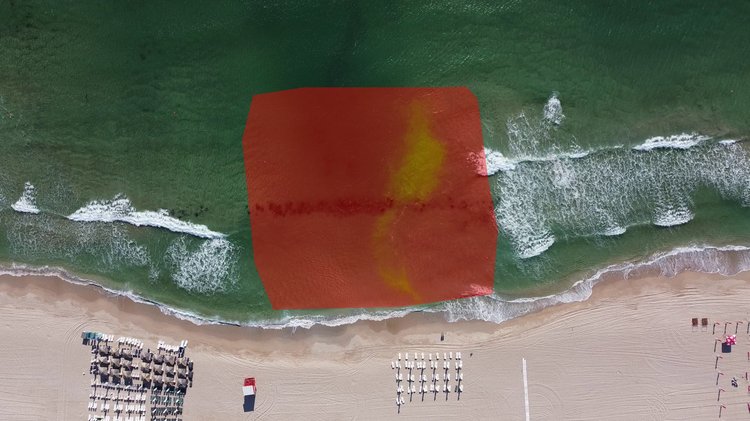} &
     \includegraphics[width=0.195\textwidth]{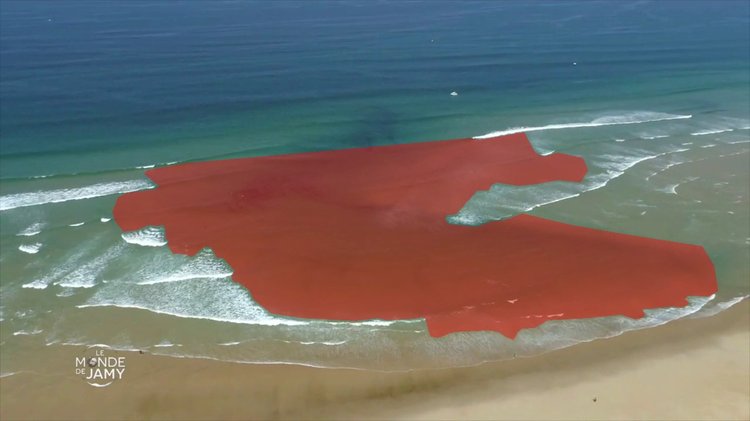} &
     \includegraphics[width=0.195\textwidth]{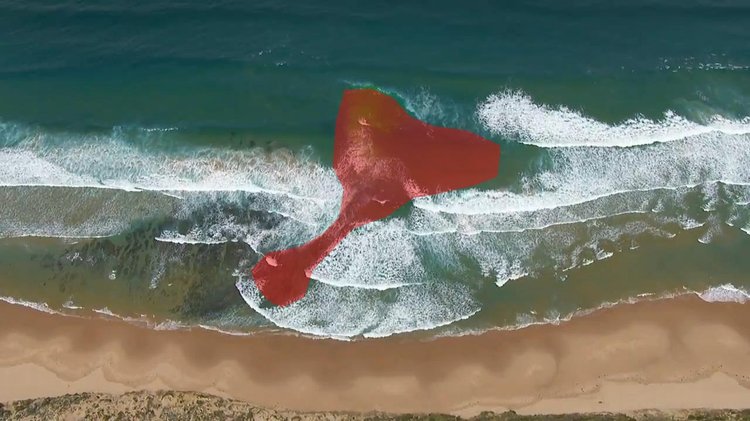} 
     \tabularnewline
     \vspace{-1mm} 
     \includegraphics[width=0.195\textwidth]{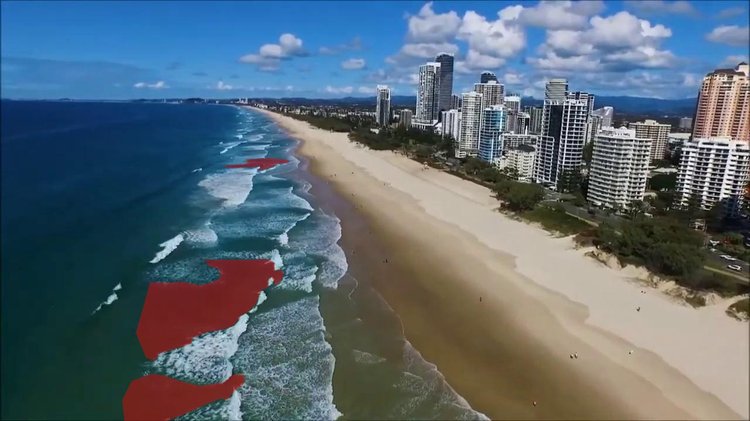} &
     \includegraphics[width=0.195\textwidth]{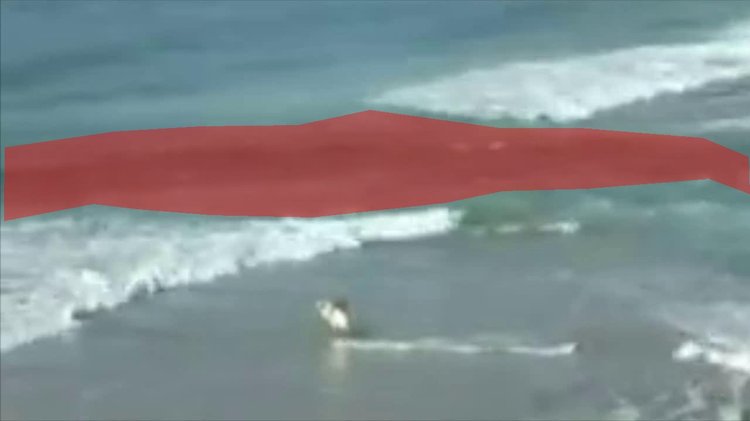} &
     \includegraphics[width=0.195\textwidth]{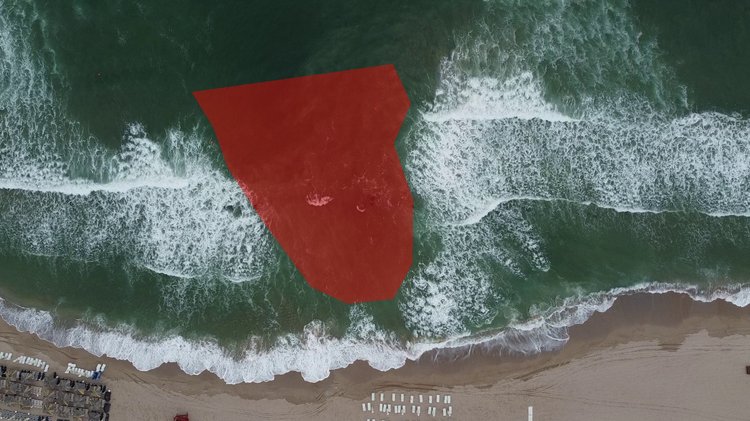} &
     \includegraphics[width=0.195\textwidth]{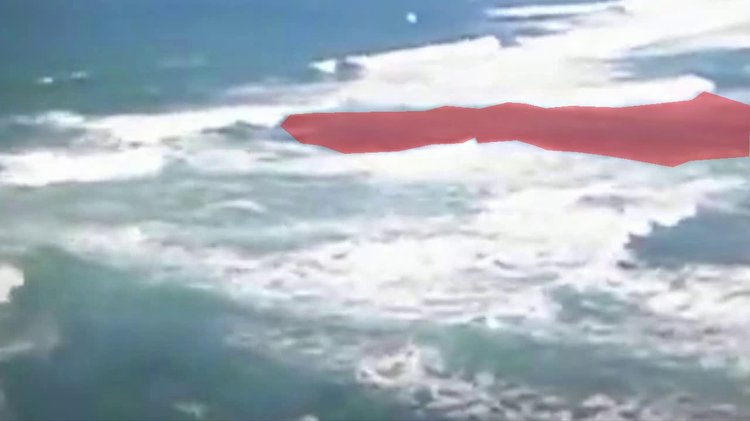} &
     \includegraphics[width=0.195\textwidth]{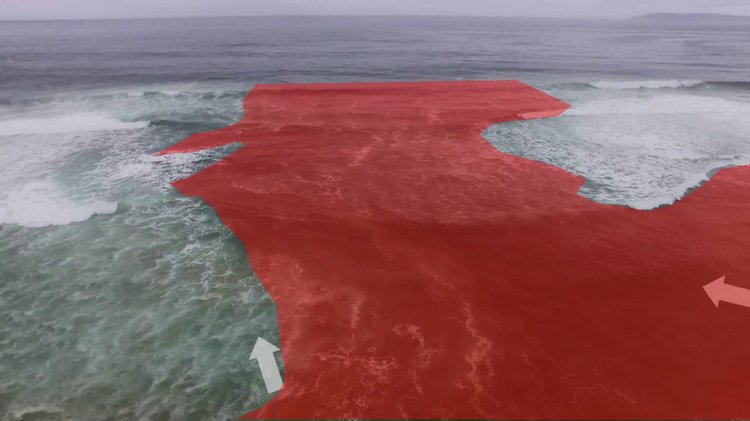} 
     \tabularnewline
     \vspace{-1mm} 
     \includegraphics[width=0.195\textwidth]{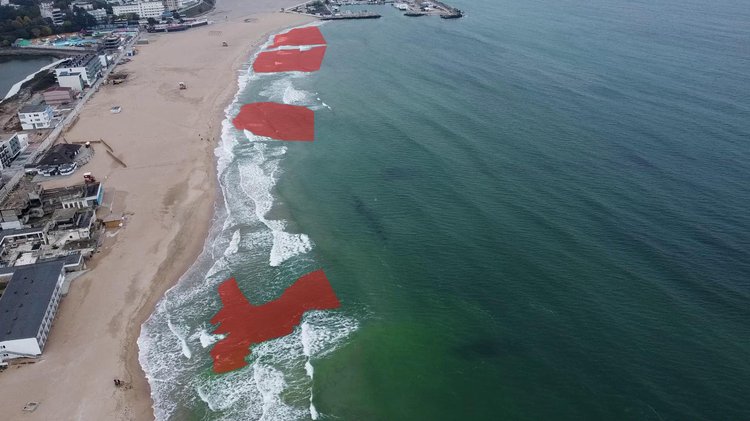} &
     \includegraphics[width=0.195\textwidth]{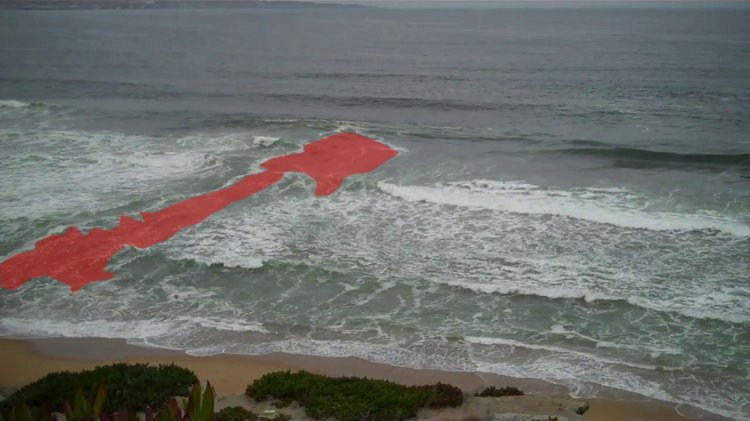} &
     \includegraphics[width=0.195\textwidth]{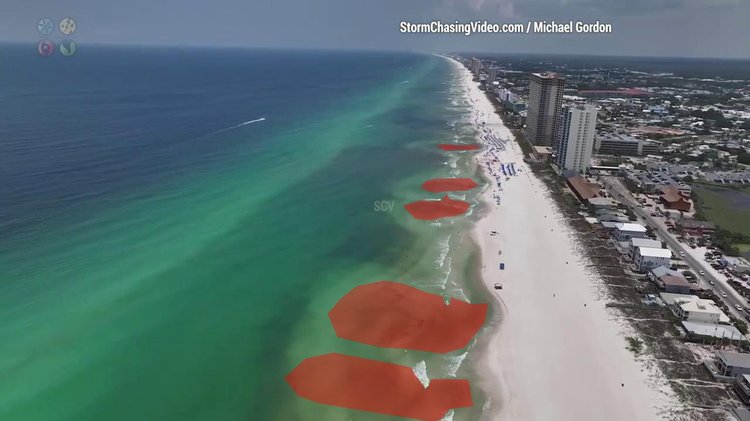} &
     \includegraphics[width=0.195\textwidth]{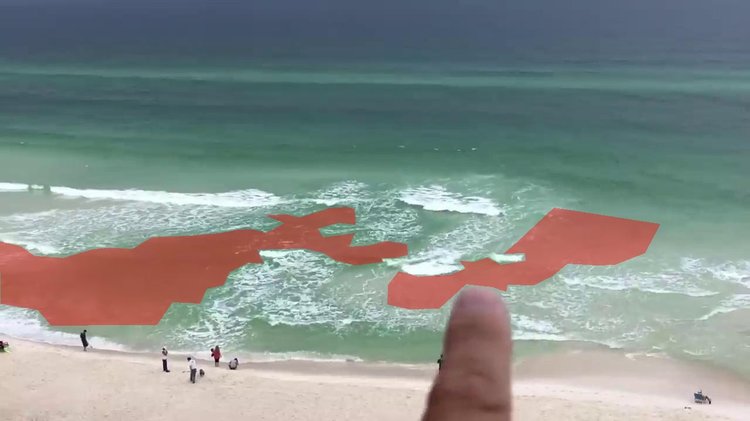} &
     \includegraphics[width=0.195\textwidth]{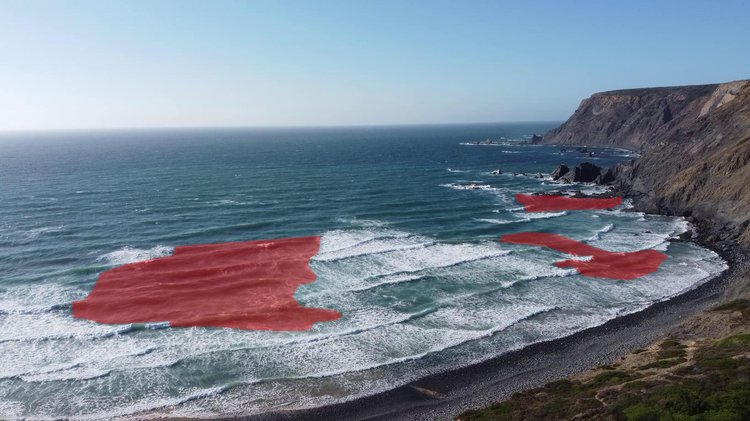} 
     \tabularnewline
     \vspace{-1mm} 
     \includegraphics[width=0.195\textwidth]{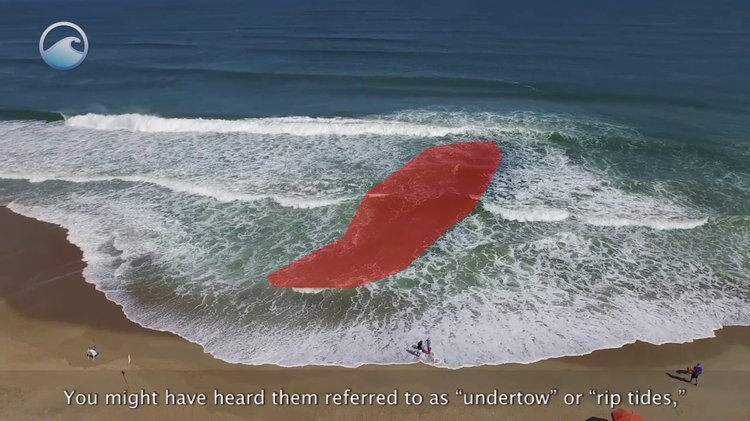} &
     \includegraphics[width=0.195\textwidth]{figures/dataset-variety/mask/RipVIS-084-middle_frame.jpg} &
     \includegraphics[width=0.195\textwidth]{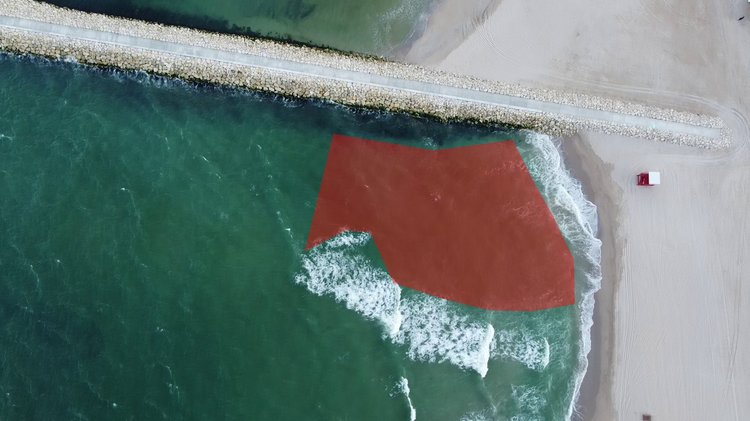} &
     \includegraphics[width=0.195\textwidth]{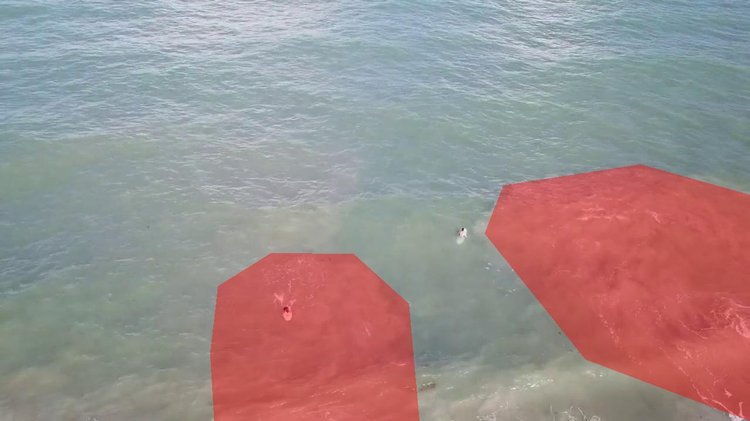} &
     \includegraphics[width=0.195\textwidth]{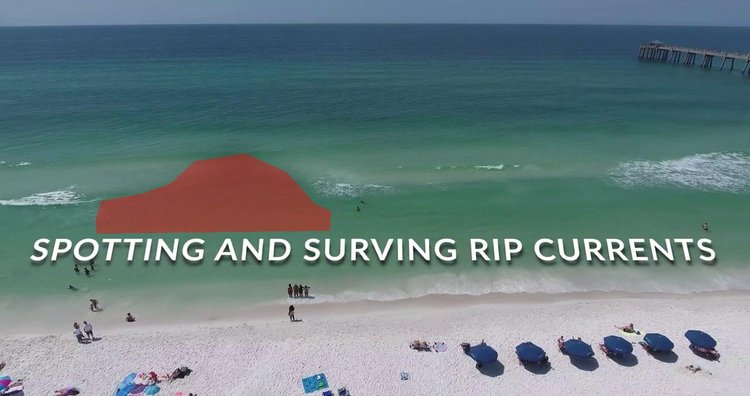} 
     \tabularnewline
\end{tabular}
  \caption{The same examples as before, with the ground truth masks overlayed on top. Pay special attention to the rip currents with sediments. How many did you get right?}
  \label{fig:dataset_diversity_masked}
\end{figure*}

\begin{figure*}
\centering
\setlength{\tabcolsep}{1pt}
\begin{tabular}{c c c c c}
     Original Image & Prediction & Prediction + TCA & Pred. + Filtered TCA  & Ground Truth \tabularnewline
     \vspace{-1mm} 
     \begin{turn}{90} {\raggedright Frame 007} \end{turn}
     \includegraphics[width=0.19\textwidth]{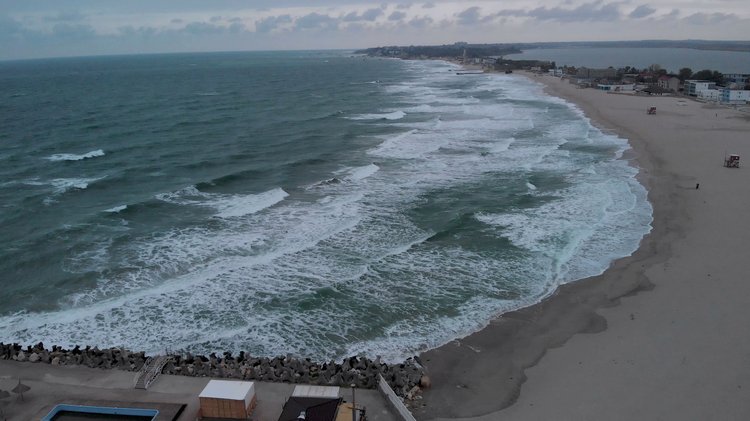} &
     \includegraphics[width=0.19\textwidth]{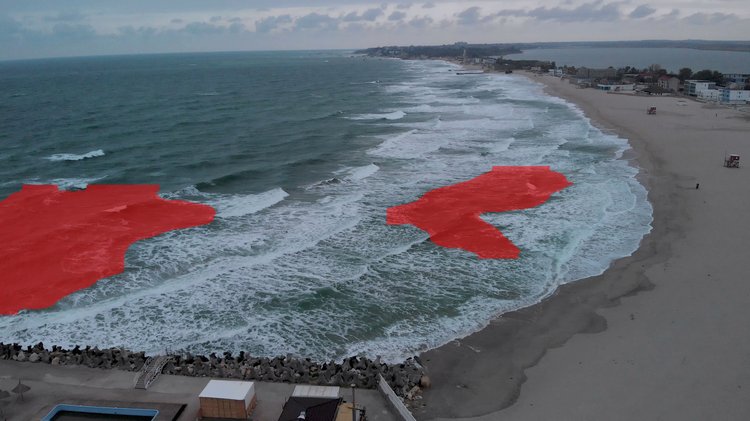} &
     \includegraphics[width=0.19\textwidth]{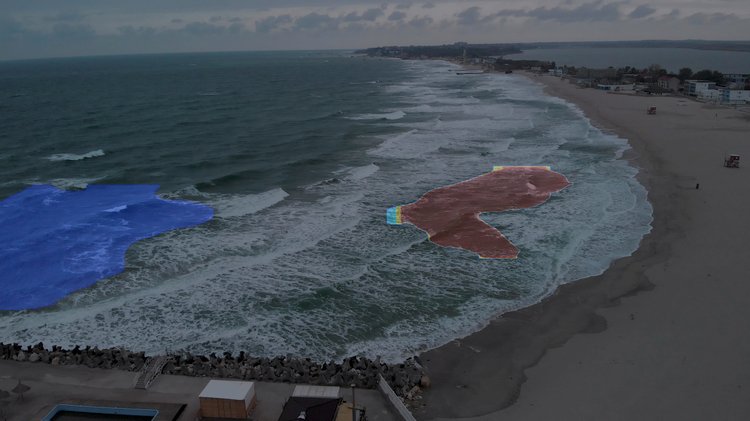} &
     \includegraphics[width=0.19\textwidth]{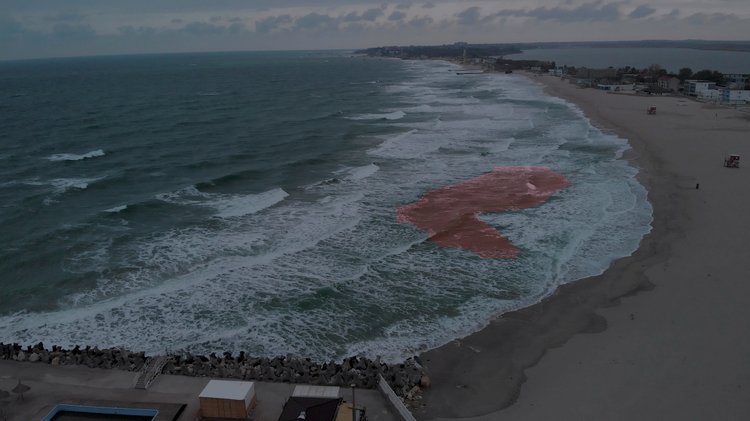} &
     \includegraphics[width=0.19\textwidth]{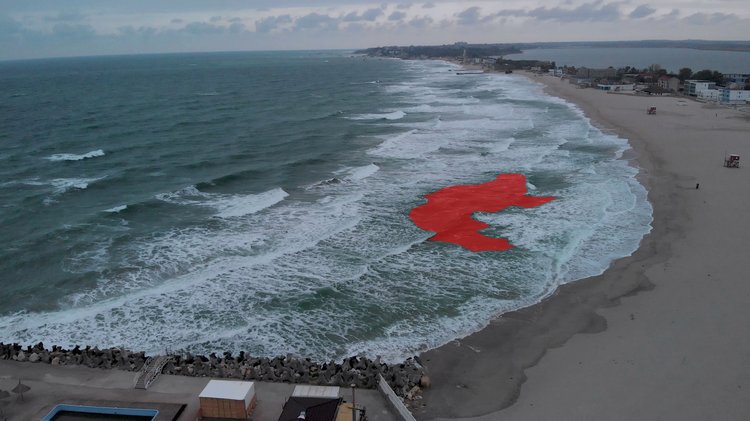} 
     \tabularnewline
     \vspace{-1mm} 
     \begin{turn}{90} {\raggedright Frame 035} \end{turn}
     \includegraphics[width=0.19\textwidth]{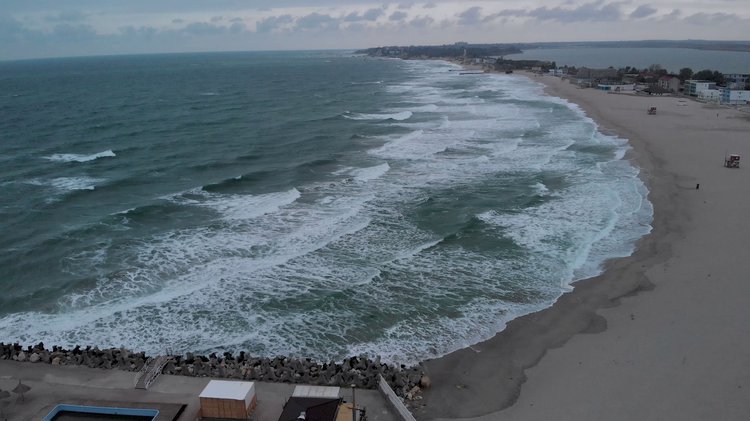} &
     \includegraphics[width=0.19\textwidth]{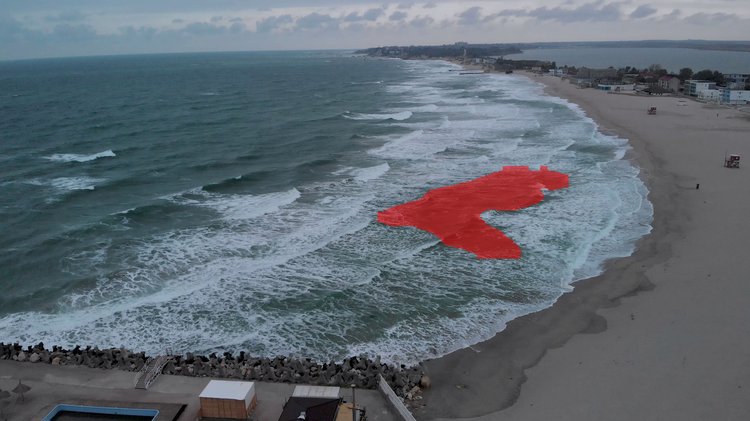} &
     \includegraphics[width=0.19\textwidth]{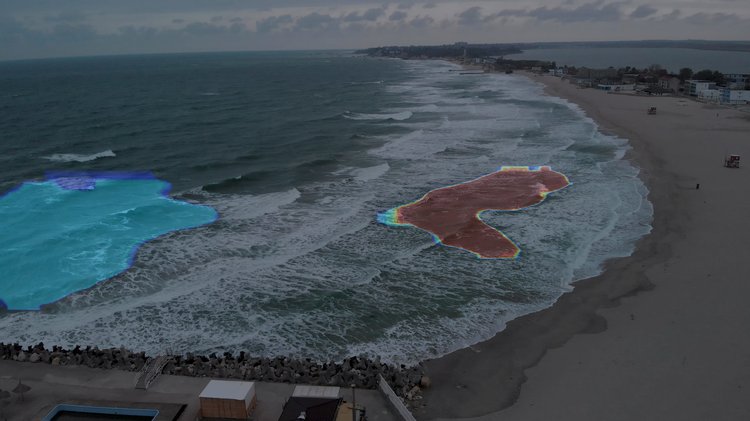} &
     \includegraphics[width=0.19\textwidth]{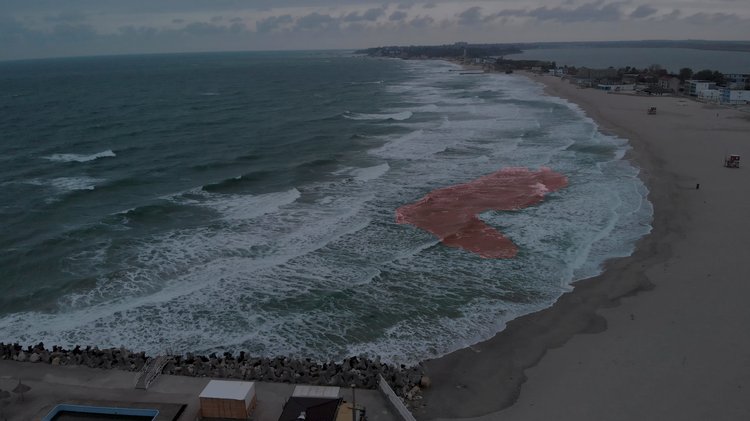} &
     \includegraphics[width=0.19\textwidth]{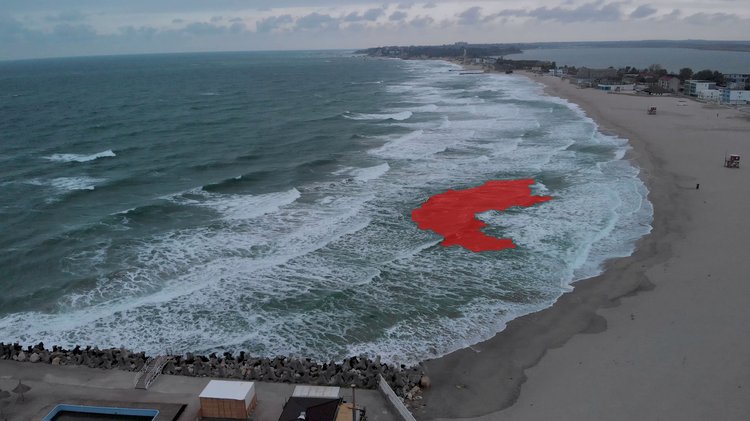} 
     \tabularnewline
     \vspace{-1mm} 
     \begin{turn}{90} {\raggedright Frame 062} \end{turn}
     \includegraphics[width=0.19\textwidth]{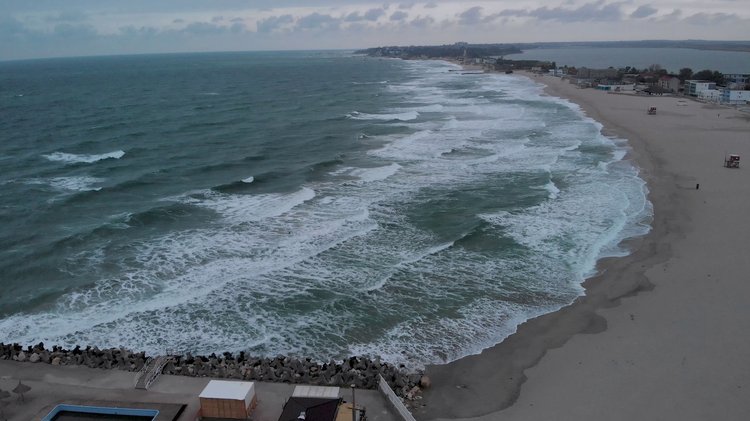} &
     \includegraphics[width=0.19\textwidth]{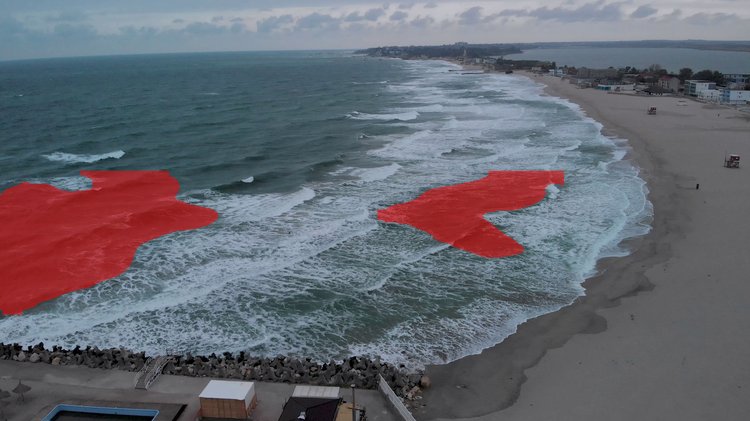} &
     \includegraphics[width=0.19\textwidth]{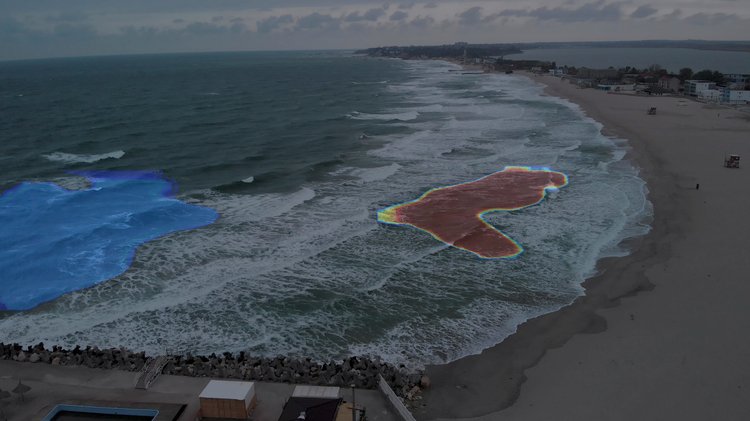} &
     \includegraphics[width=0.19\textwidth]{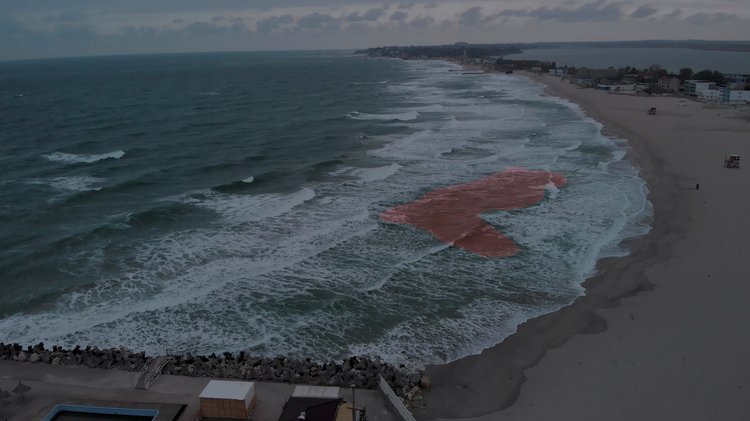} &
     \includegraphics[width=0.19\textwidth]{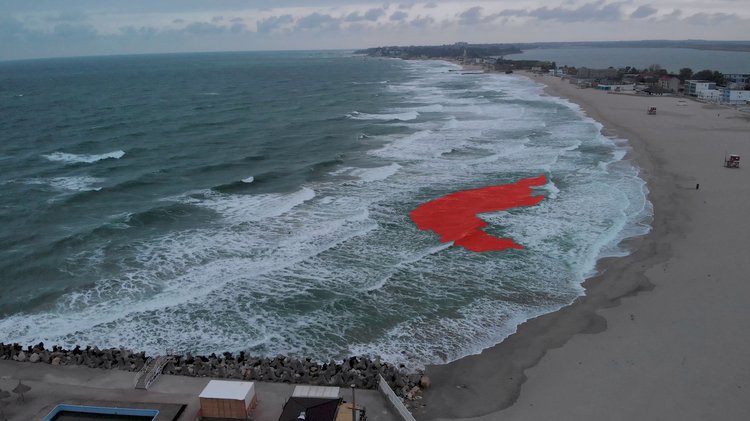} 
     \tabularnewline
     \vspace{-1mm} 
     \begin{turn}{90} {\raggedright Frame 145} \end{turn}
     \includegraphics[width=0.19\textwidth]{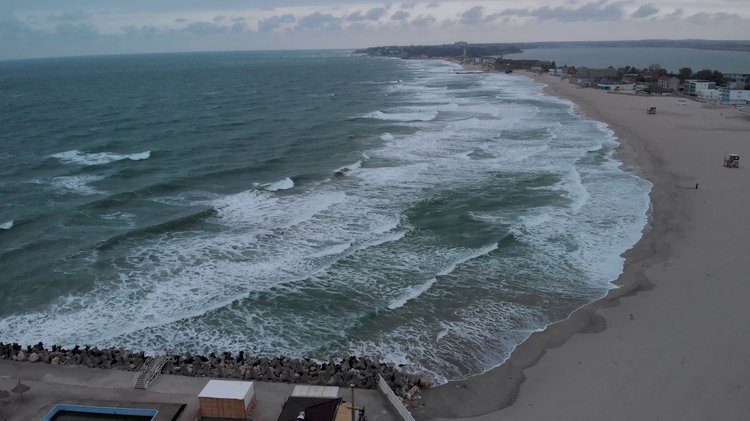} &
     \includegraphics[width=0.19\textwidth]{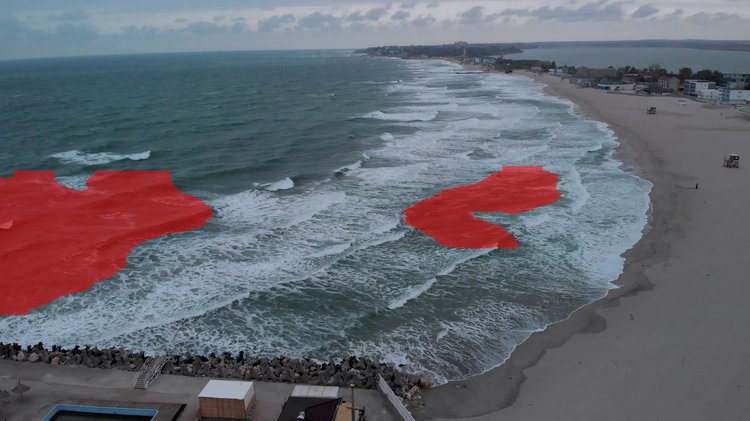} &
     \includegraphics[width=0.19\textwidth]{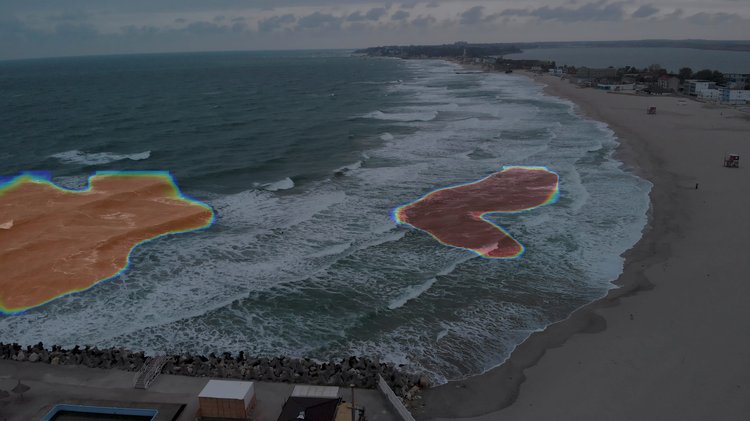} &
     \includegraphics[width=0.19\textwidth]{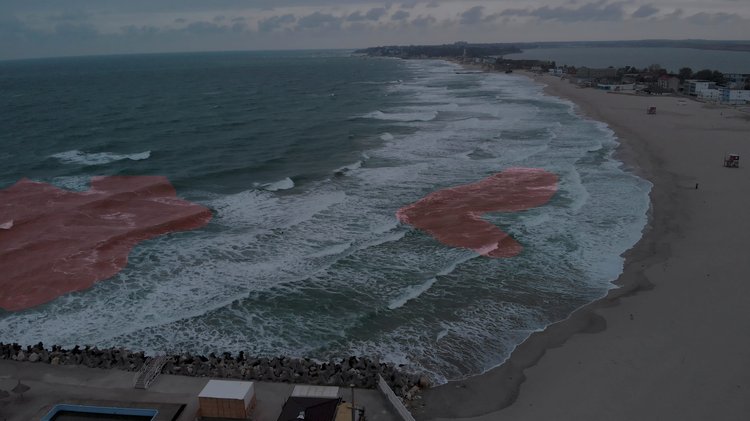} &
     \includegraphics[width=0.19\textwidth]{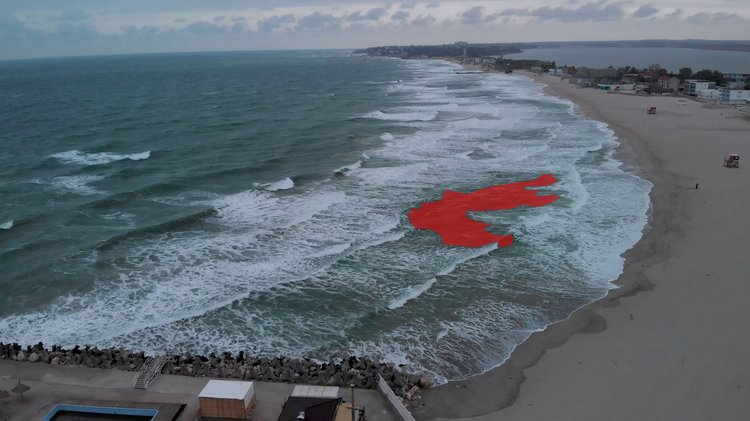} 
     \tabularnewline
     \vspace{-1mm} 
     \begin{turn}{90} {\raggedright Frame 265} \end{turn}
     \includegraphics[width=0.19\textwidth]{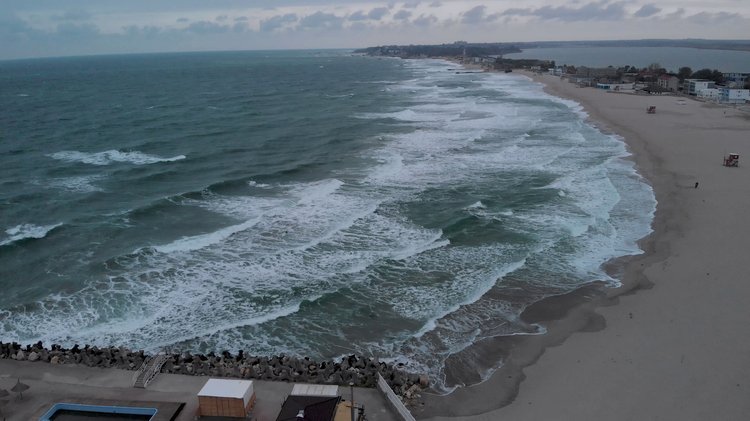} &
     \includegraphics[width=0.19\textwidth]{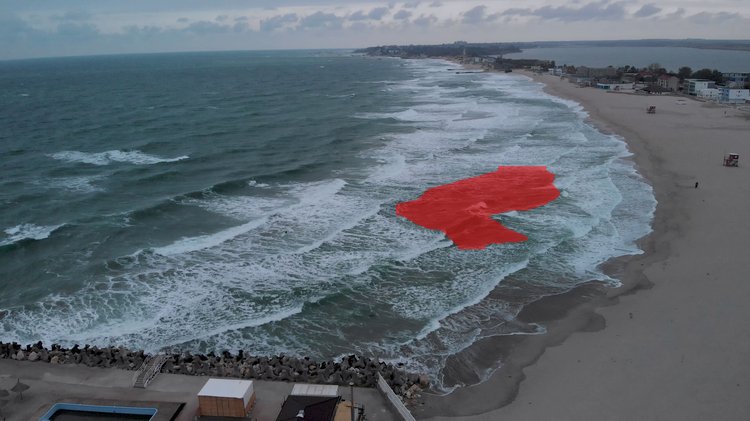} &
     \includegraphics[width=0.19\textwidth]{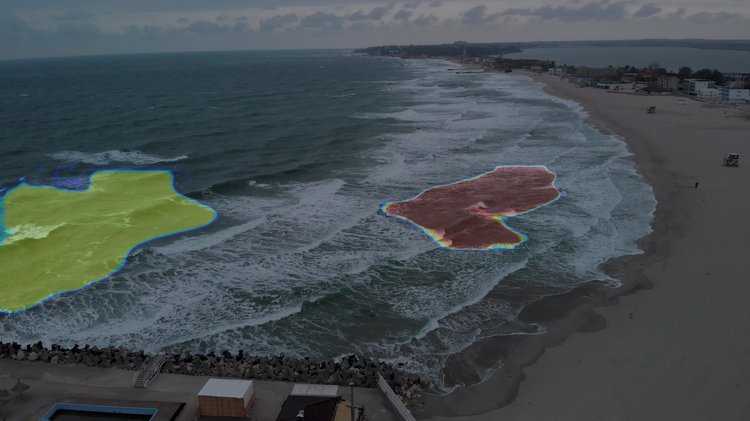} &
     \includegraphics[width=0.19\textwidth]{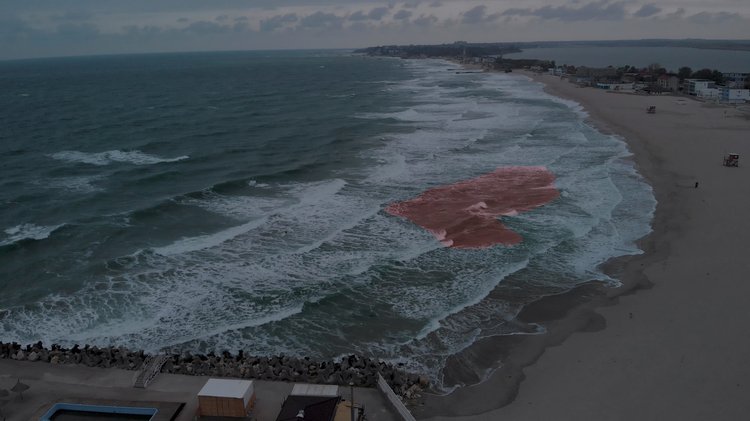} &
     \includegraphics[width=0.19\textwidth]{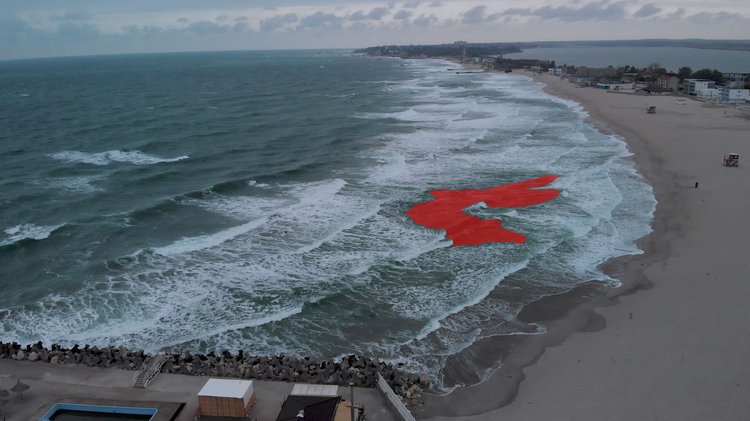} 
     \tabularnewline
\end{tabular}
  \caption{In this situation, TCA manages to filter many false positives, but not all. Too many false positives in a row get accumulated into a final detection (frames 062 - 145). Many false positives are on and off, though, and TCA helps filter most of them.}
  \label{fig:tca_mixed}
\end{figure*}

\begin{figure*}
\centering
\setlength{\tabcolsep}{1pt}
\begin{tabular}{c c c c c}
     Original Image & Prediction & Prediction + TCA & Pred. + Filtered TCA  & Ground Truth \tabularnewline
    
     \vspace{-1mm} 
     \begin{turn}{90} {\raggedright Frame 063} \end{turn}
     \includegraphics[width=0.19\textwidth]{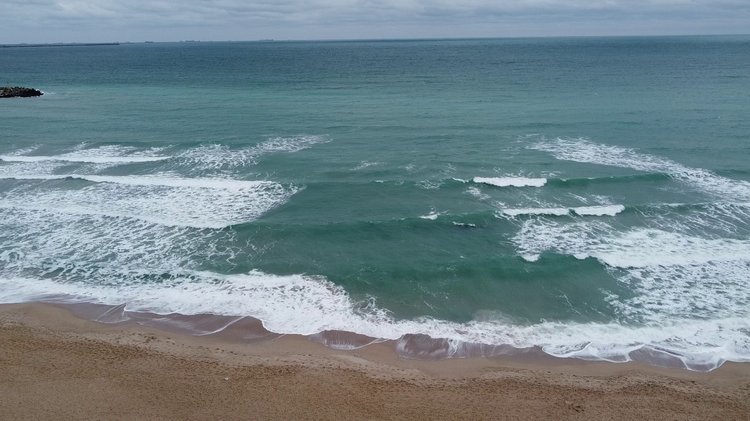} &
     \includegraphics[width=0.19\textwidth]{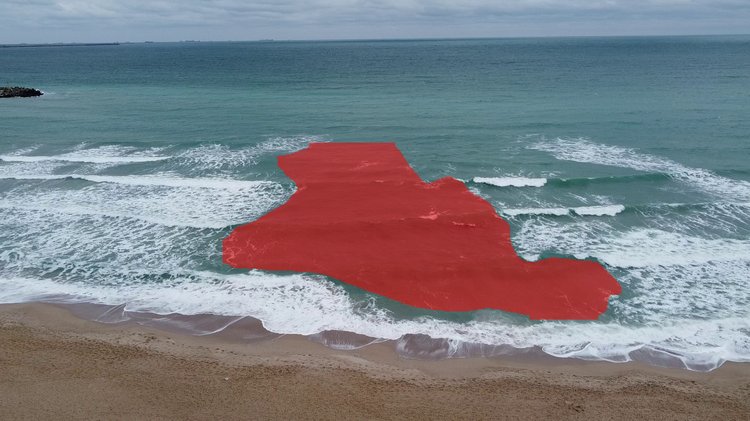} &
     \includegraphics[width=0.19\textwidth]{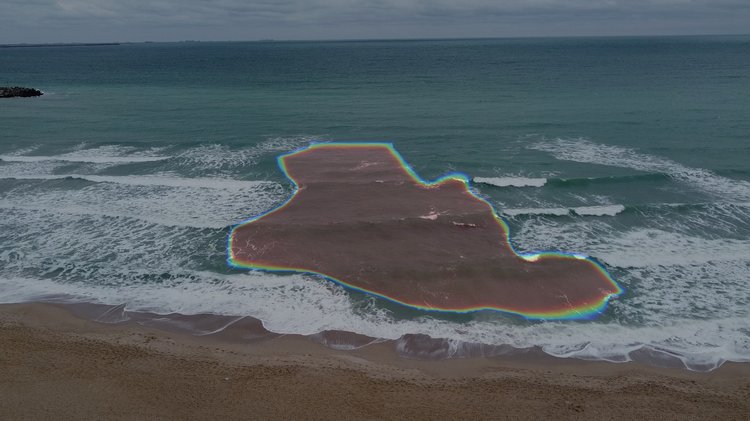} &
     \includegraphics[width=0.19\textwidth]{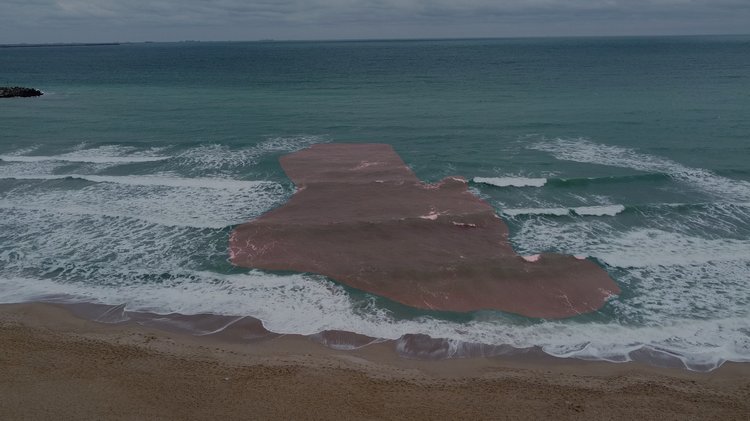} &
     \includegraphics[width=0.19\textwidth]{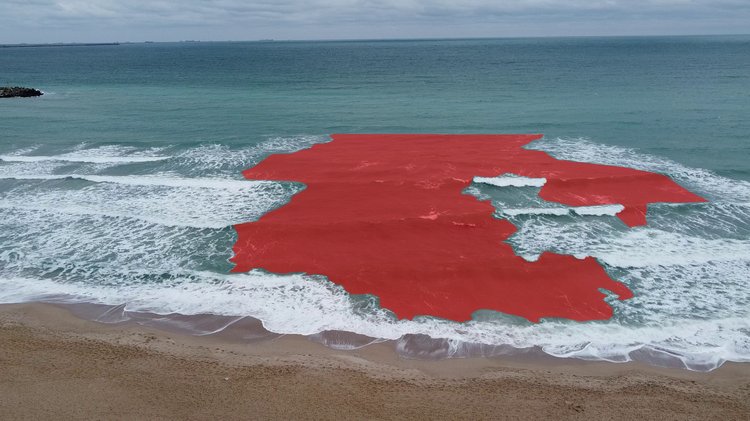} 
     \tabularnewline
     \vspace{-1mm} 
     \begin{turn}{90} {\raggedright Frame 270} \end{turn}
     \includegraphics[width=0.19\textwidth]{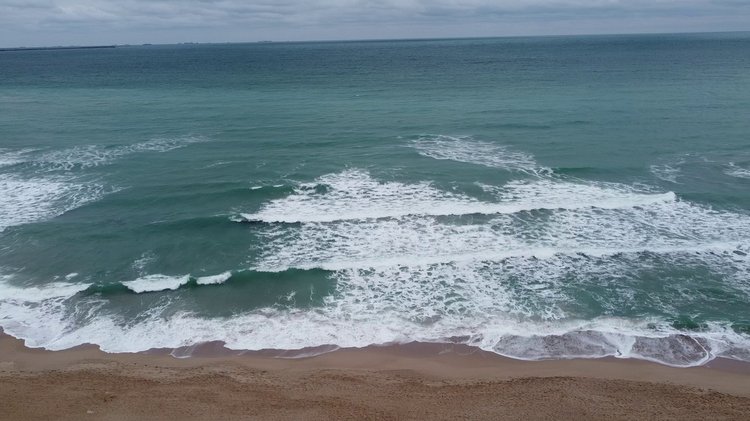} &
     \includegraphics[width=0.19\textwidth]{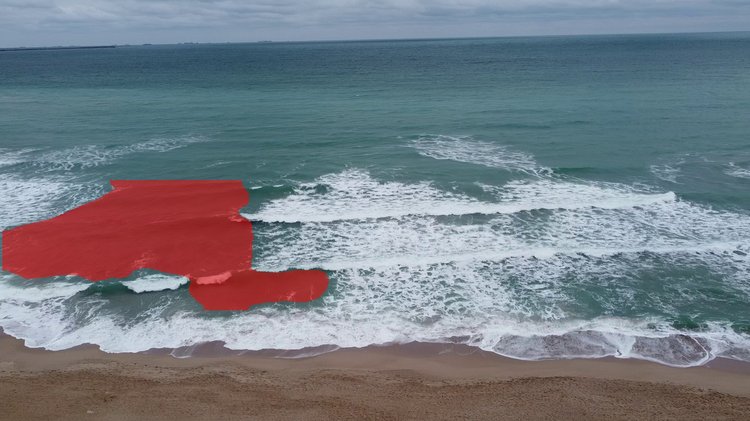} &
     \includegraphics[width=0.19\textwidth]{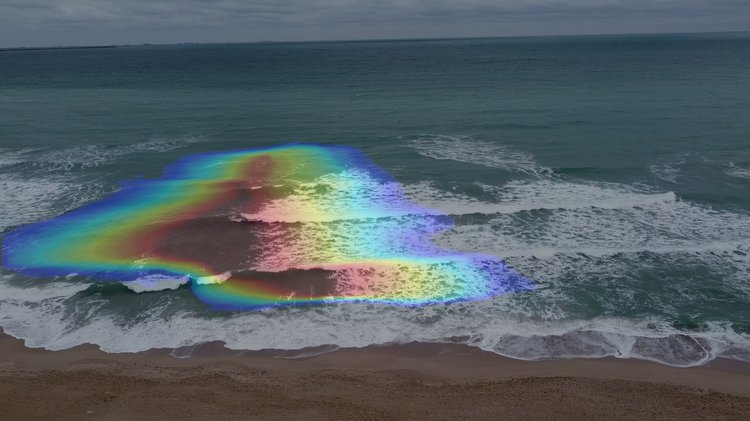} &
     \includegraphics[width=0.19\textwidth]{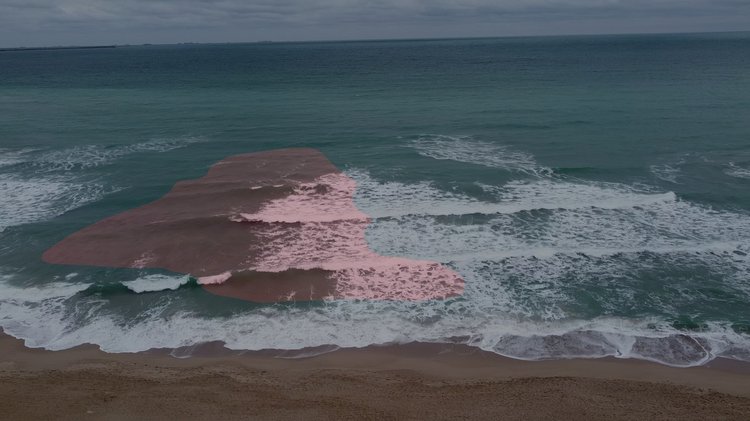} &
     \includegraphics[width=0.19\textwidth]{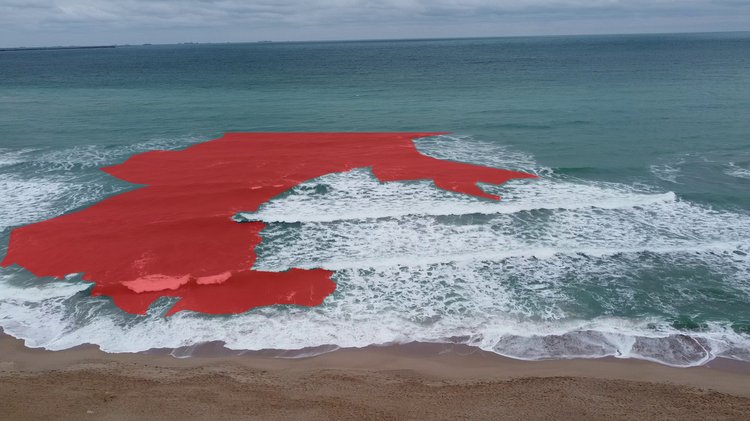} 
     \tabularnewline     
     \vspace{-1mm} 
     \begin{turn}{90} {\raggedright Frame 323} \end{turn}
     \includegraphics[width=0.19\textwidth]{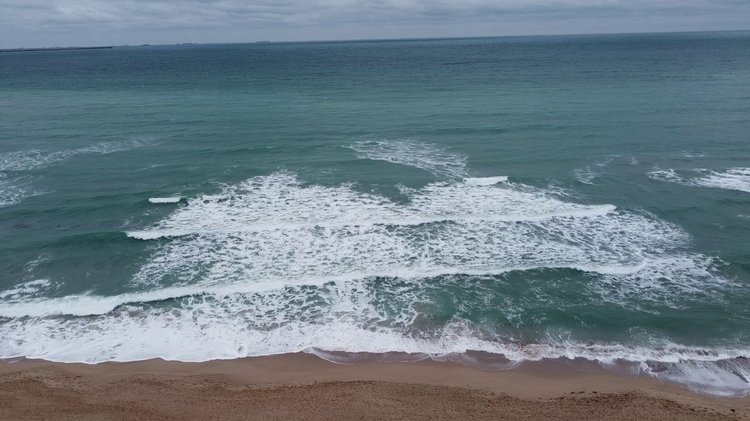} &
     \includegraphics[width=0.19\textwidth]{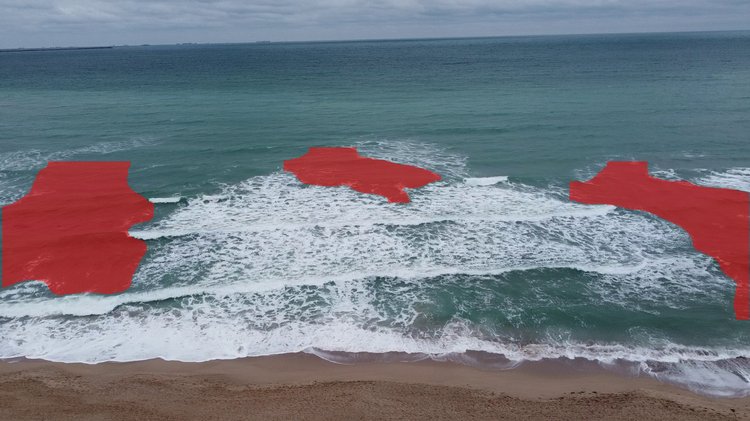} &
     \includegraphics[width=0.19\textwidth]{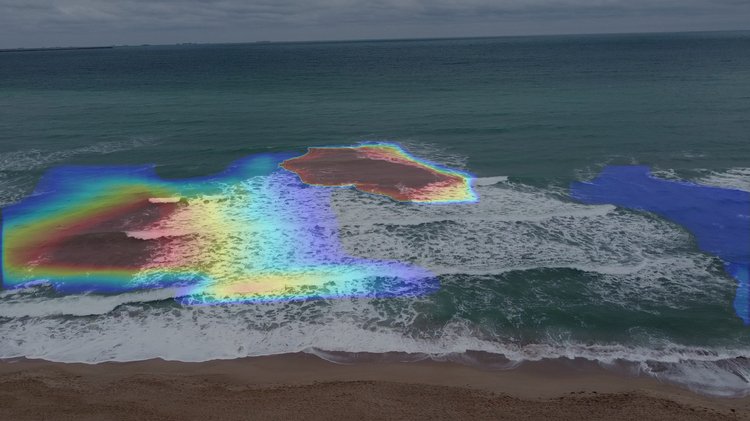} &
     \includegraphics[width=0.19\textwidth]{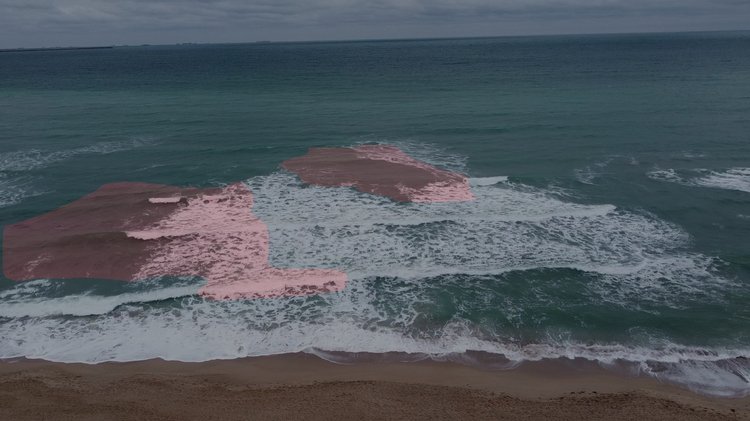} &
     \includegraphics[width=0.19\textwidth]{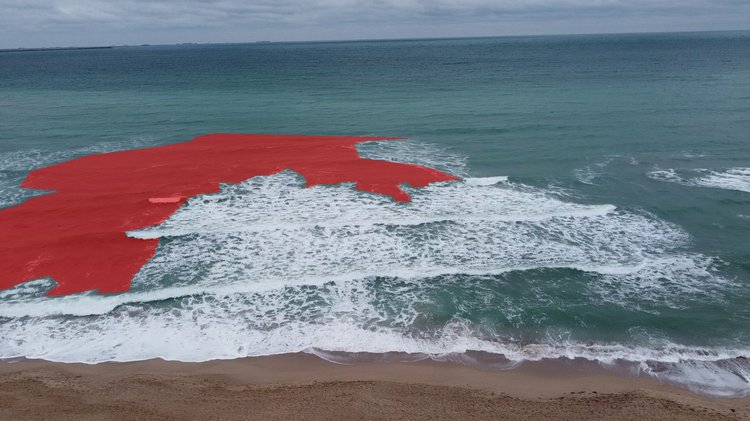} 
     \tabularnewline     
     \vspace{-1mm} 
     \begin{turn}{90} {\raggedright Frame 380} \end{turn}
     \includegraphics[width=0.19\textwidth]{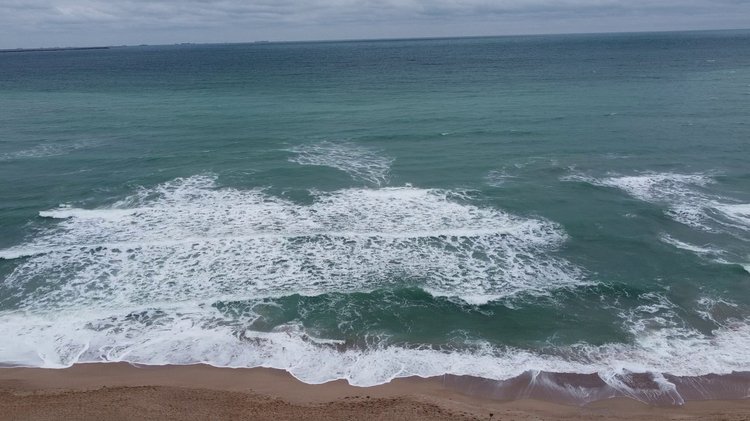} &
     \includegraphics[width=0.19\textwidth]{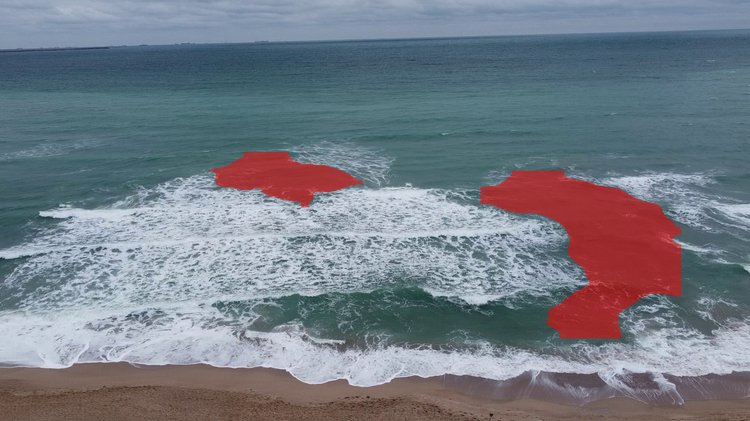} &
     \includegraphics[width=0.19\textwidth]{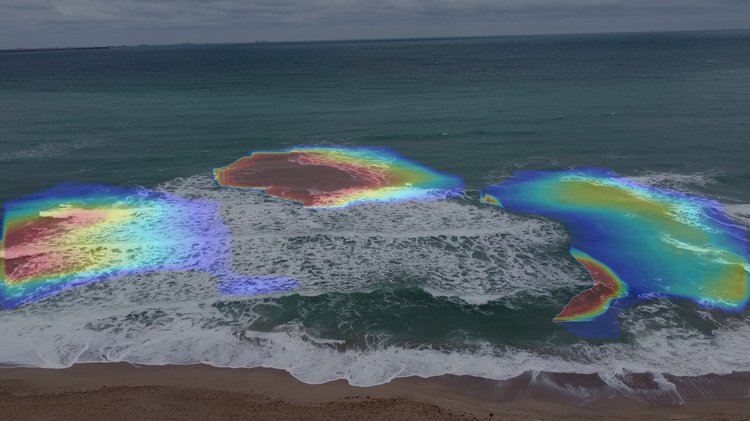} &
     \includegraphics[width=0.19\textwidth]{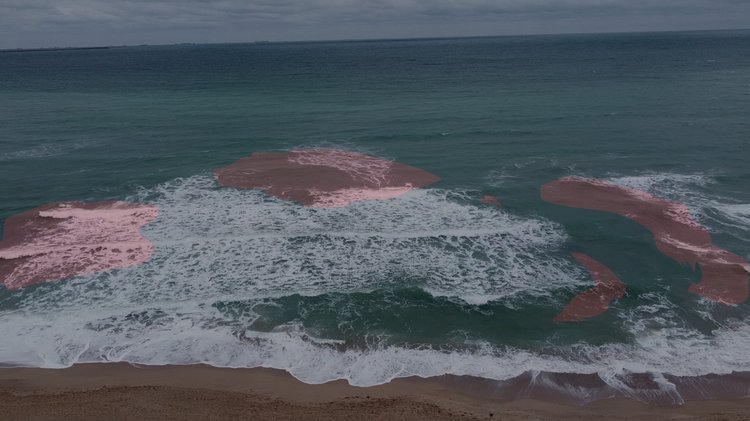} &
     \includegraphics[width=0.19\textwidth]{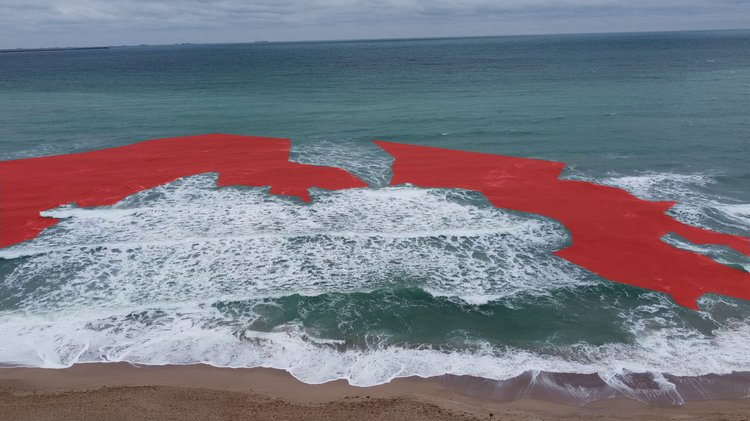} 
     \tabularnewline     
     \vspace{-1mm} 
     \begin{turn}{90} {\raggedright Frame 447} \end{turn}
     \includegraphics[width=0.19\textwidth]{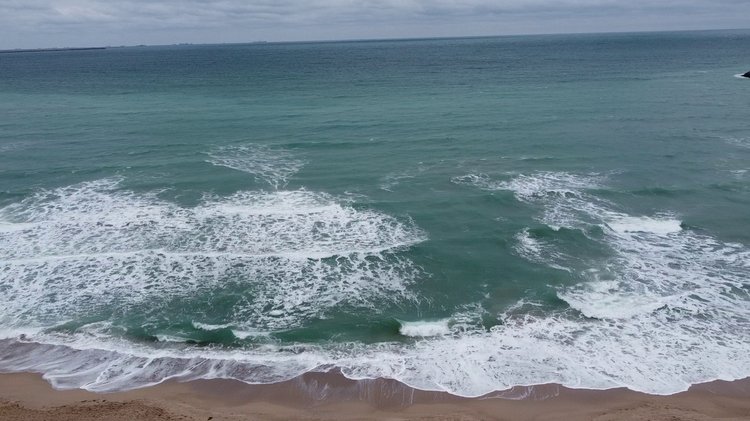} &
     \includegraphics[width=0.19\textwidth]{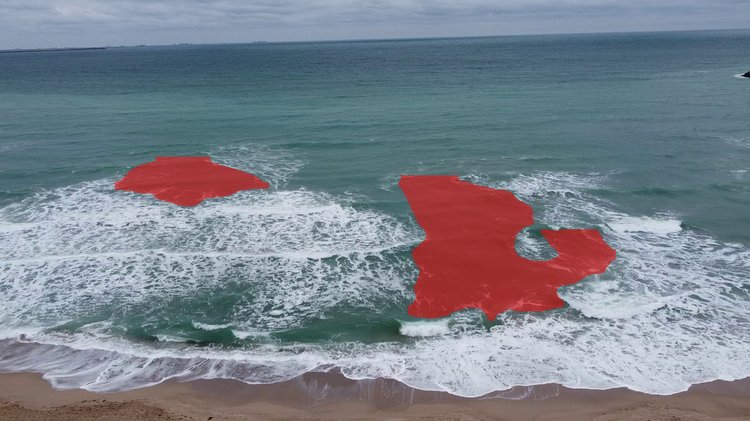} &
     \includegraphics[width=0.19\textwidth]{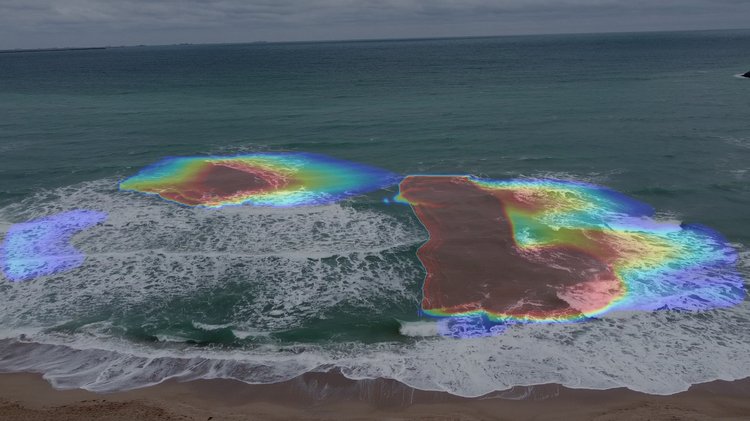} &
     \includegraphics[width=0.19\textwidth]{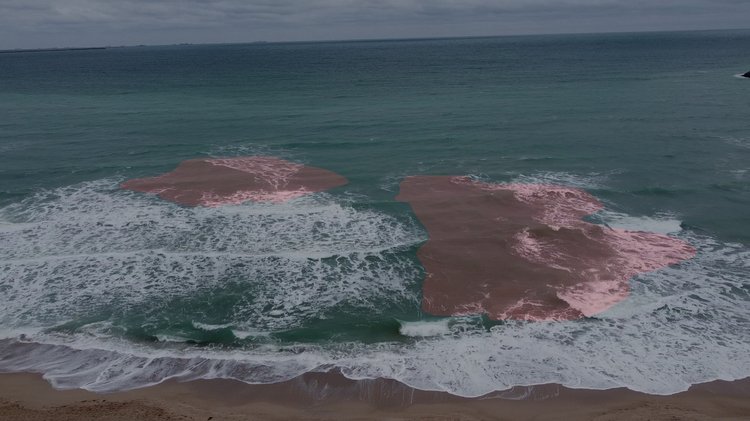} &
     \includegraphics[width=0.19\textwidth]{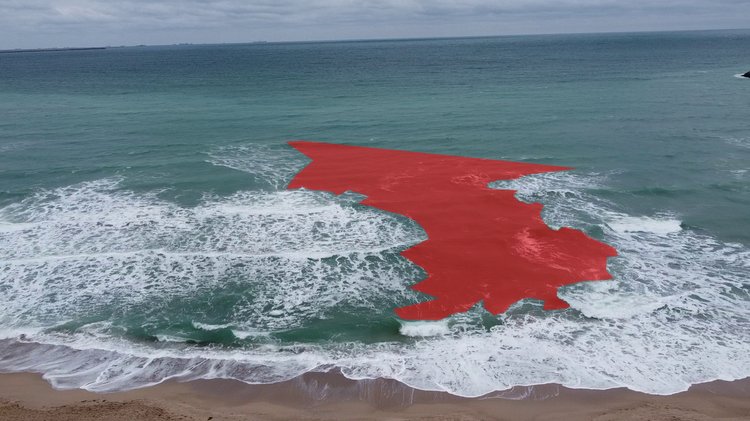} 
     \tabularnewline     
\end{tabular}
  \caption{An example where TCA does more harm than good, if the camera is moving fast enough (in this case, the drone is dashing along the beachfront).}
  \label{fig:tca_fail}
\end{figure*}

\end{document}